\documentclass{article} % For LaTeX2e
\usepackage{iclr2027_conference,times}

\iclrfinalcopy
\usepackage{amsmath,amsfonts,bm}

\def\eqref#1{equation~\ref{#1}}
\def\1{\bm{1}}

\DeclareMathAlphabet{\mathsfit}{\encodingdefault}{\sfdefault}{m}{sl}
\SetMathAlphabet{\mathsfit}{bold}{\encodingdefault}{\sfdefault}{bx}{n}

\usepackage[table]{xcolor}
\definecolor{rowgray}{gray}{0.92}
\definecolor{omnigreen}{RGB}{226,241,233}
\definecolor{omniink}{RGB}{31,45,80}
\usepackage{url}
\usepackage{tabularx}
\usepackage{booktabs}
\usepackage{graphicx}
\usepackage{float}
\usepackage{hyperref}
\newcommand{\logo}{\raisebox{-2pt}[0pt][0pt]{\includegraphics[height=1.5em]{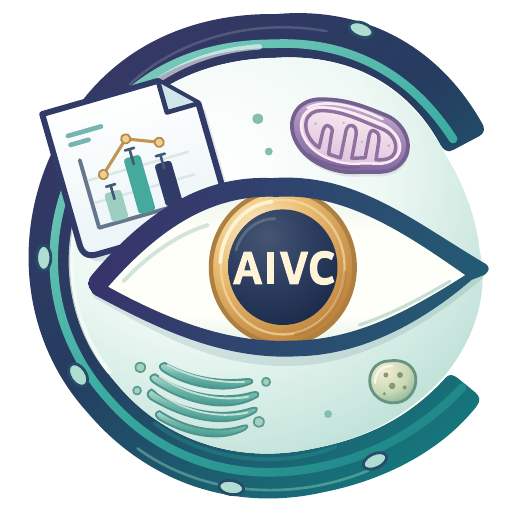}}}

\title{\logo OmniVCBench: Benchmarking Evidence-Grounded Multimodal Reasoning Towards AI Virtual Cells}

\author{Manyu Li \\
Fudan University, Shanghai, China \\
\texttt{24210240029@m.fudan.edu.cn} 
\AND
Xunkai Li \\
Beijing Institute of Technology, Beijing, China \\
\texttt{cs.xunkai.li@gmail.com} 
\AND
Yongfu Xiong \& Yi Liu \\
Chongqing Ant Consumer Finance Co,. Ltd \\
\texttt{\{xiongyongfu.xyf, larry.liuy\}@myxiaojin.cn}
\AND 
Rong-Hua Li \thanks{Corresponding Authors.} \& Guoren Wang \\
Beijing Institute of Technology, Beijing, China \\
\texttt{\{lironghuabit, wanggrbit\}@162.com}
}

\begin{document}

\maketitle

\begin{abstract}
\textit{Artificial Intelligence Virtual Cells} (AIVCs) are envisioned as scientific agents that simulate cellular responses, explain underlying mechanisms, and support hypothesis-driven discovery. Existing AIVC benchmarks, however, operate primarily at the \emph{simulation layer}, motivating complementary evaluation of how models interpret experimental evidence and formulate biological hypotheses. We introduce \textbf{\textsc{OmniVCBench}}, a figure-centric, source-traceable benchmark for the \emph{interpretation component} of an AIVC. It contains 6,077 curated single- and multi-subfigure question--answer pairs derived from figures and experimental contexts in the scientific literature. Guided by Bloom's taxonomy, we instantiate interpretation-layer counterparts of the AIVC Predict--Explain--Discover agenda through three scientific reasoning tasks. These comprise L1 evidence-conditioned inference, L2 mechanistic explanation, and L3 evidence-grounded hypothesis proposal. We further introduce \textbf{AIVC-Judge}, a task-conditioned MLLM-as-a-judge framework with category-specific, reference-aware rubrics for evaluating open-ended responses. A complementary Model-Derived Hard-Negative Mining (MDHNM) strategy converts plausible errors observed during model inference into MCQ distractors for lower-cost evaluation. Within the evaluated heterogeneous model pool, MCQ accuracy correlates positively with AIVC-Judge scores, providing a complementary view of performance alongside open-response evaluation. Evaluating both proprietary and open-weight multimodal models shows that even the best proprietary model reaches only 3.28 out of 5.00 under AIVC-Judge. Supervised fine-tuning and retrieval-augmented generation on the auxiliary \textbf{\textsc{OmniVCTrain}} corpus provide modest, configuration-dependent gains. Together, \textsc{OmniVCBench} combines source-aligned cellular figures, interpretation-role organization around Predict--Explain--Discover, and paired open-response and controlled MCQ evaluation. It provides a source-traceable resource for assessing evidence-grounded reasoning in candidate AIVC interpretation components. Code and data demo are available at \url{https://anonymous.4open.science/r/OmniVCBench}.

\end{abstract}

\section{Introduction}

\label{sec:introduction}

\begin{figure}[t]
  \vspace{0pt}
  \centering
  \includegraphics[width=0.85\linewidth]{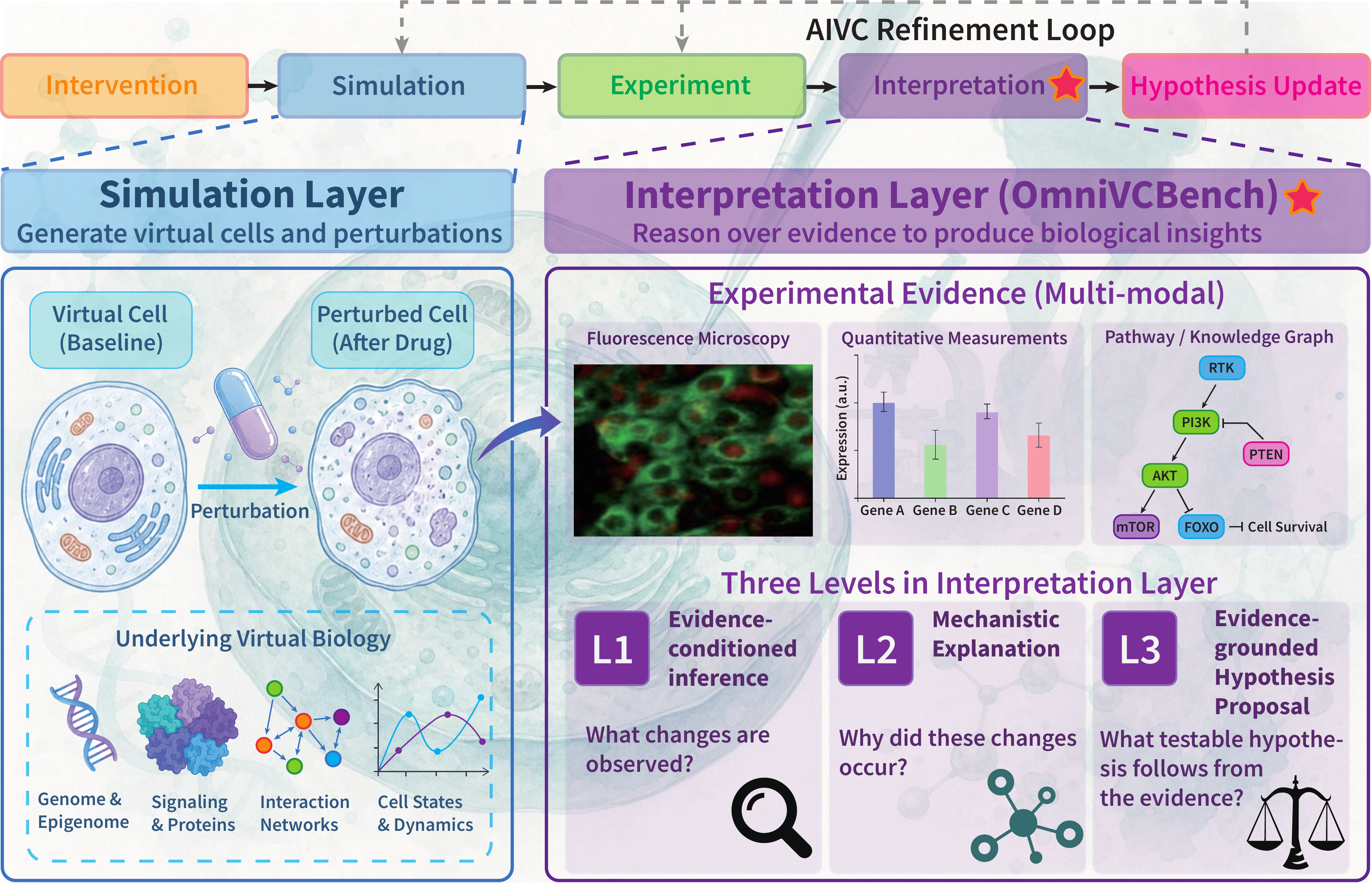}
  \vspace{-2mm}
  \caption{\textbf{The interpretation component within the AIVC workflow.} The simulation layer predicts perturbed cell states; experiments return observations; the interpretation layer compares and explains the evidence, informing the next experiment or model revision (feedback arrow). \textsc{OmniVCBench} evaluates only the interpretation component (highlighted), through three scientific task roles (L1--L3); the remaining stages are application context.}
  \label{fig:overview}
  \vspace{-3mm}
\end{figure}

Artificial intelligence virtual cells (AIVCs) aim to predict cellular behavior, explain biological mechanisms, and support hypothesis-driven discovery \citep{bunne2024build,noutahi2025virtual}. The Predict--Explain--Discover (P-E-D) agenda connects these functions into a workflow in which simulation is only the beginning: a predicted response motivates an experiment, and its value is realized when the results are distilled into a mechanistic explanation and a testable next hypothesis. Accurate state prediction alone cannot show whether a system can interpret experimental observations, recognize disagreement with new evidence, or propose what to test next. Current virtual-cell benchmarks primarily evaluate the \emph{simulation layer}, measuring how accurately models predict cellular states and phenotypes under intervention \citep{roohani2025virtual,szalata2024benchmark,wu2025perturbench,wei2026benchmarking,mao2026benchmarking,li2026mvcbench}; this complementary capability is not yet organized and tested systematically in AIVC evaluation. Figure~\ref{fig:overview} locates both capabilities within the same workflow, linking simulation to the scientific use of its predictions.

This division of labor raises a role-assignment question that current virtual-cell evaluations leave open: \textbf{what role should multimodal LLMs play in an AI virtual cell, and how do they close the loop with the simulation layer?} Simulation-layer models operate on omics profiles, yet the recorded products of cellular experiments---fluorescence microscopy fields, immunoblots, dose--response curves, and composite multi-panel figures---are visual, and using these results scientifically means reading them. We therefore distinguish the \emph{MLLM component}, which maps figures and questions to evidence-grounded answers, from the \emph{MLLM agent}, which adds planning, tool use, and iterative refinement around it; the agent's reliability depends on this component's interpretation quality. The component's domain is what we call the \emph{interpretation layer}. Confronted with an experimental result, a researcher asks in sequence: what do the observations support, how can the phenomenon be explained, and what can be tested next; these become three task roles: \textbf{L1 evidence-conditioned inference} derives a result from displayed evidence---retrospective decoding of what is observed, confirming or contradicting prior expectations, distinct from the simulation layer's forward forecasting of unobserved states; \textbf{L2 mechanistic explanation} accounts for results through biological processes; and \textbf{L3 evidence-grounded hypothesis proposal} formulates a testable extension of the observations. \textsc{OmniVCBench} operationalizes this component-level evaluation, with Section~\ref{sec:refinement-loop} exercising the component inside executable loops.

Scientific-figure benchmarks already assess important parts of this reasoning process \citep{roberts2024scifibench,li2024mmsci,zhang2025hiscibench}, including hypothesis generation and experiment proposal in biology-focused settings \citep{burgess2025microvqa,laurent2026labbench2}. We build on these precedents with \textbf{\textsc{OmniVCBench}}, a cellular-evidence benchmark organized around the interpretation roles of an AIVC. Its 6,077 questions draw on paper-linked figures and experimental contexts from OmniScience \citep{tao2026omniscience}, combining single- and multi-subfigure questions with a traceable source record per answer. Published experiments are the evidence source because their figures provide traceable observations from real experimental systems within a closed, reviewable record; they remain a proxy for the live workflow, covering curated scientific communication artifacts---annotated, post-processed figures---rather than primary instrument readouts, raw-data quality, or failed experiments. We evaluate general-purpose multimodal LLMs as candidate interpretation components from the image and question alone, an interface complementing the omics-native inputs and outputs of single-cell foundation models \citep{cui2024scgpt,theodoris2023transfer,hao2024large}.

\textbf{AIVC-Judge} scores open responses against source references with claim-level evidence traces; \textbf{Model-Derived Hard-Negative Mining (MDHNM)} turns observed inference errors into discriminative MCQ options for low-cost comparison; a unanimous three-annotator audit decides which candidates enter the benchmark; and human rescoring checks the judge on shared textual evidence. \textbf{\textsc{OmniVCTrain}} supplies an adaptation resource, and executable pilots test the component inside the loop. These pieces yield four contributions:
\begin{itemize}
  \item A source-traceable benchmark of 6,077 cellular-research questions spanning inference, explanation, and hypothesis proposal in the interpretation layer.
  \item Paired open-response and MCQ evaluation on identical items, with human-checked reference-aware scoring, model-derived distractors, and a unanimous three-annotator item audit.
  \item Evaluation of proprietary and open-weight MLLMs around four research questions: judgment reliability, cross-format agreement, adaptation gains, and where explanation and hypothesis proposal fall short.
  \item Executable evidence-acquisition pilots on real biological outputs, testing prediction revision inside the loop's evidence-acquisition and revision step (Section~\ref{sec:refinement-loop}).
\end{itemize}

\section{Related Work}

\subsection{Virtual-Cell Evaluation}
AIVC roadmaps connect prediction of cellular behavior with mechanistic explanation and scientific discovery \citep{bunne2024build,noutahi2025virtual}. Perturbation benchmarks evaluate the simulation layer: the Virtual Cell Challenge and OP3 assess intervention responses \citep{roohani2025virtual,szalata2024benchmark}; PerturBench and scPerturBench standardize generalization tests \citep{wu2025perturbench,wei2026benchmarking}; and VCBench targets in-the-wild responses \citep{mao2026benchmarking}. MVCBench adds transcriptomic and morphological outputs \citep{li2026mvcbench}, and VCWorld adds knowledge-guided reasoning to simulation \citep{wei2026vcworld}. PerturbQA, CellVerse, and SC-Arena extend evaluation to language-based biological reasoning \citep{wu2025contextualizing,zhang2026cellverse,zhao2026sc}, though over textual serializations of omics profiles rather than visual evidence. \textsc{OmniVCBench} complements these simulation-layer settings (Table~\ref{tab:benchmark-landscape}) by evaluating natural-language interpretation of observed cellular figures.

\begin{table}[t]
  \caption{Recent virtual-cell and scientific-figure benchmarks.}
  \label{tab:benchmark-landscape}
  \centering
  \scriptsize
  \setlength{\tabcolsep}{3pt}
  \begin{tabularx}{\linewidth}{@{}>{\raggedright\arraybackslash}p{0.34\linewidth}>{\raggedright\arraybackslash}p{0.14\linewidth}>{\raggedright\arraybackslash}X>{\raggedright\arraybackslash}p{0.115\linewidth}@{}}
    \toprule
    Benchmark & Scale & What it evaluates / contains & Layer \\
    \midrule
    Virtual Cell Challenge~\citep{roohani2025virtual} & $\sim$300k cells & CRISPRi perturbation responses & Simulation \\
    \rowcolor{rowgray} OP3~\citep{szalata2024benchmark} & 144 compounds & Held-out perturbation expression & Simulation \\
    PerturBench~\citep{wu2025perturbench} & 6 datasets & Perturbation-response prediction & Simulation \\
    \rowcolor{rowgray} scPerturBench~\citep{wei2026benchmarking} & 29 datasets & Unseen perturbations and contexts & Simulation \\
    VCBench~\citep{mao2026benchmarking} & 7 datasets & In-the-wild perturbation response & Simulation \\
    \rowcolor{rowgray} MVCBench~\citep{li2026mvcbench} & $\sim$1.1M profiles & Transcriptomic and morphology outputs & Simulation \\
    MicroVQA~\citep{burgess2025microvqa} & 1,042 MCQs & Microscopy interpretation, hypothesis and experiment proposal & Interpretation \\
    \rowcolor{rowgray} LABBench2~\citep{laurent2026labbench2} & $\sim$1,900 tasks & Figure, protocol, and literature QA & Interpretation \\
    SciFIBench~\citep{roberts2024scifibench} & $\sim$2,000 questions & Scientific-figure interpretation & Interpretation \\
    \rowcolor{rowgray} OmniScience~\citep{tao2026omniscience} & $\sim$1.5M records & Figure--caption--context corpus & Data layer \\
    \textbf{\textsc{OmniVCBench}} & \textbf{6,077 items} & \textbf{Evidence-grounded inference, explanation, and hypothesis proposal} & \cellcolor{omnigreen}\textbf{Interpretation} \\
    \bottomrule
  \end{tabularx}
  \vspace{-4mm}
\end{table}
\subsection{Scientific Multimodal Reasoning}
SciFIBench and MMSci assess scientific-figure understanding across disciplines, and HiSciBench organizes tasks from reading to discovery \citep{roberts2024scifibench,li2024mmsci,zhang2025hiscibench}. The closest biology-focused precedents are MicroVQA and FigQA2 within LABBench2 (Table~\ref{tab:benchmark-landscape}). MicroVQA explicitly tests mechanistic hypothesis generation and experiment proposals alongside expert image interpretation, with multi-image questions and Bloom-based cognitive analysis \citep{burgess2025microvqa}. FigQA2 uses open responses and source-specific questions across Image, Paper, and Retrieval modes \citep{laurent2026labbench2}. Building on this coverage, \textsc{OmniVCBench} curates OmniScience's paper-linked figures, captions, and contexts \citep{tao2026omniscience} into a larger cellular-evidence dataset with consistent L1--L3 organization, open-response and MCQ protocols paired on identical items, and diagnostics of visual grounding, scoring agreement, and model errors. As Table~\ref{tab:benchmark-landscape} summarizes, \textsc{OmniVCBench} complements the full AIVC pipeline at its downstream stage: where perturbation benchmarks validate what a virtual cell predicts, we evaluate how the system reads, explains, and builds on experimental evidence.

\subsection{Task Organization and Response Evaluation}
P-E-D describes scientific functions, while Bloom's taxonomy supplies a complementary account of cognitive demand \citep{noutahi2025virtual,bloom1956handbook,anderson2001taxonomy}. Mechanistic explanations, in particular, connect observations to organized biological entities and activities \citep{machamer2000thinking}. Open-response evaluation uses LLM judges to assess such semantic content \citep{liu2023g,zheng2023judging}, with documented preferences for response styles and their own generations \citep{panickssery2024llm,ye2025justice}. MCQ evaluation instead depends on plausible and discriminative distractors \citep{alhazmi2024distractor}. RefineBot uses solver reflection and iterative rewriting \citep{burgess2025microvqa}, while AutoConverter combines distractor proposal, review, selection, and correctness refinement \citep{zhang2025automated}. MDHNM pools low-scoring open responses and rewrites them into candidate options. Our focus is the connection between these response formats: applying both to the same cellular evidence supports comparing answer selection with generated reasoning.

\section{Method}
\label{sec:method}

\begin{figure}[t]
  \vspace{0pt}
  \centering
  \includegraphics[width=0.88\linewidth]{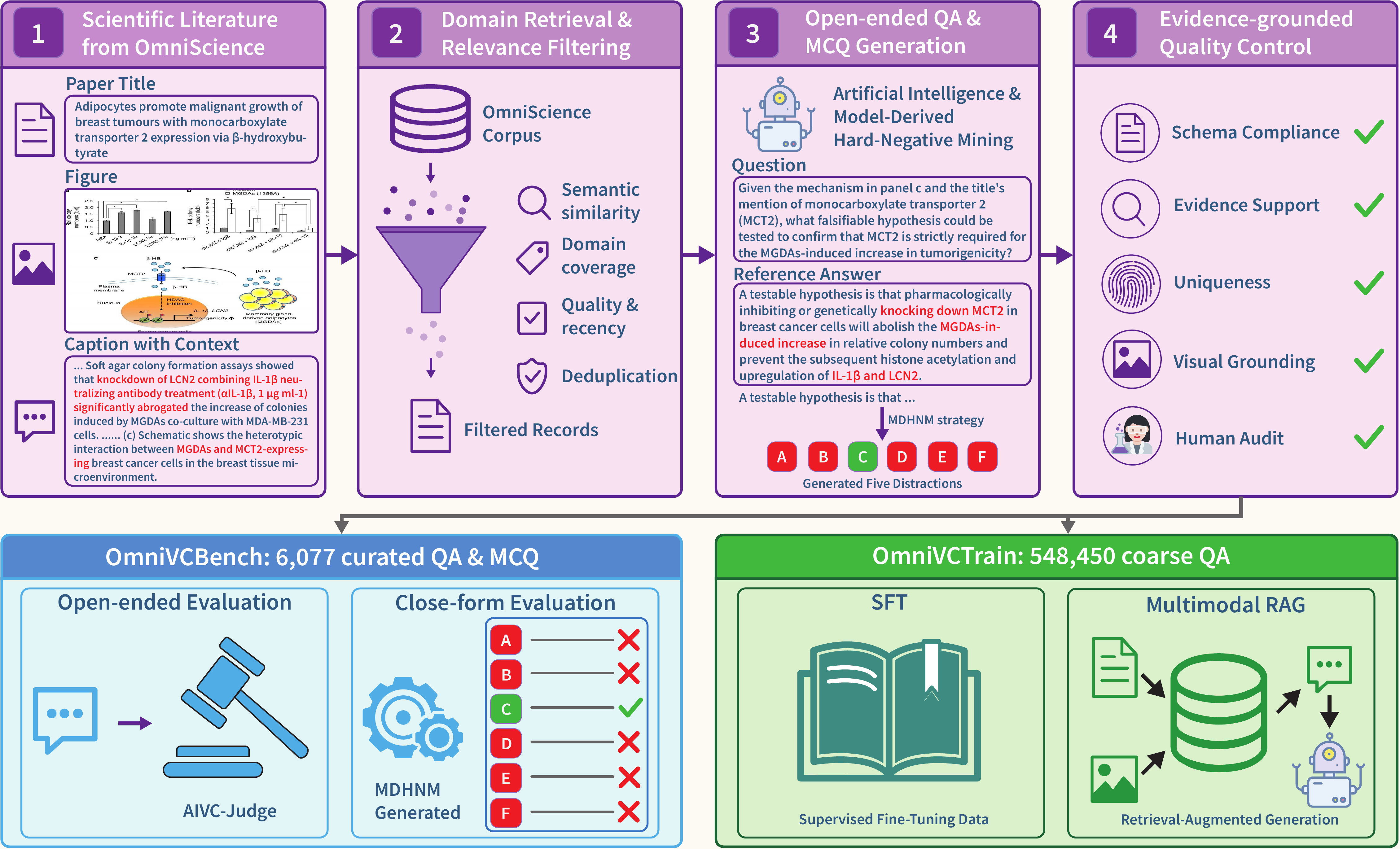}
  \vspace{-3mm}
  \caption{\textbf{Construction and paired evaluation pipeline.} Filtered OmniScience records are expanded into QA pairs and curated; retained items form \textsc{OmniVCBench} (open + MCQ tracks), while coarse QA forms \textsc{OmniVCTrain} for SFT/RAG.}
  \label{fig:construction}
  \vspace{-2mm}
\end{figure}

\subsection{Task Formulation}
\label{sec:task-formulation}

An OmniScience source record is $u=(t,\mathcal{F},C)$: the paper title $t$, scientific images with captions $\mathcal{F}=\{(I_j,c_j)\}_{j=1}^{m}$, and the surrounding paper context $C$, with no questions or answers provided. Our construction pipeline $G$ maps each retained record to a set of benchmark items grounded in its images and experimental context,
\begin{equation}
  G(u)\longrightarrow \{z_i\},
  \qquad z_i=(\mathcal{I}_i,q_i,a_i,\ell_i),
  \qquad \mathcal{I}_i\subseteq\{I_j\}_{j=1}^{m},
  \label{eq:source-to-qa}
\end{equation}
where $\mathcal{I}_i$ contains one or more images, $q_i$ is a newly generated question, $a_i$ is its reference answer, and $\ell_i\in\{1,2,3\}$ is the interpretation level. At inference time, the evaluated model receives only the image set and question; paper titles, captions, and surrounding contexts are retained as source evidence for data construction and reference-aware scoring. We evaluate the same image--question pair through open-response generation and controlled answer selection:
\begin{equation}
  \widehat{r}_i=F_{\theta}(\mathcal{I}_i,q_i),
  \qquad
  \widehat{k}_i=F_{\theta}(\mathcal{I}_i,q_i,\mathcal{O}_i),
  \label{eq:task-formulation}
\end{equation}
where $\widehat{r}_i$ is an open response and $\widehat{k}_i$ selects from a six-option MCQ set $\mathcal{O}_i$. The open protocol preserves explanatory detail and is scored by AIVC-Judge (Section~\ref{sec:aivc-judge}); the MCQ protocol supports inexpensive, deterministic evaluation across models and task levels.

The task levels specify what a response must accomplish. An L1 answer derives a result from the stated conditions and displayed observations; an L2 answer explains the biological process connecting them. An L3 answer proposes an evidence-consistent hypothesis with testable consequences. Published experiments anchor these proposals in traceable observations, and the source interpretation supplies a reference for assessment. A defensible alternative can extend that interpretation while preserving its observational basis. Accordingly, L3 concerns evidence-grounded hypothesis proposal, with novelty and experimental confirmation outside its measured scope. P-E-D describes these functional roles, whereas Bloom supplies construction-time cognitive labels. Our bounded, source-referenced formulation of proposal maps to Bloom's Evaluate (B5)---judging candidate explanations against evidence---rather than unconstrained Create (B6), which remains outside the measured scope. Distinguishing the two axes separates inference from explanation even when both receive an Analyze label. Task definitions and composition statistics appear in Appendices~\ref{app:task-levels} and~\ref{app:dataset-composition}, respectively.

\subsection{\textsc{OmniVCBench} Construction}
\label{sec:omnivcbench-construction}

\subsubsection{Open-ended QA Generation}

\textbf{Source records and domain filtering.}
OmniScience supplies paper-linked figures, titles, author-written captions, and surrounding contexts, which form the source evidence for construction \citep{tao2026omniscience}. We retrieve records using terms covering cellular perturbations, single-cell and spatial omics, pathways, mechanisms, and disease-relevant cell biology (Figure~\ref{fig:construction}). This filtering concentrates the corpus on experiments that support interpretation of cellular behavior and its biological basis. The benchmark sources are Nature Communications papers from 2011--2017, dominated by mouse and human studies, microscopy, and immunoblot assays (Appendix~\ref{app:dataset-composition}).

\textbf{Question and answer synthesis.}
For each retained record, one of two generator models is selected with equal probability per sample: Kimi-K2.6 \citep{moonshotai2026kimik26} or Qwen-VL-MAX \citep{alibabacloud2025qwenvlmax}. The selected generator receives the images and source text and produces candidate question--answer pairs. This two-generator assignment diversifies question phrasing, requested operations, and answer style across the corpus, reducing the stylistic imprint that any single construction model would leave on the benchmark. A record can support several questions when its panels expose distinct scientific operations. The question specifies the required inference, explanation, or hypothesis proposal, while the reference answer remains linked to the source evidence. This separation lets answering models operate on figures and questions, while construction and scoring retain the underlying experimental context as source evidence.

\textbf{Quality control.}
Candidate items undergo schema, evidence, answer-specificity, and visual-grounding checks. Flagged items are rewritten or removed before a manual audit in which three cell biology PhD annotators review each candidate independently---blind to one another's judgments and to the generator model's identity---and only unanimously approved items are retained. The audit retains 6,077 of 7,923 candidates (76.70\%), with unanimous decisions on 93.15\% of candidates and Fleiss' $\kappa=0.8560$ [0.8442, 0.8674] (Appendix~\ref{app:construction-funnel}), yielding 6,077 items from 2,625 source records and 1,080 papers. These checks connect source-level curation to the quality of the questions and options presented during evaluation (Appendix~\ref{app:construction-curation}).

\subsubsection{Model-Derived Hard-Negative Mining for MCQ Generation}

MDHNM constructs MCQ distractors from errors observed during open-response inference, linking options to failures on the same image and question rather than generator-proposed errors. Twelve rollouts per model from a heterogeneous model pool supply candidate responses. We retain those assigned an AIVC-Judge overall score of at most 2 (judged with the image included, as in production scoring), forming an empirical pool of low-scoring responses to the same underlying question:
\begin{equation}
  \mathcal{E}_i=\left\{r_{i,m,t}\;\middle|\;
  r_{i,m,t}=F_m(\mathcal{I}_i,q_i),\;
  J_{\ell_i}(q_i,r_{i,m,t},E_i,\mathcal{I}_i)\leq 2\right\},
  \label{eq:error-pool}
\end{equation}
where $m$ indexes models and $t$ indexes repeated rollouts. Errors shared across models are preferred as recurrent scientific failure patterns in the candidate pool.

The generation model then rewrites candidates from $\mathcal{E}_i$ to match the reference answer's syntax, detail, and length while preserving the underlying error. A reviewer then selects five distractors $\mathcal{D}_i$ using checks on scientific incorrectness, relevance, uniqueness, and answer length:
\begin{equation}
  \begin{aligned}
  |\mathcal{D}_i| &= 5, &
  \mathrm{Wrong}(d\mid z_i)&=1, &
  \mathrm{Grounded}(d\mid\mathcal{I}_i,q_i)&=1,\\
  \mathrm{Unique}(\{a_i\}\cup\mathcal{D}_i)&=1, &
  0.95\leq |d|/|a_i|&\leq1.25,
  \qquad &d&\in\mathcal{D}_i.
  \end{aligned}
  \label{eq:mdhnm-constraints}
\end{equation}
These criteria target incorrectness, relevance, semantic non-overlap, and style consistency; $|\cdot|$ denotes length in characters. The checks address both a distractor's scientific error and surface cues such as wording and length. Finally, $\mathcal{O}_i=\pi(\{a_i\}\cup\mathcal{D}_i)$ randomly orders the reference answer and five distractors into an A--F question, and the unanimous three-annotator audit applies to the resulting items as to all candidates. Appendix~\ref{app:mdhnm-details} reports the screening criteria and comparisons between the complete MDHNM workflow and direct distractor generation.

\subsubsection{\textsc{OmniVCTrain}}

The pipeline also produces \textsc{OmniVCTrain}, a corpus of 548,450 coarse image--question--answer examples for adaptation. These examples retain the same evidence-to-language interface, while benchmark items receive additional manual validation. After upstream paper deduplication, we remove training examples whose paper title or DOI matches a benchmark source. Complementary text and image overlap checks are reported in Appendix~\ref{app:leakage-audit}. We use this resource for supervised fine-tuning on the QA triples and for retrieval of related demonstrations at answer time, with source exclusions retained in both (Appendix~\ref{app:omnivctrain-details}).

\subsection{AIVC-Judge}
\label{sec:aivc-judge}

\begin{figure}[t]
  \vspace{0pt}
  \centering
  \includegraphics[width=0.85\linewidth]{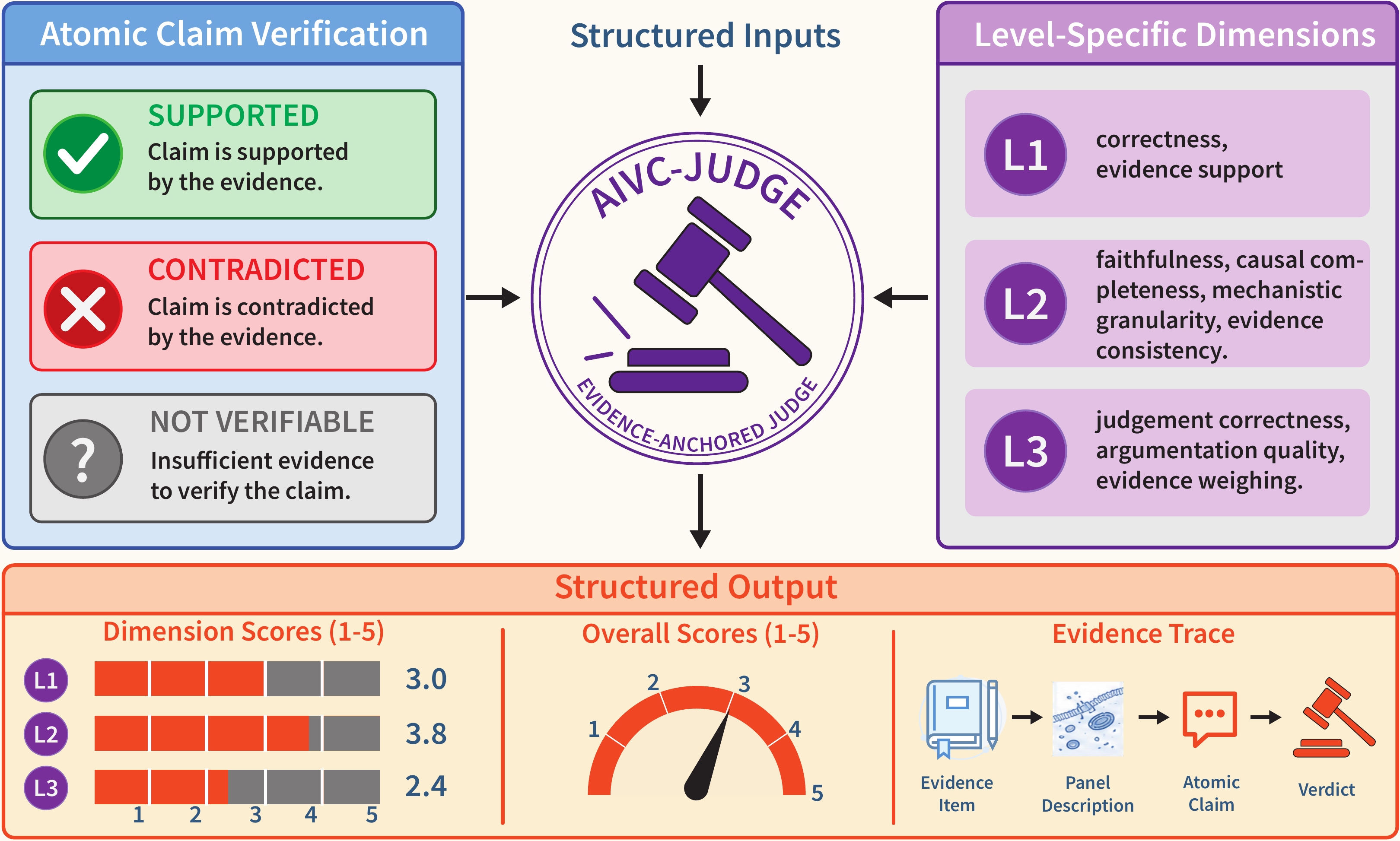}
  \vspace{-4mm}
  \caption{\textbf{AIVC-Judge.} Given the question, response, original figure image, and closed textual evidence, the judge verifies atomic claims against evidence, then scores level-specific dimensions and returns dimension scores, an overall score, and an evidence trace.}
  \label{fig:judge}
  \vspace{-2mm}
\end{figure}

Open responses require semantic evaluation that remains anchored to the source evidence. AIVC-Judge (Figure~\ref{fig:judge}) performs rubric-conditioned pointwise scoring on $(q_i,r_i,E_i,\mathcal{I}_i)$. Here, $E_i$ contains the caption, surrounding context, and reference answer, and $\mathcal{I}_i$ is the original figure image set shown to the answering model. The level-specific judge returns dimension scores, an overall score, and an evidence trace:
\begin{equation}
  J_{\ell_i}(q_i,r_i,E_i,\mathcal{I}_i)
  \longrightarrow (\mathbf{s}_i,o_i,\mathcal{C}_i),
  \label{eq:aivc-judge}
\end{equation}
where $\mathbf{s}_i$ contains level-specific dimension scores, $o_i\in[1,5]$ is the overall score, and $\mathcal{C}_i$ records the claim-level evidence trace. The judge decomposes the response into atomic claims and checks each against the source evidence. It then assigns dimension scores and a holistic overall judgment using the level-specific rubric; the overall score is not programmatically clamped by the dimension caps. This sequence separates the factual support for individual claims from the quality of the response as a whole.

L1 assesses correctness and evidence support, while L2 examines faithfulness, causal completeness, mechanistic granularity, and evidence consistency. L3 scores judgment correctness, argumentation quality, and evidence weighing under a source-referenced critique rubric (Figure~\ref{fig:judge}), assessing critical justification of the proposal rather than proposal formulation itself. Critical factual or directional errors cap the corresponding correctness or faithfulness dimension. The overall score summarizes performance under the selected rubric; dimension scores and evidence traces expose the basis for that assessment. Full checklists and scoring definitions appear in Appendix~\ref{app:judge-details}.

The production judge is DeepSeek-V4-Flash-Vision-Exp \citep{xu2026deepseek}, a multimodal judge that scores each response from the original figure image and the closed textual evidence. It scores pointwise at temperature 0 with candidate identity omitted (Appendix~\ref{app:judge-details}). Human validation scores the same responses under the production rubric on the stratified 300-item sample: mean human overall ratings correlate with the production judge at Spearman $\rho=0.877$ across 884 paired responses; annotators scored without the original figure, so this agreement validates reference-grounded textual rubric application rather than the judge's visual reading (Appendix~\ref{app:human-judge-agreement}). A frozen 24-item L3 pilot rescoring archived responses under an evidence-matched proposal-oriented rubric preserves candidate rankings while improving cross-annotator agreement (QWK 0.459$\to$0.767), indicating that L3 totals are rubric-sensitive, not a pure measure of proposal ability (Appendix~\ref{app:proposal-pilot}).

\begin{table}[t]
\vspace{0pt}
  \caption{Paired results on \textsc{OmniVCBench}: MCQ accuracy (\%, left) and AIVC-Judge scores (1--5, right). Within each block, models without Judge scores precede models sorted by Judge Avg (ascending). $^{\dagger}$ denotes the rank-16, two-epoch SFT configuration on \textsc{OmniVCTrain}. Each statistic summarizes performance over individual QA items in the corresponding track.}
  \label{tab:main-results}
  \centering
  % \scriptsize
  \footnotesize
  \setlength{\tabcolsep}{3.0pt}
  \begin{tabular}{@{}lrrrr@{\hspace{4pt}\vrule width 0.5pt\hspace{4pt}}rrrr@{}}
    \toprule
    & \multicolumn{4}{c}{MCQ accuracy (\%)$\uparrow$} & \multicolumn{4}{c}{AIVC-Judge (1--5)$\uparrow$} \\
    \cmidrule(lr){2-5}\cmidrule(l){6-9}
    Model & L1 & L2 & L3 & Avg & L1 & L2 & L3 & Avg \\
    \midrule
    \multicolumn{9}{@{}l}{\textit{Proprietary models}} \\
    Claude-Sonnet-4-5 \citep{anthropic2025claudesonnet45systemcard} & 45.3 & 52.1 & \textbf{52.5} & 49.4 & 2.55 & 2.92 & 2.62 & 2.68 \\
    \rowcolor{rowgray} Grok-4.6 \citep{xai2026introducinggrok46} & 48.6 & 53.9 & 49.2 & 50.3 & 3.26 & 3.39 & 2.56 & 3.10 \\
    GPT-5.6-luna \citep{openai2025gpt5systemcard} & 47.2 & 48.3 & 33.9 & 43.7 & 3.25 & 3.47 & 2.72 & 3.16 \\
    \rowcolor{rowgray} GPT-5.6-sol \citep{openai2025gpt5systemcard} & \textbf{58.6} & \textbf{60.6} & 44.2 & \textbf{55.1} & \textbf{3.37} & \textbf{3.68} & \textbf{2.76} & \textbf{3.28} \\
    \addlinespace
    \multicolumn{9}{@{}l}{\textit{Open-weight models}} \\
    InternVL3.5-4B \citep{wang2025internvl3} & 30.3 & 31.9 & 23.6 & 28.9 & -- & -- & -- & -- \\
    \rowcolor{rowgray} Qwen2.5-VL-3B \citep{bai2025qwen25vl} & 25.2 & 24.4 & 16.6 & 22.5 & -- & -- & -- & -- \\
    Qwen3-VL-2B \citep{bai2025qwen3vl} & 21.8 & 25.1 & 18.1 & 21.7 & -- & -- & -- & -- \\
    \rowcolor{rowgray} InternVL3.5-2B \citep{wang2025internvl3} & 22.6 & 21.9 & 17.2 & 20.9 & -- & -- & -- & -- \\
    SmolVLM2-2.2B \citep{marafioti2025smolvlm} & 17.9 & 18.7 & 15.1 & 17.3 & -- & -- & -- & -- \\
    \rowcolor{rowgray} LLaVA-Med-Mistral-7B \citep{li2023llavamed} & 17.1 & 17.5 & 12.4 & 15.9 & 2.14 & 1.68 & 1.63 & 1.86 \\
    LLaVA-OneVision-7B \citep{li2024llavaonevision} & 25.7 & 30.5 & 20.4 & 25.5 & 2.32 & 1.83 & 1.70 & 2.00 \\
    \rowcolor{rowgray} Qwen2.5-VL-7B \citep{bai2025qwen25vl} & 28.3 & 30.6 & 20.1 & 26.6 & 2.27 & 2.10 & 1.88 & 2.11 \\
    Qwen3-VL-4B \citep{bai2025qwen3vl} & 31.3 & 33.1 & 26.5 & 30.5 & 2.37 & 2.29 & 1.96 & 2.23 \\
    \rowcolor{rowgray} InternVL3.5-8B \citep{wang2025internvl3} & 30.4 & 34.3 & 26.0 & 30.3 & 2.68 & 2.24 & 1.87 & 2.32 \\
    Qwen3-VL-8B \citep{bai2025qwen3vl} & 34.2 & 37.4 & 29.6 & 33.8 & 2.48 & 2.46 & 1.99 & 2.33 \\
    \rowcolor{rowgray} Qwen3-VL-8B-SFT$^{\dagger}$ \citep{bai2025qwen3vl} & 33.8 & 36.5 & 32.7 & 34.3 & 2.54 & 2.38 & 2.14 & 2.38 \\
    \bottomrule
  \end{tabular}
  \vspace{-2mm}
\end{table}

\section{Experiments}
\label{sec:experiments}

\subsection{Setup}
\label{sec:setup}

\textbf{Models and protocols.} We evaluate four proprietary and eleven open-weight MLLMs, with the latter spanning 2B--8B parameters. The MCQ track includes these fifteen base models and an SFT variant; eleven configurations also have open-response evaluation, with the SFT variant represented by the rank-16 checkpoint. MCQ accuracy uses the original six-option keys, and open-response scores average the judge's raw overall output. Model identifiers and clustered statistics appear in Appendices~\ref{app:model-versions} and~\ref{app:cluster-stats}.

\textbf{Adaptation settings.} We adapt Qwen3-VL-8B through LoRA supervised fine-tuning in LLaMA-Factory \citep{hu2021lora,zheng2024llamafactory} and multimodal retrieval-augmented generation \citep{lewis2020retrieval}. Both use \textsc{OmniVCTrain}: SFT supplies training examples, and RAG retrieves related QA demonstrations through joint image--question embeddings. Training and retrieval configurations, including source-exclusion rules, are detailed in Appendix~\ref{app:omnivctrain-details}.

% \begin{figure}[!b]

\subsection{Benchmarking MLLMs with \textsc{OmniVCBench}}
\label{sec:main-results}

\textbf{RQ1: Can MLLMs judge reliably from experimental observations?} Table~\ref{tab:main-results} reports paired MCQ and open-response results. GPT-5.6-sol leads both tracks, reaching 55.1\% MCQ accuracy and 3.28/5 under AIVC-Judge. Open-weight base models span 15.9\%--33.8\% MCQ accuracy and score below 2.40 on open responses. For the strongest model, the open-response score declines from 3.68 on L2 to 2.76 on L3 (the decline concentrates in unprompted critique dimensions; in the frozen 24-item pilot of Appendix~\ref{app:proposal-pilot}, the same responses score 4.75--4.83).

\begin{figure}[t]
  \centering
  \includegraphics[width=0.88\linewidth]{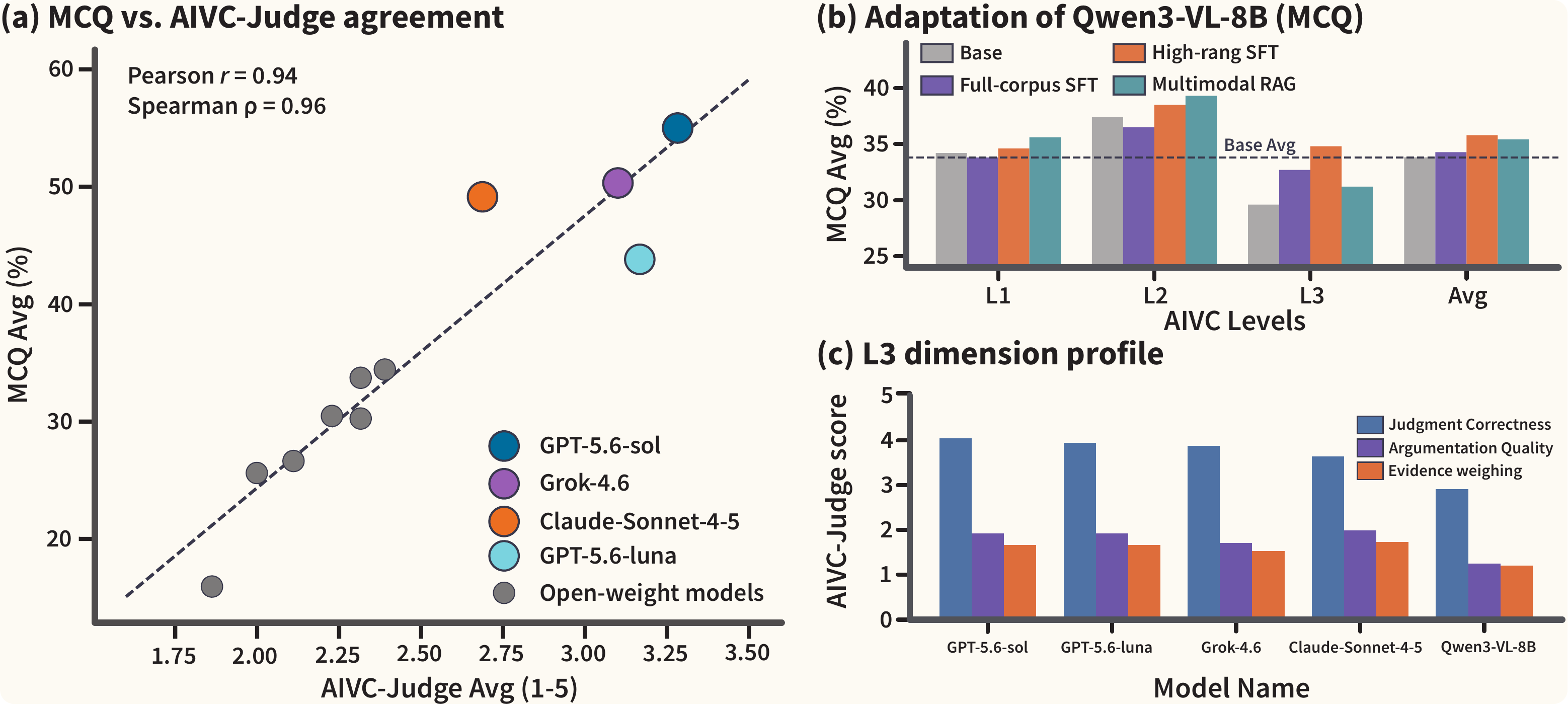}
  \vspace{-4mm}
  \caption{\textbf{Evaluation agreement and adaptation gains.} (a) MCQ accuracy vs.\ AIVC-Judge score over eleven models. (b) Qwen3-VL-8B adaptation across L1--L3. (c) L3 dimension profile.}
  \label{fig:results}
  \vspace{-2mm}
\end{figure}

\textbf{RQ2: How far do MCQs reflect open-response performance?} Across the eleven shared models, MCQ and open-response scores correlate at Spearman $\rho=0.964$ (Figure~\ref{fig:results}a). The pooled relationship includes strong separation between proprietary and open-weight systems; within the four proprietary models the correlation is much weaker (Pearson $r=0.208$), with individual rank inversions (Appendix~\ref{app:cluster-stats}). The two formats capture related but distinct aspects of performance on the same questions.

\textbf{Human reference and visual evidence.} On the stratified 300-item sample, human MCQ accuracy ranges from 55.3\% to 74.7\% across three annotators, bracketing GPT-5.6-sol's 57.3\% with images (Appendix~\ref{app:human-study}). In a no-image comparison on the same sample, GPT-5.6-sol accuracy falls to 34.7\% without the figure---still twice the 16.7\% random baseline---demonstrating a substantial visual contribution alongside question text, option cues, and domain priors; two open-weight models show no drop at all (Appendix~\ref{app:item-audits}).

\subsection{Adaptation with \textsc{OmniVCTrain}}
\label{sec:adaptation}

\textbf{RQ3: Which capabilities can added training data improve?} Table~\ref{tab:adaptation-main} compares adaptation of Qwen3-VL-8B on the MCQ track. The rank-32 SFT configuration (100k subset) improves average accuracy by 1.97 points, its largest gain at L3 (29.6\% to 34.8\%); multimodal RAG improves it by 1.60 points, and the lower-rank configuration changes it by +0.46. Both gains are significant under source-paper-clustered Holm-corrected analysis (Appendix~\ref{app:omnivctrain-details}); Figure~\ref{fig:results}b shows their distinct level distributions. The coarse QA corpus thus supports both adaptation routes, with gains depending on how it is used.

Joint image--question retrieval scores 35.41\% versus 33.62\% image-only and 33.40\% text-only, so the gain occurs with the joint representation. Across base models, RAG improvements range from 0.76 to 1.60 points (Appendix~\ref{app:rag-details}); the strongest adapted Qwen configuration reaches 35.79\%, leaving a substantial gap to the leading proprietary model.

\begin{table}[!t]
\vspace{0pt}
  \caption{Adaptation of Qwen3-VL-8B on the MCQ track. $\Delta$ denotes the change from the base model; $p$ is the $p$-value from the raw paper-block permutation test. Corrected comparisons appear in Appendix~\ref{app:omnivctrain-details}.}
  \label{tab:adaptation-main}
  % \vspace{0pt}
  \centering
  \scriptsize
  \setlength{\tabcolsep}{3.6pt}
  \begin{tabular}{@{}llrrrrrr@{}}
    \toprule
    Method & Resource & L1 & L2 & L3 & Avg & $\Delta$ & $p$ \\
    \midrule
    Base & -- & 34.2 & 37.4 & 29.6 & 33.82 & -- & -- \\
    \rowcolor{rowgray} SFT, full corpus & 548k, 2 epochs & 33.8 & 36.5 & 32.7 & 34.28 & +0.46 & 0.42 \\
    SFT, rank 32 & 100k, 2 epochs & 34.6 & 38.5 & \textbf{34.8} & \textbf{35.79} & \textbf{+1.97} & $3.2\!\times\!10^{-4}$ \\
    \rowcolor{rowgray} Multimodal RAG & 548k index, $k=3$ & \textbf{35.6} & \textbf{39.3} & 31.2 & 35.41 & +1.60 & $5.6\!\times\!10^{-4}$ \\
    \bottomrule
  \end{tabular}
  \vspace{0pt}
\end{table}

\subsection{Error analysis on \textsc{OmniVCBench}}
\label{sec:error-analysis}

\textbf{RQ4: Where do explanation and hypothesis proposal fall short?} The selected cases expose response-level differences behind the aggregate scores (Appendix~\ref{app:case-studies}): reversed effect directions and incorrect associations between interventions, panels, and biological entities at L1/L2. Paired evaluation exposes these inconsistencies: in the IFNAR1 and CCP1 examples, a keyed MCQ selection coexists with an inconsistent open explanation; conversely, correct open relations can accompany an option that changes direction or effect. Reading both responses therefore reveals errors that an MCQ score alone can conceal.

L3 additionally requires separating the motivating observation from the intervention and outcome a hypothesis predicts; a proposal can be testable yet rest on a misread observation (the hair-regeneration example). A model can also produce a defensible open hypothesis while selecting an option that reverses its own prediction or contradicts the pathway diagram (Cases~H and~I, Appendix~\ref{app:case-studies}). The cases connect this distinction to the benchmark's central purpose: assessing the relation between evidence and a scientific answer.

\subsection{Evidence Acquisition in the AIVC Refinement Loop}
\label{sec:refinement-loop}

The benchmark evaluates this stage on curated literature figures; to test the same component inside executable loops, we run four GPT-5.6-sol pilots with one shared structure: predict a hidden biological output, select one additional measurement, observe it, and revise for rescoring. The settings span four real evidence types: non-additive double-gene responses from GEARS on Norman CRISPRa data \citep{norman2019exploring,roohani2023predicting}, drug-combination responses from a locally trained CPA with combinations held out \citep{lotfollahi2023predicting}, mechanism classes from real fluorescence microscopy \citep{caie2010high,ljosa2012annotated}, and signaling interventions on Sachs data \citep{sachs2005causal}. Initial accuracies range from 66.7\% to 93.3\%, so no pilot sits at ceiling. Acquiring the selected measurement yields nine corrections and zero regressions across the three numerical pilots (77.1\%$\to$89.6\%, 93.3\%$\to$100.0\%, and 66.7\%$\to$70.0\%), while self-review improves no pilot and degrades two, and real-microscopy classification shows no gain. These are one-step replay loops without simulator retraining and with corrections confined to queried readouts, reported as revision pilots rather than a validated discovery loop (Appendix~\ref{app:refinement-loop}).

\section{Conclusion}
\label{sec:conclusion}

AIVC benchmarks primarily assess cellular simulation, while the AIVC agenda also requires interpreting evidence and proposing hypotheses. We introduce \textbf{\textsc{OmniVCBench}}, operationalizing this component through 6,077 source-traceable questions spanning inference, explanation, and hypothesis proposal. AIVC-Judge and model-derived distractors provide paired assessments of generated explanations and answer selection, with strong ranking agreement between tracks and distinct evidence-handling behaviors in case studies. The strongest model scores below three of five on L3 open responses, indicating weaker evidence-grounded argumentation. Human validation supports the shared-evidence scoring protocol, and \textsc{OmniVCTrain} adaptation yields modest gains. Returning to our opening question: four one-step evidence-acquisition pilots on real biological outputs position the MLLM as the interpretation component that closes the loop between simulation and experimental evidence, a role current models fill only partially. With the released judge implementation, rubrics, and data demo, these resources support developing interpretation components that connect predictions with evidence-grounded explanations and testable hypotheses for the next loop iteration of AIVC development and beyond.

\section*{AI Use Statement}
In this work, we used generative AI tools to generate synthetic data: candidate question--answer pairs, MCQ distractors, and auxiliary task-type annotations in \textsc{OmniVCBench} and \textsc{OmniVCTrain} were produced by multimodal LLMs, and every retained item was approved by human annotators. We also used generative AI tools to interpret results and support qualitative data analysis, through the MLLM-as-a-judge pipeline that verifies candidate responses against source evidence. We have not used generative AI tools to develop theoretical models or conceptual frameworks, formulate mathematical claims or assist in proofs, design the research methodology, or implement methods, and the remaining required-disclosure tasks are not applicable to this work. We take responsibility for the final content of this work, including text, claims, and artifacts produced with the aid of generative AI.

% \clearpage
\bibliography{iclr2027_conference}
\bibliographystyle{iclr2027_conference}
\clearpage

\appendix

\clearpage

\begingroup
\hypersetup{linkbordercolor={1 1 1}}

\begin{center}
{\LARGE\bfseries  Appendix Contents}\\[3pt]
{\color{omniink}\rule{0.8\linewidth}{1pt}}
\end{center}

\vspace{3mm}

\noindent
\begin{minipage}{\linewidth}
\setlength{\parindent}{0pt}
\setlength{\parskip}{4pt}
\color{omniink}
{\small\bfseries\color{omniink}\hyperref[app:construction-details]{A\quad Construction and Evaluation Details}}\dotfill{\small\pageref{app:construction-details}}\par
\smallskip
{\footnotesize
\hspace*{1em}\hangindent=2.8em \hyperref[app:model-versions]{A.1\ \ Model Versions and Access Routes}\dotfill\pageref{app:model-versions}\par
\hspace*{1em}\hangindent=2.8em \hyperref[app:task-levels]{A.2\ \ Interpretation-Layer Task Definitions}\dotfill\pageref{app:task-levels}\par
\hspace*{1em}\hangindent=2.8em \hyperref[app:construction-funnel]{A.3\ \ Construction Funnel}\dotfill\pageref{app:construction-funnel}\par
\hspace*{1em}\hangindent=2.8em \hyperref[app:construction-curation]{A.4\ \ Source Filtering and QA Curation}\dotfill\pageref{app:construction-curation}\par
\hspace*{1em}\hangindent=2.8em \hyperref[app:dataset-composition]{A.5\ \ Dataset Composition and Difficulty Distribution}\dotfill\pageref{app:dataset-composition}\par
\hspace*{1em}\hangindent=2.8em \hyperref[app:item-audits]{A.6\ \ Item and Option Audits}\dotfill\pageref{app:item-audits}\par
\hspace*{1em}\hangindent=2.8em \hyperref[app:mdhnm-details]{A.7\ \ MDHNM Screening Criteria}\dotfill\pageref{app:mdhnm-details}\par
\hspace*{1em}\hangindent=2.8em \hyperref[app:judge-details]{A.8\ \ AIVC-Judge Rubrics and Production Configuration}\dotfill\pageref{app:judge-details}\par
\hspace*{1em}\hangindent=2.8em \hyperref[app:human-study]{A.9\ \ Human Validation Sample and MCQ Performance}\dotfill\pageref{app:human-study}\par
\hspace*{1em}\hangindent=2.8em \hyperref[app:human-judge-agreement]{A.10\ \ Human Scoring of Open-Ended Responses}\dotfill\pageref{app:human-judge-agreement}\par
\hspace*{1em}\hangindent=2.8em \hyperref[app:proposal-pilot]{A.11\ \ Proposal-Oriented Rubric Sensitivity Pilot}\dotfill\pageref{app:proposal-pilot}\par
\hspace*{1em}\hangindent=2.8em \hyperref[app:leakage-audit]{A.12\ \ Release-Level Train--Benchmark Overlap Audit}\dotfill\pageref{app:leakage-audit}\par
\hspace*{1em}\hangindent=2.8em \hyperref[app:cluster-stats]{A.13\ \ Cluster-Robust Scores and Paper-Weighted Statistics}\dotfill\pageref{app:cluster-stats}\par
\hspace*{1em}\hangindent=2.8em \hyperref[app:run-settings]{A.14\ \ Prompts and Run Settings}\dotfill\pageref{app:run-settings}\par
}
\medskip
{\small\bfseries\color{omniink}\hyperref[app:omnivctrain-details]{B\quad \textsc{OmniVCTrain} Adaptation Details}}\dotfill{\small\pageref{app:omnivctrain-details}}\par
\smallskip
{\footnotesize
\hspace*{1em}\hangindent=2.8em \hyperref[app:sft-details]{B.1\ \ Supervised Fine-Tuning}\dotfill\pageref{app:sft-details}\par
\hspace*{1em}\hangindent=2.8em \hyperref[app:rag-details]{B.2\ \ Retrieval-Augmented Generation}\dotfill\pageref{app:rag-details}\par
}
\medskip
{\small\bfseries\color{omniink}\hyperref[app:refinement-loop]{C\quad Biological Evidence Acquisition and Prediction Revision}}\dotfill{\small\pageref{app:refinement-loop}}\par
\smallskip
{\footnotesize
\hspace*{1em}\hangindent=2.8em \hyperref[app:loop-protocol]{C.1\ \ Task Structure and Evidence Settings}\dotfill\pageref{app:loop-protocol}\par
\hspace*{1em}\hangindent=2.8em \hyperref[app:loop-results]{C.2\ \ Results}\dotfill\pageref{app:loop-results}\par
\hspace*{1em}\hangindent=2.8em \hyperref[app:loop-scope]{C.3\ \ Scope of the Pilots}\dotfill\pageref{app:loop-scope}\par
}
\medskip
{\small\bfseries\color{omniink}\hyperref[app:judge-dimensions]{D\quad AIVC-Judge Dimension-Wise Results}}\dotfill{\small\pageref{app:judge-dimensions}}\par
\medskip
{\small\bfseries\color{omniink}\hyperref[app:limitations]{E\quad Limitations and Outlook}}\dotfill{\small\pageref{app:limitations}}\par
\medskip
{\small\bfseries\color{omniink}\hyperref[app:case-studies]{F\quad Case Studies of Reasoning and Scoring}}\dotfill{\small\pageref{app:case-studies}}\par
\smallskip
{\footnotesize
\hspace*{1em}\hangindent=2.8em \hyperref[app:multi-subfigure]{Single- versus multi-subfigure difficulty}\dotfill\pageref{app:multi-subfigure}\par
\hspace*{1em}\hangindent=2.8em \hyperref[app:case-a]{Case A: LGALS3BP readout polarity (L1)}\dotfill\pageref{app:case-a}\par
\hspace*{1em}\hangindent=2.8em \hyperref[app:case-b]{Case B: IFNAR1 counterfactual across formats (L1)}\dotfill\pageref{app:case-b}\par
\hspace*{1em}\hangindent=2.8em \hyperref[app:case-c]{Case C: C8orf4 effect direction across models (L1)}\dotfill\pageref{app:case-c}\par
\hspace*{1em}\hangindent=2.8em \hyperref[app:case-d]{Case D: ROS--TGFBRI intervention--reversal chain (L2)}\dotfill\pageref{app:case-d}\par
\hspace*{1em}\hangindent=2.8em \hyperref[app:case-e]{Case E: CCP1 sign from reciprocal perturbations (L2)}\dotfill\pageref{app:case-e}\par
\hspace*{1em}\hangindent=2.8em \hyperref[app:case-f]{Case F: BRCA1 spatial integration (L2)}\dotfill\pageref{app:case-f}\par
\hspace*{1em}\hangindent=2.8em \hyperref[app:case-g]{Case G: hair-regeneration hypothesis on a faulty premise (L3)}\dotfill\pageref{app:case-g}\par
\hspace*{1em}\hangindent=2.8em \hyperref[app:case-h]{Case H: YAP--Ect2/Fgd3 paired-track divergence (L3)}\dotfill\pageref{app:case-h}\par
\hspace*{1em}\hangindent=2.8em \hyperref[app:case-i]{Case I: integrin--Lck paired-track divergence (L3)}\dotfill\pageref{app:case-i}\par
}
\medskip
{\small\bfseries\color{omniink}\hyperref[app:ethics]{G\quad Ethics, License, and Data Availability}}\dotfill{\small\pageref{app:ethics}}\par
\end{minipage}

\endgroup

\clearpage

\section{Construction and Evaluation Details}
\label{app:construction-details}

\subsection{Model Versions and Access Routes}
\label{app:model-versions}

Table~\ref{tab:model-versions} lists the exact identifiers, versions, and access routes of the evaluated systems, the construction models, the production judge, the auxiliary annotation and distractor models, and the retrieval embedder.

\begin{table}[!ht]
  \caption{Model identifiers, versions, and access routes. API models are accessed through the listed provider; open-weight checkpoints are run locally from HuggingFace.}
  \label{tab:model-versions}
  \centering
  \scriptsize
  \setlength{\tabcolsep}{3pt}
  \begin{tabularx}{\linewidth}{@{}>{\raggedright\arraybackslash}p{0.24\linewidth}>{\raggedright\arraybackslash}p{0.30\linewidth}>{\raggedright\arraybackslash}X@{}}
    \toprule
    Model & Model ID / version & Source and role \\
    \midrule
    GPT-5.6-sol & \texttt{gpt-5.6-sol} & OpenRouter API; main benchmark \\
    \rowcolor{rowgray} GPT-5.6-luna & \texttt{gpt-5.6-luna} & OpenRouter API; main benchmark \\
    Grok-4.6 & \texttt{grok-4.6} & OpenRouter API; main benchmark \\
    \rowcolor{rowgray} Claude-Sonnet-4-5 & \texttt{claude-sonnet-4-5} & OpenRouter API; main benchmark \\
    DeepSeek-V4-Flash-Vision-Exp & \texttt{deepseek-v4-flash-vision-exp} & Official DeepSeek API; production AIVC-Judge with image input, temperature 0 \\
    \rowcolor{rowgray} GPT-5.6-terra & \texttt{gpt-5.6-terra} & OpenRouter API; auxiliary task-type annotation and distractor generation \\
    Gemini-3.7-flash-high & \texttt{gemini-3.7-flash-high} & OpenRouter API; auxiliary task-type annotation and distractor generation \\
    \rowcolor{rowgray} Gemini-3.1-flash-lite & \texttt{gemini-3.1-flash-lite} & OpenRouter API; direct and mined distractor ablation generator \\
    Kimi-K2.6 & \texttt{Kimi-K2.6} & Official Moonshot AI API; QA generation, evidence transcription, and MDHNM distractor rewriting \\
    \rowcolor{rowgray} Qwen-VL-MAX & \texttt{qwen-vl-max} & Alibaba Cloud Bailian API; QA generation \\
    glm-5v-turbo & \texttt{glm-5v-turbo} & OpenRouter API; MDHNM error-pool generation \\
    \rowcolor{rowgray} minimax-m3 & \texttt{minimax-m3} & OpenRouter API; MDHNM error-pool generation \\
    HuatuoGPT-Vision-7B & \texttt{HuatuoGPT-Vision-7B} & HuggingFace checkpoint; MDHNM error-pool generation \\
    \rowcolor{rowgray}InternVL3.5-4B & \texttt{InternVL3.5-4B} & HuggingFace checkpoint; local inference \\
    Qwen2.5-VL-3B & \texttt{Qwen2.5-VL-3B} & HuggingFace checkpoint; local inference \\
    \rowcolor{rowgray}Qwen3-VL-2B & \texttt{Qwen3-VL-2B} & HuggingFace checkpoint; local inference \\
    InternVL3.5-2B & \texttt{InternVL3.5-2B} & HuggingFace checkpoint; local inference \\
    \rowcolor{rowgray}SmolVLM2-2.2B & \texttt{SmolVLM2-2.2B} & HuggingFace checkpoint; local inference \\
    LLaVA-Med-Mistral-7B & \texttt{LLaVA-Med-Mistral-7B} & HuggingFace checkpoint; local inference \\
    \rowcolor{rowgray}LLaVA-OneVision-7B & \texttt{LLaVA-OneVision-7B} & HuggingFace checkpoint; local inference \\
    Qwen2.5-VL-7B & \texttt{Qwen2.5-VL-7B} & HuggingFace checkpoint; local inference \\
    \rowcolor{rowgray}Qwen3-VL-4B & \texttt{Qwen3-VL-4B} & HuggingFace checkpoint; local inference \\
    InternVL3.5-8B & \texttt{InternVL3.5-8B} & HuggingFace checkpoint; local inference \\
    \rowcolor{rowgray}Qwen3-VL-8B & \texttt{Qwen3-VL-8B} & HuggingFace checkpoint; local inference \\
    Qwen3-VL-8B-SFT & \texttt{Qwen3-VL-8B} + LoRA & Rank 16, two epochs; local inference; rank-32 comparison in Table~\ref{tab:adaptation-main} \\
    \rowcolor{rowgray} Qwen3-VL-Embedding-8B & \texttt{Qwen3-VL-Embedding-8B} & HuggingFace checkpoint; local inference; multimodal RAG index construction \\
    \bottomrule
  \end{tabularx}
\end{table}

\subsection{Interpretation-Layer Task Definitions}
\label{app:task-levels}

Table~\ref{tab:task-levels} reports the operational definitions of the three interpretation levels; P-E-D denotes functional alignment, not label equivalence.

\begin{table}[!ht]
  \caption{Interpretation-layer task definitions. P-E-D denotes functional alignment, not label equivalence.}
  \label{tab:task-levels}
  \centering
  \scriptsize
  \begin{tabularx}{0.95\linewidth}{@{}lXllr@{}}
    \toprule
    Level & Required operation & P-E-D alignment & Bloom levels & $n$ \\
    \midrule
    L1 & Infer a value, direction, relation, or outcome from displayed evidence & Predict analogue & 2, 3, 4 & 2,576 \\
    L2 & Explain an observed result through a biologically grounded mechanism & Explain & 4 & 1,763 \\
    L3 & Propose a testable hypothesis grounded in the displayed evidence & Discover prerequisite & 5 & 1,738 \\
    \bottomrule
  \end{tabularx}
\end{table}

\textbf{L3 task-type audit.} To verify that released L3 items match the hypothesis-proposal definition, two independent annotator models (Gemini-3.7-flash-high and GPT-5.6-terra) each labeled all 1,738 L3 items by task type (Table~\ref{tab:l3-tasktype}). The two labelers agree closely (observed agreement 0.9937; Cohen's $\kappa=0.9068$) and both assign over 96\% of items to \emph{propose\_novel}; \emph{design\_experiment}, \emph{other}, and \emph{evaluate\_given} are rare. The audit identifies hypothesis proposal as the dominant requested operation within L3. It concerns task type, while Table~\ref{tab:level-cross} records the construction-time Bloom labels. Level labels are construction-time assignments, and boundary items exist: Case F (Appendix~\ref{app:case-f}) requests a spatial-relation comparison that sits close to the L1 definition although labeled L2.

\begin{table}[!ht]
  \caption{Task-type audit of all 1,738 L3 items by two independent annotator models (counts). Observed agreement 0.9937; Cohen's $\kappa=0.9068$.}
  \label{tab:l3-tasktype}
  \centering
  \small
  \setlength{\tabcolsep}{4pt}
  \begin{tabular}{@{}lrr@{}}
    \toprule
    Task type & Gemini-3.7-flash-high & GPT-5.6-terra \\
    \midrule
    propose\_novel & 1,679 & 1,676 \\
    \rowcolor{rowgray} design\_experiment & 31 & 35 \\
    other & 21 & 20 \\
    \rowcolor{rowgray} evaluate\_given & 7 & 7 \\
    \bottomrule
  \end{tabular}
\end{table}

\subsection{Construction Funnel}
\label{app:construction-funnel}

Table~\ref{tab:construction-funnel} reports the retained counts at the principal construction stages. The final audit stage was performed by three cell biology PhD annotators who reviewed candidates independently, blind to one another's judgments and to the generator model's identity; labels were merged only after all three completed their review, and a candidate item entered the benchmark only when all three approved it. The table intentionally reports only the stage totals and does not treat records and QA items as interchangeable units.

\begin{table}[!ht]
  \caption{Retained counts in the \textsc{OmniScience}--\textsc{OmniVCBench} construction funnel.}
  \label{tab:construction-funnel}
  \centering
  \small
  \begin{tabular}{@{}lr@{}}
    \toprule
    Stage & Retained count \\
    \midrule
    OmniScience initial figure--caption pairs & 1,525,179 \\
    Stage 0 keyword filtering & 98,632 \\
    QA generation + collaborative filtering & 7,923 \\
    Unanimous three-annotator audit & 6,077 \\
    \bottomrule
  \end{tabular}
\end{table}

The audit decided 7,923 candidate items: 6,077 were accepted by all three annotators and retained, 1,303 were rejected by all three, and 543 carried disagreement and were not retained. The retention rate is 76.70\%, and the three annotators fully agreed on 93.15\% of candidates. Pairwise Cohen's $\kappa$ is 0.8545 [0.8397, 0.8687] (annotators 1--2), 0.8506 [0.8357, 0.8650] (annotators 1--3), and 0.8630 [0.8487, 0.8769] (annotators 2--3); overall Fleiss' $\kappa$ is 0.8560 [0.8442, 0.8674], with bracketed values denoting 95\% confidence intervals.

\subsection{Source Filtering and QA Curation}
\label{app:construction-curation}

All captions and surrounding contexts are the author-written text of the source papers, with no model rewriting in the OmniScience corpus \citep{tao2026omniscience}; these captions are typically self-contained, describing both the visual content and the experimental context of their figures. The initial retrieval vocabulary covers six AIVC-relevant themes: virtual cells and cellular digital twins; perturbation responses; single-cell and spatial omics; multi-omics integration; mechanisms and pathways; and disease-relevant cell biology. Representative terms are shown in Table~\ref{tab:filtering-terms}.

\begin{table}[!ht]
  \caption{Representative terms used to retrieve AIVC-relevant source records from OmniScience.}
  \label{tab:filtering-terms}
  \centering
  \small
  \begin{tabularx}{\linewidth}{@{}lX@{}}
    \toprule
    Theme & Representative terms \\
    \midrule
    Virtual cells & virtual cell; AI virtual cell; digital twin cell; whole-cell model \\
    Perturbations & perturbation; CRISPR screen; knockout; drug or dose response \\
    Single-cell/spatial & single-cell RNA-seq; spatial transcriptomics; UMAP; RNA velocity \\
    Multi-omics & multi-omics; ATAC-seq; proteomics; chromatin accessibility \\
    Mechanisms & signalling pathway; gene-regulatory network; causal mechanism \\
    Disease biology & cancer; drug resistance; therapeutic hypothesis; patient-derived model \\
    \bottomrule
  \end{tabularx}
\end{table}

Candidate QA pairs undergo schema, source-evidence, answer-specificity, and visual-grounding checks. Schema checks verify fields, panel references, and answer length. Source-evidence checks compare reference answers with captions and context. Answer checks assess whether the question specifies its requested inference or explanation. L3 references anchor hypothesis assessment to source observations, while hypothesis validity depends on evidence consistency and testable consequences. Visual checks screen for text-only shortcuts. Flagged candidates are rewritten or removed before the unanimous three-annotator audit (Appendix~\ref{app:construction-funnel}). Option-level uniqueness is assessed separately for the six-option track (Appendix~\ref{app:mdhnm-details}).

Table~\ref{tab:level-cross} (left) reports the empirical relation between the final interpretation levels and P-E-D labels. The small off-diagonal counts show why we treat P-E-D as a functional alignment rather than an identity: cognitive depth and answer function provide distinct information.

\begin{table}[!htb]
  \caption{Interpretation level by P-E-D label (left) and by Bloom level (right; item counts) in the final benchmark. Parentheses give the within-level percentage of the predominant P-E-D label.}
  \label{tab:level-cross}
  \centering
  \small
  \setlength{\tabcolsep}{4pt}
  \begin{tabular}{@{}lrrrr@{\hspace{10pt}}rrrrr@{}}
    \toprule
    \multicolumn{5}{c}{P-E-D label} & \multicolumn{5}{c}{Bloom level} \\
    \cmidrule(lr){1-5}\cmidrule(l){6-10}
    Level & Predict & Explain & Discover & Total & B2 & B3 & B4 & B5 & Total \\
    \midrule
    L1 & 2,362 (91.7\%) & 214 & 0 & 2,576 & 261 & 773 & 1,542 & 0 & 2,576 \\
    \rowcolor{rowgray} L2 & 0 & 1,763 (100\%) & 0 & 1,763 & 0 & 0 & 1,763 & 0 & 1,763 \\
    L3 & 9 & 9 & 1,720 (99.0\%) & 1,738 & 0 & 0 & 0 & 1,738 & 1,738 \\
    \bottomrule
  \end{tabular}
\end{table}

\subsection{Dataset Composition and Difficulty Distribution}
\label{app:dataset-composition}

An \emph{item} is one question--answer pair, a \emph{source record} is an OmniScience figure record, and a \emph{paper} is a source article. An \emph{image path} identifies a materialized panel crop or full figure. A source record can yield multiple images and questions, so these units are counted separately. Tables~\ref{tab:level-cross} and~\ref{tab:bloom-ped-dist} report the construction-label distribution. Retained items use Bloom labels B2--B5, with 83.0\% assigned to Analyze or Evaluate. These labels describe the dataset's construction taxonomy. Within B4, the task-role labels distinguish 1,542 inference items from 1,763 mechanistic-explanation items. L3 items carry B5 labels and request evidence-grounded hypotheses; these construction labels are distinct from a validated measurement of cognitive difficulty.

\begin{table}[!htb]
  \caption{Distribution over Bloom's taxonomy levels (left) and P-E-D functions (right) in the final benchmark.}
  \label{tab:bloom-ped-dist}
  \centering
  \scriptsize
  \begin{tabular}{@{}clrr@{\hspace{10pt}}llrr@{}}
    \toprule
    \multicolumn{4}{c}{Bloom's taxonomy} & \multicolumn{4}{c}{P-E-D function} \\
    \cmidrule(lr){1-4}\cmidrule(l){5-8}
    Level & Category & $n$ & Share & Function & Operational meaning & $n$ & Share \\
    \midrule
    2 & Understand & 261 & 4.3\% & Predict & Evidence-conditioned inference & 2,371 & 39.0\% \\
    3 & Apply & 773 & 12.7\% & Explain & Mechanistic explanation & 1,986 & 32.7\% \\
    4 & Analyze & 3,305 & 54.4\% & Discover & Testable-hypothesis proposal grounded in displayed evidence & 1,720 & 28.3\% \\
    5 & Evaluate & 1,738 & 28.6\% & & & & \\
    \bottomrule
  \end{tabular}
\end{table}

\textbf{Corpus units and provenance.} The released corpora derive from 1,525,179 OmniScience source records. Keyword filtering retains 98,632 records; 96,007 of them, spanning 33,802 papers, expand into 548,450 \textsc{OmniVCTrain} examples (327,601 materialized image paths), while 2,625 records from 1,080 papers yield the 6,077 benchmark items (4,428 image paths). Source coverage is therefore 6.29\% and 0.17\% of the upstream corpus, respectively. Here \emph{multi-subfigure} items integrate evidence across subfigures (panels) of a single composite figure: each item is materialized as one image (the whole figure or a panel crop), 50.9\% of items are built over multi-panel figures whose evidence spans several panels, and items reference a mean of 1.78 panels. Table~\ref{tab:corpus-units} reports unit-level statistics.

\begin{table}[!ht]
  \caption{Unit-level statistics of the released corpora.}
  \label{tab:corpus-units}
  \centering
  \small
  \setlength{\tabcolsep}{4pt}
  \begin{tabular}{@{}lrr@{}}
    \toprule
    Unit & \textsc{OmniVCTrain} & \textsc{OmniVCBench} \\
    \midrule
    Released QA rows & 548,450 & 6,077 \\
    \rowcolor{rowgray} Unique source records & 96,007 & 2,625 \\
    Unique papers & 33,802 & 1,080 \\
    \rowcolor{rowgray} Materialized image paths & 327,601 & 4,428 \\
    QA per source record (mean / median) & 5.71 / 6 & 2.32 / 2 \\
    \rowcolor{rowgray} QA per paper (mean / median) & 16.23 / 12 & 5.63 / 4 \\
    \bottomrule
  \end{tabular}
\end{table}

\textbf{Subject, source, and length distribution.} Subject labels are OmniScience's coarse source metadata rather than manual content annotations; the two released corpora concentrate in biology and medicine subjects (Table~\ref{tab:subject-dist}). All benchmark items derive from Nature Communications, whereas \textsc{OmniVCTrain} draws mainly from Nature Communications (64.0\%), bioRxiv (10.5\%), Cell Reports (8.7\%), Scientific Reports (5.7\%), and PLOS ONE (4.8\%). Table~\ref{tab:text-length} reports question and answer lengths.

\begin{table}[!ht]
  \caption{Subject metadata distribution (QA-row weights; source-record weights differ by $<1$ point in each cell).}
  \label{tab:subject-dist}
  \centering
  \small
  \begin{tabular}{@{}lrr@{}}
    \toprule
    Subject & \textsc{OmniVCTrain} & \textsc{OmniVCBench} \\
    \midrule
    Biology & 75.5\% & 73.3\% \\
    \rowcolor{rowgray} Medicine & 21.7\% & 26.4\% \\
    Physics & 0.2\% & 0.4\% \\
    \rowcolor{rowgray} Others (train only) & 2.6\% & 0.0\% \\
    \bottomrule
  \end{tabular}
\end{table}

\begin{table}[!ht]
  \caption{Question and answer lengths (Unicode characters / whitespace-separated words).}
  \label{tab:text-length}
  \centering
  \small
  \setlength{\tabcolsep}{3.5pt}
  \begin{tabular}{@{}lrrrr@{}}
    \toprule
    Field & Median chars & Mean chars & Median words & Mean words \\
    \midrule
    Train questions & 147 & 154.2 & 24 & 25.3 \\
    \rowcolor{rowgray} Train answers & 235 & 249.1 & 35 & 37.3 \\
    Bench questions & 234 & 238.0 & 37 & 37.3 \\
    \rowcolor{rowgray} Bench answers & 358 & 364.1 & 52 & 53.6 \\
    \bottomrule
  \end{tabular}
\end{table}

\textbf{Source-paper publication years.} Crossref DOI records resolve the publication years of all 1,080 source papers, with standardized titles matching the benchmark records. The papers span 2011--2017 (Table~\ref{tab:year-dist}). MCQ accuracy varies across years without a monotone trend in the evaluated models (Table~\ref{tab:year-accuracy}). The overlap audit in Appendix~\ref{app:leakage-audit} addresses the released training and benchmark corpora; the present evaluation uses this historical source distribution.

\begin{table}[!ht]
  \caption{Publication-year distribution of the 1,080 source papers and the 6,077 benchmark items (Crossref, by DOI). Papers are deduplicated by DOI; QA counts keep the per-paper item weight.}
  \label{tab:year-dist}
  \centering
  \small
  \setlength{\tabcolsep}{4pt}
  \begin{tabular}{@{}lrr@{}}
    \toprule
    Year & Papers & QA items \\
    \midrule
    2011 & 30 (2.8\%) & 182 (3.0\%) \\
    \rowcolor{rowgray} 2012 & 121 (11.2\%) & 723 (11.9\%) \\
    2013 & 233 (21.6\%) & 912 (15.0\%) \\
    \rowcolor{rowgray} 2014 & 25 (2.3\%) & 110 (1.8\%) \\
    2015 & 160 (14.8\%) & 737 (12.1\%) \\
    \rowcolor{rowgray} 2016 & 152 (14.1\%) & 861 (14.2\%) \\
    2017 & 359 (33.2\%) & 2,552 (42.0\%) \\
    \bottomrule
  \end{tabular}
\end{table}

\begin{table}[!ht]
  \caption{Item-macro MCQ accuracy (\%) by source-paper publication year. Year cells cover 110--2,552 items each (Table~\ref{tab:year-dist}); no model shows a monotone year trend.}
  \label{tab:year-accuracy}
  \centering
  \scriptsize
  \setlength{\tabcolsep}{3.5pt}
  \begin{tabular}{@{}lrrrrrrr@{}}
    \toprule
    Model & 2011 & 2012 & 2013 & 2014 & 2015 & 2016 & 2017 \\
    \midrule
    GPT-5.6-sol & 51.1 & 53.3 & 61.0 & 57.3 & 55.8 & 54.1 & 53.8 \\
    \rowcolor{rowgray} Grok-4.6 & 49.5 & 55.3 & 42.7 & 59.1 & 57.9 & 50.4 & 49.1 \\
    Claude-Sonnet-4-5 & 49.5 & 43.9 & 52.6 & 44.6 & 48.2 & 48.6 & 50.6 \\
    \rowcolor{rowgray} GPT-5.6-luna & 47.3 & 38.0 & 48.9 & 38.2 & 41.7 & 42.0 & 44.7 \\
    Qwen3-VL-8B-SFT & 29.1 & 32.6 & 36.6 & 38.2 & 30.3 & 35.8 & 34.8 \\
    \rowcolor{rowgray} Qwen3-VL-8B & 29.1 & 32.2 & 35.8 & 35.5 & 30.8 & 33.7 & 34.8 \\
    Qwen3-VL-4B & 26.4 & 27.4 & 31.7 & 34.6 & 31.5 & 31.6 & 30.4 \\
    \rowcolor{rowgray} InternVL3.5-8B & 25.8 & 30.7 & 31.5 & 32.7 & 28.5 & 28.2 & 31.2 \\
    InternVL3.5-4B & 23.1 & 25.6 & 29.9 & 33.6 & 30.0 & 28.7 & 29.4 \\
    \rowcolor{rowgray} Qwen2.5-VL-7B & 25.3 & 25.6 & 27.5 & 28.2 & 24.4 & 27.1 & 27.1 \\
    LLaVA-OneVision-7B & 22.5 & 23.4 & 27.7 & 25.5 & 25.4 & 24.9 & 25.9 \\
    \rowcolor{rowgray} Qwen2.5-VL-3B & 20.3 & 19.2 & 25.8 & 17.3 & 23.1 & 23.1 & 22.3 \\
    InternVL3.5-2B & 18.1 & 18.1 & 22.5 & 21.8 & 20.6 & 22.0 & 20.9 \\
    \rowcolor{rowgray} Qwen3-VL-2B & 14.8 & 18.7 & 24.8 & 18.2 & 21.3 & 23.3 & 21.6 \\
    SmolVLM2-2.2B & 17.0 & 18.8 & 17.0 & 17.3 & 15.5 & 18.2 & 17.2 \\
    \rowcolor{rowgray} LLaVA-Med-Mistral-7B & 17.6 & 15.8 & 17.5 & 20.0 & 15.3 & 16.1 & 15.1 \\
    \bottomrule
  \end{tabular}
\end{table}

\textbf{Species, assay, and cell-type coverage.} All 1,080 source papers were annotated for species, experimental assays, and cell types (Table~\ref{tab:coverage}). The corpus is dominated by mouse (49.9\% of papers) and human (38.2\%) studies, with smaller shares of rat (5.1\%), bacteria (3.4\%), fruit fly (2.5\%), yeast (2.3\%), zebrafish (2.2\%), plant (1.5\%), and worm (1.0\%). Microscopy (61.9\%) and immunoblot (41.8\%) are the most frequent assays, and the top cell types are standard cell lines (HeLa 7.4\%, HEK293T 4.1\%, HEK293 3.4\%).

\begin{table}[!ht]
  \caption{Coverage of the 1,080 source papers (\% of papers carrying each label). Labels are multi-valued and not normalized (variants such as MEF/MEFs and MCF-7/MCF7 co-occur), so entries within a block need not sum to 100\%.}
  \label{tab:coverage}
  \centering
  \small
  \setlength{\tabcolsep}{4pt}
  \begin{tabular}{@{}lr@{\hspace{1.8em}}lr@{\hspace{1.8em}}lr@{}}
    \toprule
    \multicolumn{2}{@{}l}{\emph{Species}} & \multicolumn{2}{l}{\emph{Assays}} & \multicolumn{2}{l}{\emph{Cell types (top 10)}} \\
    \midrule
    mouse & 49.9 & microscopy & 61.9 & HeLa & 7.4 \\
    \rowcolor{rowgray} human & 38.2 & immunoblot & 41.8 & HEK293T & 4.1 \\
    rat & 5.1 & genetic perturbation & 37.4 & HEK293 & 3.4 \\
    \rowcolor{rowgray} bacteria & 3.4 & qPCR & 36.9 & U2OS & 2.3 \\
    fruit fly & 2.5 & binding interaction & 19.4 & fibroblasts & 2.3 \\
    \rowcolor{rowgray} yeast & 2.3 & flow cytometry & 19.4 & MDA-MB-231 & 2.2 \\
    zebrafish & 2.2 & viability/proliferation & 17.1 & MEF & 2.2 \\
    \rowcolor{rowgray} plant & 1.5 & chromatin assay & 12.7 & endothelial cells & 2.2 \\
    worm & 1.0 & microarray & 10.0 & neurons & 2.2 \\
    \rowcolor{rowgray} other & 6.0 & RNA sequencing & 9.9 & MCF-7 & 2.1 \\
    \bottomrule
  \end{tabular}
\end{table}

\subsection{Item and Option Audits}
\label{app:item-audits}

The response-derived difficulty statistic has median 0.25 and mean 0.3166. Its point-biserial correlation is defined per item as the leave-one-item-out point-biserial correlation of correctness across models; it is available for 5,606/6,077 items (mean 0.3143, median 0.3877, 10th percentile $-0.1839$, 90th percentile 0.7269), with the remaining 471 items answered incorrectly by every model (zero variance, undefined correlation). An eleven-model sensitivity analysis yields mean 0.2922 (5,518 items). This is a descriptive discrimination statistic within the evaluated model pool, not a human answerability measure. Correct-option counts are A/B/C/D/E/F = 1,001/999/964/1,011/1,036/1,066 (15.86\%--17.54\%), and every distractor originates from an observed model error. These statistics are based on model responses and do not measure human answerability or inter-annotator agreement.

A stratified 300-item sample---127 L1, 87 L2, and 86 L3 questions from 251 source papers, balanced by single- and multi-subfigure structure---supports the input comparison below, the distractor ablations of Appendix~\ref{app:mdhnm-details}, and the case studies of Appendix~\ref{app:case-studies}.

\textbf{No-image shortcut test.} Five models answer the stratified 300-item sample using only the question and six options (Table~\ref{tab:shortcut}). GPT-5.6-sol drops from 57.3\% with images to 34.7\% without images, a decrease of 22.6 percentage points. InternVL3.5-2B and SmolVLM2-2.2B each score 13.7\% without images, against a 16.7\% random baseline. LLaVA-OneVision-7B changes from 23.0\% to 24.0\%, and Qwen3-VL-2B from 20.3\% to 21.7\%. The contribution of visual input is therefore model-dependent. GPT-5.6-sol benefits from images and also extracts useful information from the question and options, domain priors, and potentially memorized source content (Appendix~\ref{app:limitations}).

\begin{table}[!ht]
  \caption{No-image shortcut test on the stratified 300-item sample: MCQ accuracy (\%) when models receive only the question and the six options, without the image or any transcribed evidence. The six-option random baseline is 16.7\%. For reference, GPT-5.6-sol scores 57.3\% with the image on the same items.}
  \label{tab:shortcut}
  \centering
  \small
  \begin{tabular}{@{}lrrrrr@{}}
    \toprule
    Model  & L1 & L2 & L3 & Avg \\
    \midrule
    GPT-5.6-sol  & 37.8 & 39.1 & 25.6 & 34.7 \\
    LLaVA-OneVision-7B & 23.6 & 25.3 & 23.3 & 24.0  \\
    Qwen3-VL-2B  & 22.0 & 23.0 & 19.8 & 21.7 \\
    InternVL3.5-2B & 11.0 & 16.1 & 15.1 & 13.7 \\
    SmolVLM2-2.2B & 11.8 & 16.1 & 14.0 & 13.7 \\
    \bottomrule
  \end{tabular}
\end{table}

% \FloatBarrier
\subsection{MDHNM Screening Criteria}
\label{app:mdhnm-details}

For each open-ended item, twelve rollouts per pool model provide naturally occurring candidate errors. We use HuatuoGPT-Vision-7B \citep{chen2024huatuogptvisioninjectingmedicalvisual}, minimax-m3 \citep{minimaxai2026minimaxm3} and glm-5v-turbo \citep{hong2026glm5vturbo} to generate the error pool. Responses with AIVC-Judge overall scores above 2, judged with the original image included as in production scoring, are removed from the error pool. Kimi-K2.6 then rewrites at most five retained errors using minimal edits, preserving the wrong inference while aligning answer length, syntax, units, and level of detail. The final review applies the criteria in Table~\ref{tab:mdhnm-criteria}; near-duplicates are merged or removed before option ordering. These construction criteria are complemented by the unanimous three-annotator audit (Appendix~\ref{app:construction-funnel}).

\begin{table}[!htb]
  \caption{Criteria for selecting MDHNM distractors.}
  \label{tab:mdhnm-criteria}
  \centering
  \small
  \begin{tabularx}{\linewidth}{@{}lX@{}}
    \toprule
    Criterion & Requirement \\
    \midrule
    Strict incorrectness & The screening target is an option contradicted by the evidence or incompatible with the requested answer; an alternative hypothesis requires a substantive scientific distinction. \\
    Plausibility & The error should reflect a biologically or visually plausible model failure. \\
    Relevance and grounding & The option must answer the question and refer only to entities or patterns present in the item. \\
    Diversity & The five distractors should represent distinct error modes rather than paraphrases. \\
    Uniqueness & The screening target is one correct option with semantically distinct distractors; subsequent audits examine violations of this target. \\
    Style consistency & Length, grammar, specificity, and units should not reveal the correct answer. \\
    \bottomrule
  \end{tabularx}
\end{table}

\subsubsection{Direct-Generation Distractor Ablation}
\label{app:mdhnm-ablation}

We compare MDHNM distractors with distractors written directly by strong generators on the stratified 300-item sample (Appendix~\ref{app:item-audits}). Each baseline generator---Grok-4.6 \citep{xai2026introducinggrok46}, Gemini-3.1-flash-lite \citep{google2026gemini31flash}, and Kimi-K2.6 \citep{moonshotai2026kimik26}---receives only the question and the reference answer and writes five distractors directly, without access to the MDHNM error pool; the original reference answer is inserted once among the six options, with option letters matching the corresponding released MDHNM item. A fixed solver, GPT-5.6-sol, answers at temperature 0 from the original image, the question, and the six options. The primary metric is the fixed solver's MCQ accuracy: the easier a generator's distractors, the higher the solver scores. Confidence intervals for accuracy differences use 20,000 source-paper cluster bootstrap replicates, and paired significance uses an exact two-sided McNemar test.

\begin{table}[!ht]
  \caption{Direct-generation distractor ablation on the stratified 300-item sample: MCQ accuracy (\%) of the fixed solver (GPT-5.6-sol, temperature 0) on items whose distractors were written directly by each baseline generator, versus the released MDHNM items on the same questions. Discordant pairs are (correct only on direct) / (correct only on MDHNM); $p$ is the exact two-sided McNemar test, and bracketed intervals are source-paper cluster bootstrap 95\% CIs (20,000 replicates).}
  \label{tab:mdhnm-ablation}
  \centering
  \small
  \setlength{\tabcolsep}{3.5pt}
  \begin{tabular}{@{}lrrrrr@{}}
    \toprule
    Distractor generator & Direct & MDHNM & $\Delta$ (pp) [95\% CI] & Discordant & McNemar $p$ \\
    \midrule
    Grok-4.6 & 87.7 & 57.3 & $+30.3$ [$+23.8$, $+36.7$] & 111/20 & $<10^{-6}$ \\
    \rowcolor{rowgray} Gemini-3.1-flash-lite & 95.3 & 57.3 & $+38.0$ [$+32.2$, $+43.9$] & 120/6 & $<10^{-6}$ \\
    Kimi-K2.6 & 93.3 & 57.3 & $+36.0$ [$+30.1$, $+41.9$] & 116/8 & $<10^{-6}$ \\
    \bottomrule
  \end{tabular}
\end{table}

\begin{table}[!ht]
  \caption{Per-level breakdown of the direct-generation ablation (127 L1, 87 L2, 86 L3 items). The MDHNM column is the fixed solver's accuracy on the released items; it repeats within each level block because it does not depend on the baseline generator.}
  \label{tab:mdhnm-ablation-levels}
  \centering
  \small
  \setlength{\tabcolsep}{4pt}
  \begin{tabular}{@{}llrrr@{}}
    \toprule
    Level & Distractor generator & Direct & MDHNM & $\Delta$ (pp) \\
    \midrule
    L1 & Grok-4.6 & 86.6 & 63.8 & $+22.8$ \\
    \rowcolor{rowgray} L1 & Gemini-3.1-flash-lite & 91.3 & 63.8 & $+27.6$ \\
    L1 & Kimi-K2.6 & 92.9 & 63.8 & $+29.1$ \\
    \midrule
    \rowcolor{rowgray} L2 & Grok-4.6 & 86.2 & 65.5 & $+20.7$ \\
    L2 & Gemini-3.1-flash-lite & 97.7 & 65.5 & $+32.2$ \\
    \rowcolor{rowgray} L2 & Kimi-K2.6 & 92.0 & 65.5 & $+26.4$ \\
    \midrule
    L3 & Grok-4.6 & 90.7 & 39.5 & $+51.2$ \\
    \rowcolor{rowgray} L3 & Gemini-3.1-flash-lite & 98.8 & 39.5 & $+59.3$ \\
    L3 & Kimi-K2.6 & 95.3 & 39.5 & $+55.8$ \\
    \bottomrule
  \end{tabular}
\end{table}

\textbf{Same-evidence ablation.} The direct-generation baselines above receive less evidence than the MDHNM pipeline. To compare direct generation with error conditioning under matched inputs, we built a same-evidence comparison in which two generator arms both receive the original image, question, reference answer, and caption/context under identical budgets (five distractors, temperature 0, 8,192 output tokens, at most two attempts), with the correct option's text and position frozen; the direct arm writes distractors from scratch while the mined arm is conditioned on mined inference errors. Two solvers answer the resulting items: GPT-5.6-sol and Grok-4.6, neither of which contributed to the released items' construction. Table~\ref{tab:mdhnm-same-evidence} reports accuracy across five distractor arms. Direct distractors remain far easier than the released MDHNM items for both solvers, and the paired direct-versus-mined comparison for the same Gemini-3.1-flash-lite generator is small and not significant after Holm correction: the direct arm exceeds the mined arm by $+3.7$pp for GPT-5.6-sol [$+0.3$, $+7.1$] and $+2.7$pp for Grok-4.6 [$-0.3$, $+5.7$] (CI on the accuracy difference; significance by exact McNemar, $p=.052$ and $p=.115$). One arm carries a length artifact: the correct option is the longest in 57.3\% of Gemini-3.7-flash-high direct items (0.037--0.150 elsewhere); a symmetric length/identity filter ($n=299$) leaves all conclusions unchanged.

\begin{table}[!ht]
  \caption{Same-evidence distractor ablation on the stratified 300-item sample: MCQ accuracy (\%) of two solvers on five distractor arms. Both generator arms receive identical evidence and budgets; mined arms are conditioned on mined inference errors, direct arms are not. original MDHNM denotes the released items. For the paired gemini-3.1-flash-lite arms, a positive direct-minus-mined difference means the direct distractors are easier.}
  \label{tab:mdhnm-same-evidence}
  \centering
  \small
  \setlength{\tabcolsep}{4pt}
  \begin{tabular}{@{}lrr@{}}
    \toprule
    Distractor arm & GPT-5.6-sol & Grok-4.6 \\
    \midrule
    gemini-3.1-flash-lite direct & 93.7 & 94.0 \\
    \rowcolor{rowgray} gemini-3.1-flash-lite mined & 90.0 & 91.3 \\
    gemini-3.7-flash-high direct & 94.3 & 93.7 \\
    \rowcolor{rowgray} gpt-5.6-terra direct & 92.0 & 91.7 \\
    original MDHNM & 57.3 & 48.3 \\
    \bottomrule
  \end{tabular}
\end{table}

Under the question-and-reference setting, GPT-5.6-sol scores 30.3--38.0 percentage points higher on directly generated distractors than on released MDHNM items. The same-evidence study extends the comparison to two solvers. With Gemini-3.1-flash-lite as the generator, direct-minus-mined differences are 3.7 and 2.7 points. Neither contrast reaches significance under the reported Holm correction. Regenerated mined arms yield 90.0\% and 91.3\% solver accuracy, compared with 57.3\% and 48.3\% for the released items. The observed difficulty difference therefore concerns the complete released construction configuration. We report MDHNM as the benchmark's distractor-construction procedure, with difficulty and option validity assessed separately: error-pool mining supplies candidate failure seeds, while the iterative rewriting, screening criteria, and unanimous audit funnel drive the final discriminative difficulty.

% \FloatBarrier
\subsection{AIVC-Judge Rubrics and Production Configuration}
\label{app:judge-details}

Table~\ref{tab:judge-rubrics} lists the scoring dimensions selected for each task level. Each dimension uses a 1--5 integer scale with checklist-based guidance. For L1, a key numerical, directional, or entity error caps correctness at 2. For L2, a claim contradicted by the closed evidence caps faithfulness at 2. These rules penalize decisive scientific errors even in otherwise fluent responses.

\begin{table}[!htb]
  \caption{Level-conditioned dimensions used by AIVC-Judge. The recorded L3 dimensions assess source-referenced critique and argumentation.}
  \label{tab:judge-rubrics}
  \centering
  \small
  \begin{tabularx}{\linewidth}{@{}lX@{}}
    \toprule
    Level & Scoring dimensions \\
    \midrule
    L1 & correctness; evidence support \\
    L2 & faithfulness; causal completeness; mechanistic granularity; evidence consistency \\
    L3 & judgment correctness; argumentation quality; evidence weighing \\
    \bottomrule
  \end{tabularx}
\end{table}

The judge returns atomic claims, evidence verdicts, a scoring rationale, dimension scores, and a raw overall score. The raw overall score is a holistic judgment made after claim checking and dimension scoring. The parser rounds overall scores to integers and clips them to the 1--5 scale. Records with dimension ratings but no overall score contribute only to dimension-based analyses. Main tables average these item-level overall scores; dimension averages provide separate diagnostics. JSON validation checks the required fields and the 1--5 score range. Malformed responses are retried, and persistent failures are logged separately from valid scores.

\begin{table}[!htb]
  \caption{Production AIVC-Judge configuration used for the reported open-response scores.}
  \label{tab:judge-production}
  \centering
  \small
  \begin{tabularx}{\linewidth}{@{}lX@{}}
    \toprule
    Component & Reported configuration \\
    \midrule
    Judge & DeepSeek-V4-Flash-Vision-Exp (\texttt{deepseek-v4-flash-vision-exp}); single pointwise judge; temperature 0 \\
    Response input & question and candidate response \\
    Closed evidence & figure caption, surrounding paper context, and complete reference answer \\
    Visual input & original figure image set, as shown to the answering model \\
    Reference key points & disabled; the complete reference answer provides the scoring anchor \\
    Anonymization & evaluated model identity omitted from the judge input \\
    Output & claim verdicts, rationale, dimension scores, and raw overall score \\
    \bottomrule
  \end{tabularx}
\end{table}

\subsection{Human Validation Sample and MCQ Performance}
\label{app:human-study}
\label{app:visual-judge-audit}

The validation sample contained 300 questions from 251 source papers, stratified by AIVC level and single-/multi-subfigure structure. It comprised 127 L1, 87 L2, and 86 L3 questions. Three human annotators, denoted Human 1--3, participated in both MCQ answering and open-response scoring. They are cell biology PhD annotators, forming a group independent of the three construction auditors (Appendix~\ref{app:construction-funnel}). For MCQ answering, they received the original image, question, and six options, with the reference answer and answer key withheld. Responses followed the sample order with fixed option positions.

MCQ accuracy used all 300 assigned questions, counting missing answers as incorrect (Table~\ref{tab:human-mcq}). Human 1 provided 283 valid answers; Humans 2 and 3 each provided 300. Accuracy was 55.3\%, 71.3\%, and 74.7\%, respectively, with the lowest level-wise accuracy on L3 for each annotator.

\begin{table}[!htb]
  \caption{Human MCQ accuracy (\%) on the fixed 300-question sample. Level denominators are L1/L2/L3 = 127/87/86. Human 1's 17 missing answers count as incorrect. Brackets denote 95\% source-paper cluster bootstrap confidence intervals.}
  \label{tab:human-mcq}
  \centering
  \small
  \setlength{\tabcolsep}{4pt}
  \begin{tabular}{@{}llrrrr@{}}
    \toprule
    Annotator & Correct/assigned & Overall [95\% CI] & L1 & L2 & L3 \\
    \midrule
    Human 1 & 166/300 & 55.3 [50.0, 60.7] & 55.9 & 57.5 & 52.3 \\
    Human 2 & 214/300 & 71.3 [65.8, 76.7] & 76.4 & 81.6 & 53.5 \\
    Human 3 & 224/300 & 74.7 [69.1, 79.5] & 75.6 & 87.4 & 60.5 \\
    \bottomrule
  \end{tabular}
\end{table}

Option agreement was estimated on the 283 questions with valid answers from all three annotators. Fleiss' $\kappa$ was 0.589 [0.543, 0.635], and nominal Krippendorff's $\alpha$ was also 0.589 [0.543, 0.635]. Table~\ref{tab:human-iaa} reports pairwise agreement, treating the six answer options as nominal categories. Missing answers were excluded from agreement estimation and were never assigned an artificial option.

\begin{table}[!htb]
  \caption{Human--human agreement on selected MCQ options. Each comparison uses the same 283 complete questions. Brackets denote 95\% source-paper cluster bootstrap confidence intervals.}
  \label{tab:human-iaa}
  \centering
  \small
  \begin{tabular}{@{}lrr@{}}
    \toprule
    Annotator pair & Cohen's $\kappa$ [95\% CI] & Exact agreement \\
    \midrule
    Human 1--Human 2 & 0.456 [0.391, 0.520] & 54.8\% \\
    Human 1--Human 3 & 0.457 [0.391, 0.523] & 54.8\% \\
    Human 2--Human 3 & 0.855 [0.805, 0.897] & 88.0\% \\
    \bottomrule
  \end{tabular}
\end{table}

\subsection{Human Scoring of Open-Ended Responses}
\label{app:human-judge-agreement}

The same three annotators (Human 1--3, Appendix~\ref{app:human-study}) scored existing responses from GPT-5.6-sol, Qwen3-VL-8B, and LLaVA-Med-Mistral-7B on the stratified 300-item sample (Appendix~\ref{app:item-audits}). Each candidate supplied one nonempty response per question, yielding 900 candidate--question response cells. Human scorers received the question, anonymous candidate response, caption, surrounding context, and reference answer, but neither the original image nor other scorers' ratings. The production judge scores the same responses with the original figure image as additional visual input; human--judge agreement therefore concerns the shared rubric and closed textual evidence rather than identical visual inputs.

\paragraph{Scores, coverage, and statistical units.}
The primary validation uses raw overall scores on the same 1--5 scale as the main results. Available raw ratings numbered 887, 900, and 900 for Humans 1--3, and 897 for the production judge. All four scorers had valid raw ratings on 884 response cells from 297 questions and 249 papers. Pairwise analyses use each scorer pair's available intersection; the panel analysis requires valid scores from all four scorers. No missing score was imputed for correlations or agreement coefficients. Repeated ratings were resolved by retaining the last valid record for each annotator--candidate--question combination.

The secondary analysis uses rubric-weighted dimension scores, with weights of 0.60/0.40 for the two L1 dimensions. Weights for L2 are 0.40/0.25/0.20/0.15, and those for L3 are 0.40/0.30/0.30, in the dimension order of Table~\ref{tab:judge-rubrics}. These weights are unequal, and weighted scores are retained without rounding. Valid dimension records numbered 890, 900, and 900 for Humans 1--3, and 898 for the judge. Their common intersection contained 888 cells from 297 questions and 249 papers.

All Human-validation confidence intervals use 2,000 source-paper cluster bootstrap replicates, with percentile endpoints at 2.5\% and 97.5\%. Each replicate resamples source papers and retains their associated questions, candidate responses, and paired ratings. Point estimates weight response cells equally; paper clustering accounts for their shared sources in interval estimation. Spearman correlations retain score ties, and pairwise raw-score QWK uses the five integer categories with quadratic weights $((a-b)/4)^2$. The human panel mean remains unrounded and is summarized using correlation, bias, and mean absolute error (MAE).

\paragraph{Candidate ranking and overall agreement.}
All three annotators and the production judge ranked GPT-5.6-sol above Qwen3-VL-8B, followed by LLaVA-Med-Mistral-7B (Table~\ref{tab:human-judge-candidates}). A fixed-300 sensitivity assigning missing scores the scale minimum of 1 preserved this ordering. Absolute score levels differed across annotators, as shown by their candidate means.

\begin{table}[!htb]
  \caption{Observed mean raw overall scores for the three response candidates.}
  \label{tab:human-judge-candidates}
  \centering
  \small
  \setlength{\tabcolsep}{4pt}
  \begin{tabular}{@{}lrrr@{}}
    \toprule
    Scorer & GPT-5.6-sol & Qwen3-VL-8B & LLaVA-Med-Mistral-7B \\
    \midrule
    Human 1 & 3.163 & 2.186 & 1.727  \\
    Human 2 & 4.127 & 2.657  & 1.917  \\
    Human 3 & 3.933  & 2.447  & 1.797  \\
    Production judge & 3.354 & 2.257 & 1.790 \\
    \bottomrule
  \end{tabular}
\end{table}

The panel mean correlated with the production judge at Spearman $\rho=0.877$ [0.855, 0.897] across 884 raw-score cells. Pearson correlation was $r=0.880$ [0.857, 0.899], with MAE 0.464 and a human-minus-judge bias of $+0.188$. Within-candidate Spearman correlations ranged from 0.778 to 0.865 (Table~\ref{tab:human-judge-agreement}). The rubric-weighted analysis gave $\rho=0.893$ [0.873, 0.909] across 888 cells. Equal dimension weighting yielded a similar correlation of 0.892 [0.872, 0.909] on the same cells.

\begin{table}[!htb]
  \caption{Production-judge agreement with the mean of the three human ratings. The unit is a candidate--question response cell. Rows use raw overall scores except the final weighted-score sensitivity. Bias is human minus judge; brackets denote 95\% source-paper cluster bootstrap confidence intervals.}
  \label{tab:human-judge-agreement}
  \centering
  \small
  \setlength{\tabcolsep}{4pt}
  \begin{tabular}{@{}lrrrr@{}}
    \toprule
    Scope & Paired cells & Spearman [95\% CI] & MAE & Bias \\
    \midrule
    All, raw overall & 884 & 0.877 [0.855, 0.897] & 0.464 & +0.188 \\
    L1 & 377 & 0.909 [0.887, 0.926] & 0.389 & +0.193 \\
    L2 & 257 & 0.875 [0.837, 0.905] & 0.450 & +0.014 \\
    L3 & 250 & 0.709 [0.637, 0.774] & 0.592 & +0.360 \\
    \midrule
    GPT-5.6-sol & 291 & 0.840 [0.797, 0.874] & 0.608 & +0.379 \\
    Qwen3-VL-8B & 296 & 0.865 [0.823, 0.896] & 0.430 & +0.169 \\
    LLaVA-Med-Mistral-7B & 297 & 0.778 [0.721, 0.829] & 0.357 & +0.020 \\
    \midrule
    All, rubric-weighted & 888 & 0.893 [0.873, 0.909] & 0.487 & +0.293 \\
    \bottomrule
  \end{tabular}
\end{table}

\paragraph{Human--human agreement and L3 diagnostics.}
Pairwise human raw-score correlations ranged from 0.820 to 0.953, with QWK from 0.738 to 0.944 (Table~\ref{tab:human-human-agreement}). Human--judge correlations ranged from 0.832 to 0.854, and QWK ranged from 0.779 to 0.851. These comparisons quantify both agreement among annotators and agreement with the production judge.

\begin{table}[!htb]
  \caption{Pairwise agreement on raw overall scores, using each pair's available response cells. QWK denotes quadratic-weighted Cohen's $\kappa$ on integer 1--5 ratings. Brackets denote 95\% source-paper cluster bootstrap confidence intervals.}
  \label{tab:human-judge-per-annotator}
  \label{tab:human-human-agreement}
  \centering
  \small
  \setlength{\tabcolsep}{4pt}
  \begin{tabular}{@{}lrrr@{}}
    \toprule
    Scorer pair & Paired cells & Spearman [95\% CI] & QWK [95\% CI] \\
    \midrule
    Human 1--Human 2 & 887 & 0.820 [0.788, 0.849] & 0.738 [0.694, 0.776] \\
    Human 1--Human 3 & 887 & 0.835 [0.807, 0.859] & 0.800 [0.766, 0.829] \\
    Human 2--Human 3 & 900 & 0.953 [0.942, 0.962] & 0.944 [0.930, 0.956] \\
    \midrule
    Human 1--judge & 884 & 0.838 [0.810, 0.863] & 0.851 [0.821, 0.876] \\
    Human 2--judge & 897 & 0.832 [0.800, 0.860] & 0.779 [0.740, 0.815] \\
    Human 3--judge & 897 & 0.854 [0.830, 0.876] & 0.837 [0.806, 0.863] \\
    \bottomrule
  \end{tabular}
\end{table}

L3 showed lower overall agreement than L1 or L2, with panel--judge $\rho=0.709$ on 250 raw-score cells. The recorded L3 rubric includes competing explanations, reasons for accepting or rejecting alternatives, and evidential limitations in its checklists. Its dimension scores therefore provide source-referenced critique and argumentation diagnostics. The validation supports overall score agreement and candidate ordering under this rubric, while fine-grained L3 dimension agreement is lower.

\subsection{Proposal-Oriented Rubric Sensitivity Pilot}
\label{app:proposal-pilot}

The L3 rubric mismatch noted above motivates a direct test of how sensitive L3 scores are to the rubric itself. We froze 24 L3 items from the stratified 300-item sample's 86 L3 items, selected by a fixed hash order over item IDs (24 source papers; selection fixed before inspecting any scores), and archived all three candidates' open responses, yielding 72 fixed response cells. Two additional cell biology PhD annotators, denoted Human 4 and Human 5 and independent of Human 1--3, rescored every cell under three frozen conditions: (i) image and question only, under a proposal-oriented rubric; (ii) full source evidence (caption, context, and reference answer), under the same rubric; and (iii) full source evidence, under a reimplementation of the critique-oriented checklist of Table~\ref{tab:judge-rubrics}. The proposal rubric's core dimensions are observational fidelity, scientific plausibility, and falsifiable specificity, each an integer from 1 to 5; a discriminative-test dimension is scored only when the item requests it, and unrequested argumentation does not enter the core score. Scoring was blind to candidate identity, options, and answer keys, at temperature 0.

\textbf{Agreement and ranking.} On identical full-source inputs, cross-annotator agreement on overall scores is QWK 0.767 [0.646, 0.868] under the proposal rubric, versus QWK 0.466 [0.332, 0.585] under the critique checklist; on the 56 cells valid under both, the paired values are 0.767 versus 0.459, a difference of $+0.308$ [0.130, 0.468] under 2,000 source-paper cluster bootstrap resamples. The candidate ranking GPT-5.6-sol $\succ$ Qwen3-VL-8B $\succ$ LLaVA-Med-Mistral-7B is preserved on the 17-item intersection valid for every annotator and condition, and on 19 items under a parser-relaxed sensitivity analysis.

\textbf{Score levels and input effect.} Proposal-rubric totals exceed critique totals for the strongest candidate ($+0.800$ [0.600, 0.950] for Human 4 and $+1.750$ [1.417, 2.042] for Human 5 on paired cells), so historical L3 totals are rubric-sensitive rather than a pure measure of hypothesis-proposal ability. Under the proposal rubric with full evidence, Human 4 and Human 5 rate GPT-5.6-sol at 4.750 and 4.833 overall; among core dimensions, falsifiable specificity is the weakest for the open-weight candidates (2.611/3.292 for Qwen3-VL-8B and 1.611/1.625 for LLaVA-Med-Mistral-7B under Human 4/Human 5). Adding source evidence under the fixed proposal rubric changes totals by $+0.074$ [$-0.060$, 0.196] for Human 4 and $-0.028$ [$-0.167$, 0.125] for Human 5, so the source-evidence fields have only a small average effect once the rubric is fixed.

\textbf{Stress controls.} On four frozen base items, concise hypotheses and non-reference alternative hypotheses score 4.25--5.00 overall, pure rhetorical expansion adds at most $+0.25$, and injected observation errors reduce totals by $1.00$ for Human 4 and $2.00$ for Human 5. These four-item controls are diagnostic only, but they are consistent with the proposal rubric rewarding valid alternatives and penalizing observational errors without rewarding verbosity.

We read this pilot as measurement-sensitivity evidence: it supports stable candidate rankings and improved cross-annotator agreement under the proposal-oriented rubric, while establishing neither a validated hypothesis-ability scale nor full-benchmark rescoring, which remain future work.

\subsection{Release-Level Train--Benchmark Overlap Audit}
\label{app:leakage-audit}

This appendix audits the released files themselves---the 548,450-row \textsc{OmniVCTrain} parquet against the 6,077-item \textsc{OmniVCBench} release---rather than inferring isolation from the construction scripts.

\textsc{OmniVCTrain} covers 33,802 papers and the benchmark 1,080. Four cross-set checks---normalized DOI/paper-ID exact match, normalized title exact match, and title character-TFIDF cosine at thresholds $.90$ and $.95$---return zero candidates (Table~\ref{tab:leak-source}). The largest nearest-title cosine observed is $0.8896$, below the $.90$ threshold; that nearest pair (a DAF-16/FOXO record) was manually reviewed and involves a different DOI and a different study.

\begin{table}[!ht]
  \caption{Source-level cross-set overlap checks between the released \textsc{OmniVCTrain} (33,802 papers) and \textsc{OmniVCBench} (1,080 papers).}
  \label{tab:leak-source}
  \centering
  \small
  \begin{tabular}{@{}lrr@{}}
    \toprule
    Check & Cross-set candidates & Confirmed duplicates \\
    \midrule
    Normalized DOI / paper ID (exact) & 0 & 0 \\
    \rowcolor{rowgray} Normalized title (exact) & 0 & 0 \\
    Title char-TFIDF cosine $\geq .90$ & 0 & 0 \\
    \rowcolor{rowgray} Title char-TFIDF cosine $\geq .95$ & 0 & 0 \\
    \bottomrule
  \end{tabular}
\end{table}

Text overlap is computed after Unicode normalization, lowercasing, and word tokenization; candidate pairs are retrieved by bottom-12 signatures (word trigrams for questions, 5-grams for captions) and accepted at Jaccard $\geq .80$ or containment $\geq .90$. Table~\ref{tab:leak-text} reports the counts: questions (548,390 unique train / 6,077 bench), captions (96,007 / 2,625), and answers (exact match only; 544,652 / 6,077); exact and near pairs are all zero. Signature retrieval is not a semantic-paraphrase detector, so Table~\ref{tab:leak-text-synth} reports detection rates on synthetic perturbations.

\begin{table}[!ht]
  \caption{Text-level cross-set overlap between the released corpora. Near pairs require Jaccard $\geq .80$ or containment $\geq .90$ under bottom-12 signature retrieval.}
  \label{tab:leak-text}
  \centering
  \small
  \begin{tabular}{@{}lrrrr@{}}
    \toprule
    Field & Train uniques & Bench uniques & Exact pairs & Near pairs \\
    \midrule
    Question & 548,390 & 6,077 & 0 & 0 \\
    \rowcolor{rowgray} Caption & 96,007 & 2,625 & 0 & 0 \\
    Answer (exact only) & 544,652 & 6,077 & 0 & -- \\
    \bottomrule
  \end{tabular}
\end{table}

\begin{table}[!ht]
  \caption{Synthetic-text sensitivity of the overlap detector ($n=500$ perturbations per row): signature-retrieval recall and acceptance recall (Jaccard $\geq .80$ or containment $\geq .90$).}
  \label{tab:leak-text-synth}
  \centering
  \small
  \begin{tabular}{@{}lrr@{}}
    \toprule
    Perturbation & Retrieval & Acceptance \\
    \midrule
    Question: append 10\% words & 100\% & 100\% \\
    \rowcolor{rowgray} Question: delete one word & 100\% & 98.2\% \\
    Question: substitute one word & 100\% & 87.8\% \\
    \rowcolor{rowgray} Caption: append 10\% words & 100\% & 100\% \\
    Caption: delete one word & 100\% & 100\% \\
    \rowcolor{rowgray} Caption: substitute one word & 100\% & 100\% \\
    \bottomrule
  \end{tabular}
\end{table}

Image overlap combines SHA-256 with perceptual hashes (pHash and dHash on EXIF-corrected grayscale images), accepting candidates at Hamming distance $\leq 6$. We hash 327,601 train and 4,428 benchmark images (Table~\ref{tab:leak-image}): zero exact pairs; 22,768 perceptual candidate pairs involving 606 benchmark images under a single hash; and 45 pairs involving 8 benchmark images passing both hashes at distance $\leq 6$, all of which were manually reviewed without confirming any identical, cropped, or recomposed figure. Perceptual similarity alone cannot establish leakage, since microscopy images and charts share generic layouts; Table~\ref{tab:leak-image-synth} reports detection rates on synthetic image perturbations.

\begin{table}[!ht]
  \caption{Image-level cross-set overlap checks (327,601 train and 4,428 benchmark images hashed).}
  \label{tab:leak-image}
  \centering
  \small
  \setlength{\tabcolsep}{3.5pt}
  \begin{tabular}{@{}lrrr@{}}
    \toprule
    Check & Pairs & Bench images involved & Confirmed duplicates \\
    \midrule
    SHA-256 exact & 0 & 0 & 0 \\
    \rowcolor{rowgray} Single-hash perceptual candidates & 22,768 & 606 & -- \\
    Dual-hash (pHash and dHash) Hamming $\leq 6$ & 45 & 8 & 0 \\
    \bottomrule
  \end{tabular}
\end{table}

\begin{table}[!ht]
  \caption{Synthetic-image sensitivity of the perceptual-hash detector ($n=300$ perturbations per row).}
  \label{tab:leak-image-synth}
  \centering
  \small
  \begin{tabular}{@{}lr@{}}
    \toprule
    Perturbation & Detection rate \\
    \midrule
    JPEG quality 75 & 100\% \\
    \rowcolor{rowgray} Resize to half & 100\% \\
    Brightness $\times 1.2$ & 97.7\% \\
    \rowcolor{rowgray} Crop 2\% & 95.0\% \\
    Add 2\% border & 97.3\% \\
    \bottomrule
  \end{tabular}
\end{table}

Across all four groups of checks, no duplicates were confirmed after manual review; we therefore describe the released corpora as having no confirmed overlap rather than claiming absolute leak-freeness.

\subsection{Cluster-Robust Scores and Paper-Weighted Statistics}
\label{app:cluster-stats}

This appendix recomputes the main scores and cross-track correlations on the full benchmark (6,077 items, 1,080 papers) from item-level files without rounding, covering the sixteen MCQ configurations and the eleven open-track models. Confidence intervals are source-paper cluster bootstrap; paper-weighted statistics first average within each source paper and then weight papers equally. Table~\ref{tab:cluster-scores} reports both weightings for every model; the item-level means agree with the rounded values of Table~\ref{tab:main-results} (main text).

Using the original 0/1 predictions on all 6,077 items, exact overall MCQ accuracies are GPT-5.6-sol 55.09\%, Grok-4.6 50.30\%, Claude-Sonnet-4-5 49.37\%, GPT-5.6-luna 43.74\%, Qwen3-VL-8B-SFT 34.28\%, and base Qwen3-VL-8B 33.82\%. Across the eleven models shared by both tracks, model-level Pearson correlation is 0.94 (permutation $p=6.0\times10^{-5}$) and Spearman correlation is 0.96 ($p=2.0\times10^{-5}$).

\textbf{Correlation robustness.} Across the eleven shared models, unrounded correlations are Pearson $r=0.938$ (95\% model-bootstrap CI $[0.843,0.993]$) and Spearman $\rho=0.964$ ($[0.767,1.000]$). Open-weight models correlate strongly ($r=.955$, $n=7$); the four proprietary systems have a weaker correlation ($r=.208$). The pooled statistic therefore includes between-group separation. Leave-one-family-out Pearson correlations remain $0.922$--$0.971$ on strict common items with paper weighting (Table~\ref{tab:cluster-lofo}). Human validation checks the raw overall score directly on the stratified 300-item sample (Appendix~\ref{app:human-judge-agreement}).

\begin{table}[!ht]
  \caption{Item-level and paper-weighted scores with source-paper cluster bootstrap 95\% CIs, computed from unrounded item-level files (6,077 items, 1,080 papers). Top block: MCQ accuracy (\%), sixteen configurations; bottom block: AIVC-Judge (1--5), eleven models.}
  \label{tab:cluster-scores}
  \centering
  \scriptsize
  \setlength{\tabcolsep}{3pt}
  \begin{tabular}{@{}lll@{}}
    \toprule
    Model & Item-level [95\% CI] & Paper-weighted [95\% CI] \\
    \midrule
    \multicolumn{3}{@{}l}{\emph{MCQ accuracy (\%)}} \\
    GPT-5.6-sol & 55.093 [53.768, 56.444] & 56.033 [54.267, 57.794] \\
    \rowcolor{rowgray} Grok-4.6 & 50.304 [48.862, 51.729] & 51.492 [49.738, 53.286] \\
    Claude-Sonnet-4-5 & 49.366 [47.937, 50.768] & 50.123 [48.313, 51.889] \\
    \rowcolor{rowgray} GPT-5.6-luna & 43.739 [42.415, 45.135] & 44.478 [42.678, 46.314] \\
    Qwen3-VL-8B-SFT & 34.277 [33.018, 35.566] & 34.111 [32.444, 35.821] \\
    \rowcolor{rowgray} Qwen3-VL-8B & 33.816 [32.599, 35.047] & 34.233 [32.558, 35.937] \\
    Qwen3-VL-4B & 30.476 [29.248, 31.725] & 31.259 [29.595, 32.902] \\
    \rowcolor{rowgray} InternVL3.5-8B & 30.278 [29.001, 31.550] & 30.557 [28.982, 32.177] \\
    InternVL3.5-4B & 28.879 [27.617, 30.157] & 29.525 [27.870, 31.198] \\
    \rowcolor{rowgray} Qwen2.5-VL-7B & 26.625 [25.455, 27.830] & 26.971 [25.440, 28.560] \\
    LLaVA-OneVision-7B & 25.539 [24.413, 26.683] & 25.564 [24.078, 27.074] \\
    \rowcolor{rowgray} Qwen2.5-VL-3B & 22.495 [21.433, 23.576] & 22.583 [21.164, 24.038] \\
    Qwen3-VL-2B & 21.688 [20.603, 22.761] & 22.122 [20.694, 23.572] \\
    \rowcolor{rowgray} InternVL3.5-2B & 20.866 [19.794, 21.964] & 21.308 [19.911, 22.781] \\
    SmolVLM2-2.2B & 17.311 [16.407, 18.244] & 17.596 [16.361, 18.898] \\
    \rowcolor{rowgray} LLaVA-Med-Mistral-7B & 15.863 [14.961, 16.807] & 16.538 [15.243, 17.912] \\
    \midrule
    \multicolumn{3}{@{}l}{\emph{AIVC-Judge (1--5)}} \\
    GPT-5.6-sol & 3.285 [3.243, 3.325] & 3.286 [3.237, 3.335] \\
    \rowcolor{rowgray} GPT-5.6-luna & 3.161 [3.121, 3.201] & 3.140 [3.090, 3.190] \\
    Grok-4.6 & 3.100 [3.059, 3.142] & 3.111 [3.060, 3.163] \\
    \rowcolor{rowgray} Claude-Sonnet-4-5 & 2.681 [2.638, 2.724] & 2.702 [2.650, 2.757] \\
    Qwen3-VL-8B-SFT & 2.379 [2.341, 2.416] & 2.387 [2.340, 2.435] \\
    \rowcolor{rowgray} Qwen3-VL-8B & 2.333 [2.297, 2.371] & 2.342 [2.294, 2.390] \\
    InternVL3.5-8B & 2.320 [2.283, 2.356] & 2.316 [2.268, 2.365] \\
    \rowcolor{rowgray} Qwen3-VL-4B & 2.226 [2.189, 2.262] & 2.206 [2.162, 2.252] \\
    Qwen2.5-VL-7B & 2.108 [2.076, 2.141] & 2.077 [2.037, 2.118] \\
    \rowcolor{rowgray} LLaVA-OneVision-7B & 2.003 [1.970, 2.036] & 1.984 [1.943, 2.026] \\
    LLaVA-Med-Mistral-7B & 1.861 [1.830, 1.893] & 1.861 [1.820, 1.903] \\
    \bottomrule
  \end{tabular}
\end{table}

Table~\ref{tab:cluster-correlation} reports the MCQ--open-response model-level correlation under three item sets and both weightings. The strict common set contains 5,654 items from 1,058 papers. The paper-cluster bootstrap quantifies sensitivity to paper sampling; it does not increase the number of independent model points, which remains eleven.

\begin{table}[!ht]
  \caption{MCQ--open-response model-level correlation ($n=11$ models) by item set and weighting. Bracketed values are source-paper cluster bootstrap 95\% CIs; $p$ values are permutation-based.}
  \label{tab:cluster-correlation}
  \centering
  \scriptsize
  \setlength{\tabcolsep}{3pt}
  \begin{tabular}{@{}lll@{}}
    \toprule
    Scope & Pearson & Spearman \\
    \midrule
    Published full-track, item & 0.938 ($p{=}6{\times}10^{-5}$) & 0.964 ($p{=}2{\times}10^{-5}$) \\
    \rowcolor{rowgray} Published full-track, paper & 0.945 & 0.955 \\
    Model-specific matched, item & 0.936 & 0.964 \\
    \rowcolor{rowgray} Model-specific matched, paper & 0.943 & 0.955 \\
    Strict common, item & 0.935 [0.919, 0.946] ($p{=}8{\times}10^{-5}$) & 0.964 [0.927, 0.973] ($p{=}3{\times}10^{-5}$) \\
    \rowcolor{rowgray} Strict common, paper & 0.943 [0.923, 0.956] & 0.955 [0.927, 0.973] \\
    \bottomrule
  \end{tabular}
\end{table}

Table~\ref{tab:cluster-lofo} reports leave-one-family-out correlations on the strict common set with paper weighting; the Pearson range is $0.922$--$0.971$ and the Spearman range $0.917$--$0.967$. Family definitions were frozen before this analysis, and removing single-model families is a leverage check rather than a subgroup claim.

\begin{table}[!ht]
  \caption{Leave-one-family-out MCQ--open-response correlation (strict common items, paper-weighted; $n$ = remaining models).}
  \label{tab:cluster-lofo}
  \centering
  \small
  \begin{tabular}{@{}lrrr@{}}
    \toprule
    Excluded family & $n$ & Pearson & Spearman \\
    \midrule
    OpenAI GPT & 9 & 0.956 & 0.967 \\
    \rowcolor{rowgray} xAI Grok & 10 & 0.932 & 0.964 \\
    Anthropic Claude & 10 & 0.971 & 0.964 \\
    \rowcolor{rowgray} Qwen3-VL (incl.\ SFT) & 8 & 0.942 & 0.929 \\
    Qwen2.5-VL & 10 & 0.939 & 0.939 \\
    \rowcolor{rowgray} InternVL & 10 & 0.943 & 0.952 \\
    LLaVA & 9 & 0.922 & 0.917 \\
    \bottomrule
  \end{tabular}
\end{table}

% \FloatBarrier
% \clearpage

\subsection{Prompts and Run Settings}
\label{app:run-settings}

AIVC-Judge uses temperature 0 for pointwise scoring. Its inputs are the question, candidate response, original figure image, caption, surrounding context, and reference answer. Candidate identity is omitted (Table~\ref{tab:judge-production}). Answering models receive the original image and question, plus six options for MCQ. Proprietary systems use the API routes listed in Table~\ref{tab:model-versions}; open-weight checkpoints run locally. The construction-audit protocol and agreement statistics appear in Appendix~\ref{app:construction-funnel}, and human answering and scoring protocols appear in Appendices~\ref{app:human-study} and~\ref{app:human-judge-agreement}. Prompt templates and evaluation code are provided through \url{https://anonymous.4open.science/r/OmniVCBench}.

\section{\textsc{OmniVCTrain} Adaptation Details}
\label{app:omnivctrain-details}

This appendix details the adaptation experiments behind Section~\ref{sec:adaptation}. Holm-corrected values retain the original adaptation comparison family, including configurations omitted from the main-table presentation. The rank-32 configuration gains 1.97 percentage points (95\% paper-cluster bootstrap CI $[0.93,3.02]$; corrected $p=.006$), and multimodal RAG gains 1.60 points (CI $[0.71,2.48]$; corrected $p=.010$). The rank-16 SFT change is 0.46 points (CI $[-0.62,1.54]$). The principal results on the MCQ track are reported in Table~\ref{tab:adaptation-main} (main text); open-response scores in Table~\ref{tab:main-results} cover only the rank-16 SFT configuration, and open-track scoring of the strongest MCQ configurations remains future work. The subsections below describe the supervised fine-tuning and retrieval-augmented generation protocols and report the corresponding ablations.

\subsection{Supervised Fine-Tuning}
\label{app:sft-details}

We format \textsc{OmniVCTrain} as multimodal instruction triples and adapt Qwen3-VL-8B with LLaMA-Factory. Table~\ref{tab:sft-ablation} compares two LoRA settings: rank 16 on the full corpus (548,016 loaded examples) and rank 32 on a 100,000-example subset, both for two epochs. The two settings differ in rank, data size, and batch size jointly, so this is a configuration comparison rather than a rank-only ablation. An independent audit of the archived training logs and inference scripts recovers the shared configuration (learning rate $10^{-4}$, cosine schedule, warmup ratio 0.03, bf16, sequence cutoff 2,048, seed 42, frozen visual tower and projector, LoRA $\alpha$ twice the rank, target modules down/gate/k/o/q/up/v\_proj, effective batch sizes 64 and 56; PEFT 0.18.1, Transformers 5.8.0, Torch 2.8.0) and reproduces all three adaptation accuracies of Table~\ref{tab:adaptation-main} exactly from the archived predictions (34.2768\%, 35.7907\%, and 35.4122\% before rounding). Rank 32 yields the larger observed improvement, 1.97 percentage points over the 33.82\% base model. Source-paper-clustered comparisons and corrected significance are reported in Section~\ref{sec:adaptation}.

\begin{table}[!ht]
  \caption{Principal SFT configurations for Qwen3-VL-8B. Raw paper-block permutation tests compare each setting with the 33.82\% base model on all 6,077 items; Holm-corrected values over the original adaptation comparison family are quoted in the main text.}
  \label{tab:sft-ablation}
  \centering
  \small
  \begin{tabular}{@{}lrrrrrr@{}}
    \toprule
    Setting & Data & Epochs & Rank & Avg & $\Delta$ & $p$ \\
    \midrule
    Full corpus & 548k & 2 & 16 & 34.28 & +0.46 & 0.42 \\
    Rank 32, 100k subset & 100k & 2 & 32 & 35.79 & +1.97 & $3.2\times10^{-4}$ \\
    \bottomrule
  \end{tabular}
\end{table}

% \FloatBarrier
\subsection{Retrieval-Augmented Generation}
\label{app:rag-details}

We construct a 548,450-entry index with Qwen3-VL-Embedding-8B \citep{li2026qwen3}. Each entry jointly encodes its image and question into a 4,096-dimensional, L2-normalized vector. At test time, cosine similarity retrieves $k=3$ QA demonstrations after excluding entries from the same source article and entries with the same figure basename. Retrieved questions and answers are prepended as text demonstrations; the test image remains the visual input to the evaluated MLLM. Retrieved images are discarded, so this baseline evaluates text-demonstration prompting conditioned on multimodal retrieval rather than interleaved multimodal in-context learning, which remains an unexplored adaptation route.

Table~\ref{tab:rag-signal-ablation} compares joint image--question retrieval with image-only and text-only controls at $k=3$. Joint retrieval increases accuracy from 33.82\% to 35.41\%, while the unimodal controls score 33.62\% and 33.40\%, respectively. Table~\ref{tab:rag-cross-model} reports joint-retrieval gains for three base models, ranging from 0.76 to 1.60 percentage points.

\begin{table}[!ht]
  \caption{Retrieval-signal ablation for Qwen3-VL-8B with $k=3$. The base accuracy is 33.82\%. $p$: raw paper-block permutation test vs.\ base.}
  \label{tab:rag-signal-ablation}
  \centering
  \small
  \begin{tabular}{@{}lrrrrrr@{}}
    \toprule
    Retrieval signal & L1 & L2 & L3 & Avg & $\Delta$ & $p$ \\
    \midrule
    Joint image--question & 35.6 & 39.3 & 31.2 & 35.41 & +1.60 & $5.6\times10^{-4}$ \\
    Image only & 34.1 & 37.5 & 28.9 & 33.62 & $-0.20$ & .64 \\
    \rowcolor{rowgray} Text only & 34.1 & 37.3 & 28.5 & 33.40 & $-0.41$ & .31 \\
    \bottomrule
  \end{tabular}
\end{table}

\begin{table}[!ht]
  \caption{Joint multimodal RAG across base models. Improvements are positive across all three models but vary in statistical strength; the InternVL3.5-8B change does not reach significance. $p$: raw paper-block permutation test vs.\ the corresponding base model.}
  \label{tab:rag-cross-model}
  \centering
  \small
  \begin{tabular}{@{}lrrrr@{}}
    \toprule
    Model & Base & $+$RAG & $\Delta$ & $p$ \\
    \midrule
    Qwen3-VL-8B & 33.82 & 35.41 & +1.60 & $5.6\times10^{-4}$ \\
    \rowcolor{rowgray} Qwen3-VL-2B & 21.69 & 23.19 & +1.50 & $2.9\times10^{-3}$ \\
    InternVL3.5-8B & 30.28 & 31.04 & +0.76 & .10 \\
    \bottomrule
  \end{tabular}
\end{table}

\section{Biological Evidence Acquisition and Prediction Revision}
\label{app:refinement-loop}

This appendix documents the four executable pilots behind Section~\ref{sec:refinement-loop}. Their purpose is to test whether the interpretation component measured by \textsc{OmniVCBench} can acquire evidence and revise predictions inside the refinement loop of Figure~\ref{fig:overview}, over real biological outputs rather than curated literature figures. All four pilots use GPT-5.6-sol via the route listed in Table~\ref{tab:model-versions}, with fixed cases chosen before any model call; every model response is archived, and all scores are deterministic recomputations from the archived responses.

\subsection{Task Structure and Evidence Settings}
\label{app:loop-protocol}

\begin{figure}[!ht]
  \centering
  \includegraphics[width=\linewidth]{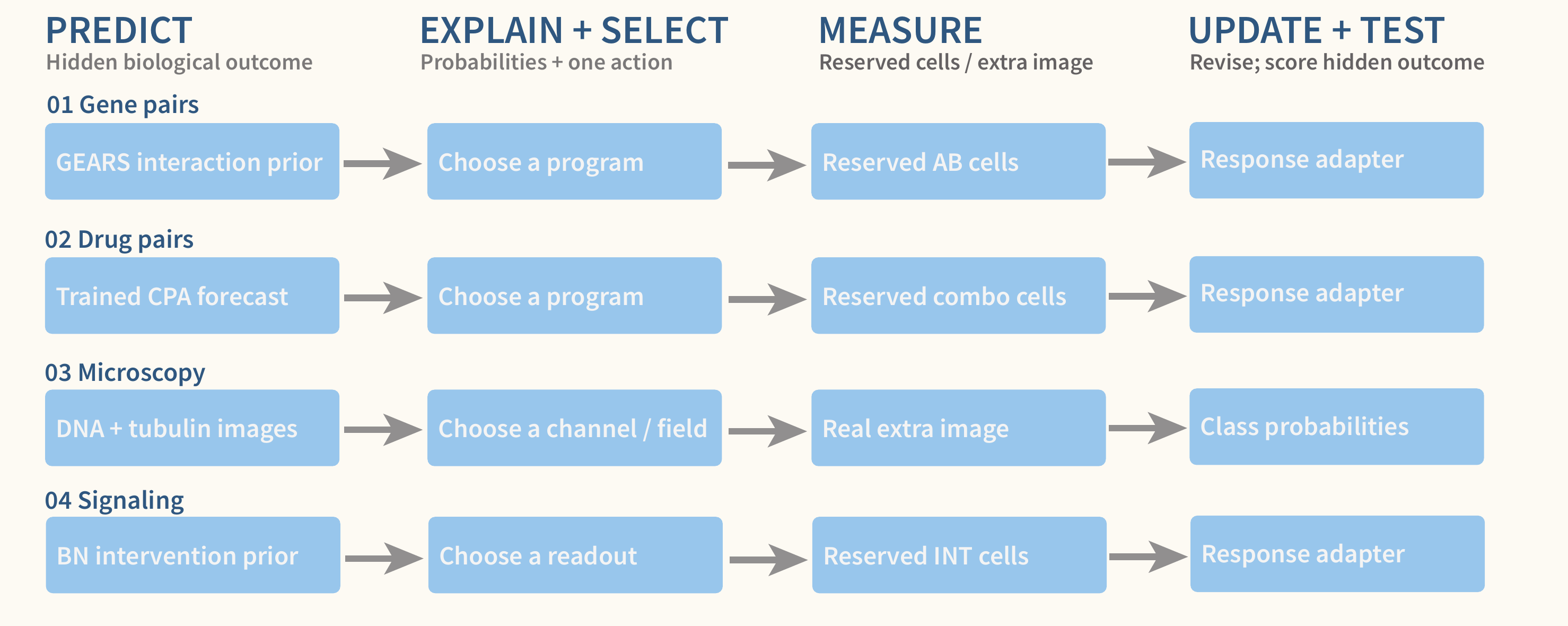}
  \caption{\textbf{Shared pilot structure.} Each pilot runs predict, explain and select, measure, then update and test. Three pilots revise predictions through a response adapter after active evidence acquisition; the microscopy pilot revises its interpretation. These are one-step replay loops: simulator weights and free-text biological mechanisms are not validated or updated.}
  \label{fig:loop-workflow}
\end{figure}

The four pilots share one structure (Figure~\ref{fig:loop-workflow}). The model first predicts a hidden biological output from initial evidence and explains its prediction. A self-review pass then re-examines the same evidence without new information. Independently of the self-review, the model selects one additional measurement to acquire from reserved cells or images; the reserved data are revealed, the model revises its prediction, and the revision is rescored against the hidden ground truth. Each case therefore contributes three scored answers---initial, self-reviewed, and post-evidence---and 36 cases yield the 108 archived model calls. Three pilots update a response-calibration layer on top of the simulator's outputs, and the revision propagates to unqueried readouts through that layer; the microscopy pilot updates class probabilities and the accompanying explanation. Simulator weights are never retrained within a run, and no new wet-lab measurement is performed. Table~\ref{tab:loop-settings} summarizes the four evidence settings.

\begin{table}[!ht]
  \caption{The four evidence-acquisition pilots. Each case admits exactly one additional measurement; readouts are the scored prediction fields. The response adapter is the calibration layer updated by the acquired measurement.}
  \label{tab:loop-settings}
  \centering
  \scriptsize
  \setlength{\tabcolsep}{3pt}
  \begin{tabularx}{\linewidth}{@{}>{\raggedright\arraybackslash}p{0.16\linewidth}>{\raggedright\arraybackslash}p{0.24\linewidth}X>{\raggedright\arraybackslash}p{0.20\linewidth}>{\raggedright\arraybackslash}p{0.09\linewidth}@{}}
    \toprule
    Pilot & Simulator / data & Prediction target & Evidence choice & Cases / readouts \\
    \midrule
    Gene interactions & GEARS \citep{roohani2023predicting}; Norman K562 CRISPRa \citep{norman2019exploring} & non-additive double-gene response $AB-A-B$ per program & one cellular program in reserved AB cells & 12 / 48 \\
    \rowcolor{rowgray} Drug combinations & CPA trained locally \citep{lotfollahi2023predicting}; ComboSciPlex A549 & program mean change under held-out combinations & one program in 48 reserved cells & 10 / 30 \\
    Real microscopy & BBBC021 \citep{caie2010high,ljosa2012annotated}; MCF-7, 5 MoA classes & phenotype-supported mechanism class & F-actin channel or a second field of the same well & 9 / 9 \\
    \rowcolor{rowgray} Signaling & discrete Bayesian network; Sachs discretized cells \citep{sachs2005causal} & high-state fraction change under each intervention & one non-target readout in 96 reserved cells & 5 / 30 \\
    \bottomrule
  \end{tabularx}
\end{table}

\subsection{Results}
\label{app:loop-results}

\begin{figure}[!ht]
  \centering
  \includegraphics[width=\linewidth]{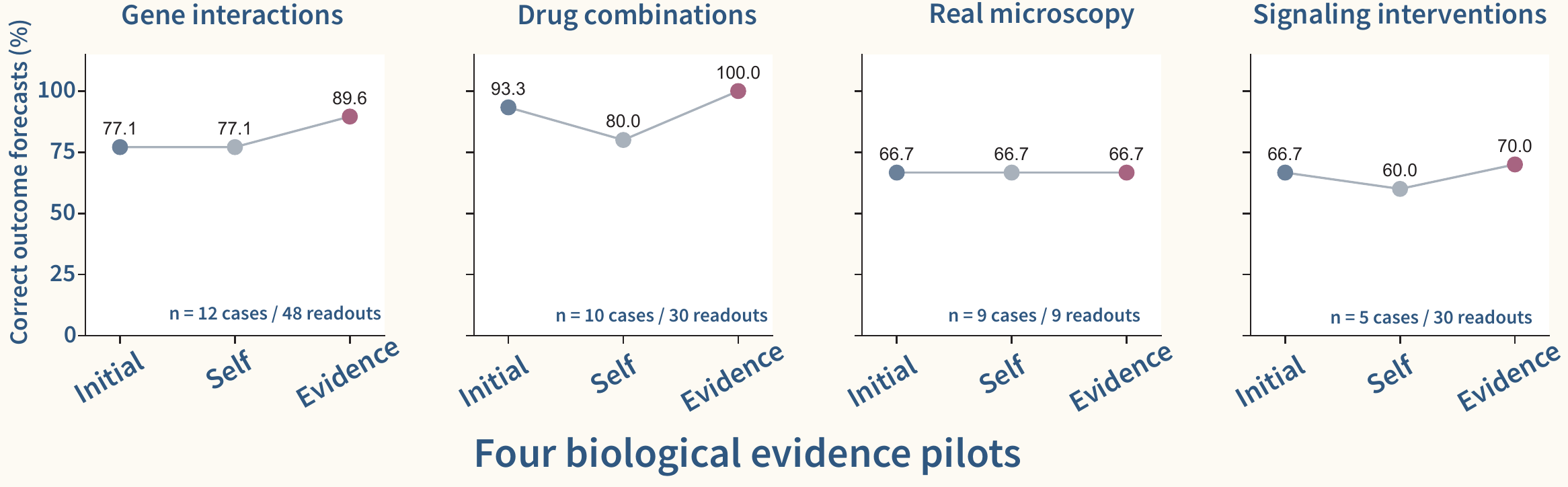}
  \caption{\textbf{Per-pilot outcomes.} Correct-outcome rate under the initial prediction, self-review with the same evidence, and revision after one chosen measurement, for each of the four pilots. Case counts and readout counts differ across pilots; results are descriptive and support no pooled accuracy or causal-discovery claim.}
  \label{fig:loop-results}
\end{figure}

\begin{figure}[!ht]
  \centering
  \includegraphics[width=0.72\linewidth]{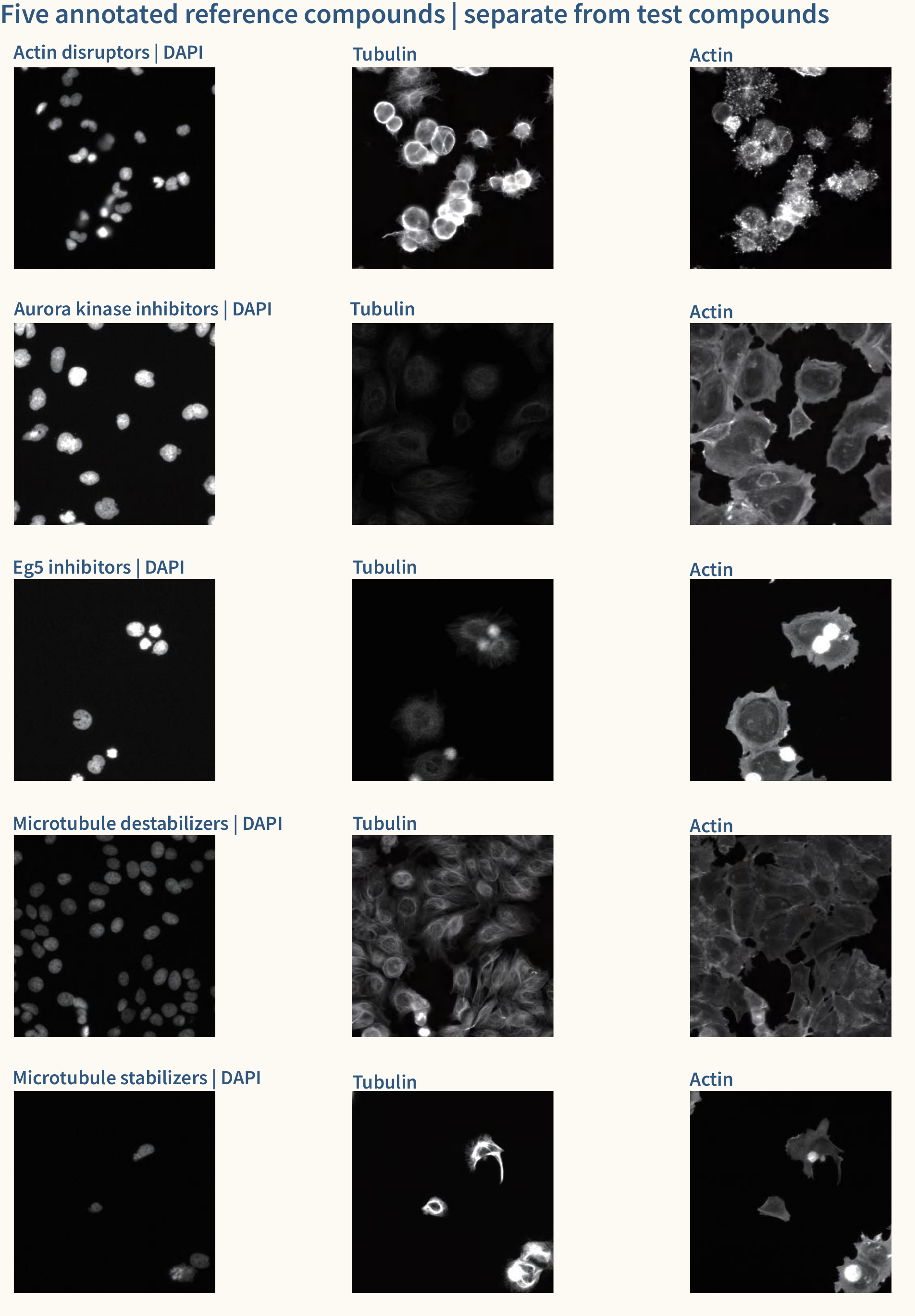}
  \caption{\textbf{Real microscopy evidence.} Reference compounds of the microscopy pilot (BBBC021), shown as DNA, tubulin, and actin channels; test compounds are disjoint from these references. The task reads real fluorescence pixels, not synthetic bar plots.}
  \label{fig:loop-atlas}
\end{figure}

Figure~\ref{fig:loop-results} and Table~\ref{tab:loop-results} report per-pilot outcomes. Initial accuracy is below 100\% in all four pilots (66.7\%--93.3\%). Acquiring the chosen measurement corrects predictions in the three numerical pilots, with nine wrong-to-right and zero right-to-wrong transitions in total; self-review alone improves no pilot and degrades two. The microscopy pilot shows no classification gain when reading real fluorescence pixels (Figure~\ref{fig:loop-atlas}). Two qualifications accompany these numbers. First, corrections are local: unqueried-readout accuracy does not improve after the calibration update, so a local fix must not be read as a general improvement of the remaining outputs. Second, the gene-pair pilot is class-skewed---38 of its 48 readouts have near-zero ground truth, so a constant near-zero baseline reaches 79.2\%, above the model's initial 77.1\%; macro-F1 and continuous errors are archived in the run records. Case-bootstrap intervals in the run records describe small-sample replay uncertainty, not independent biological replicates.

\begin{table}[!ht]
  \caption{Per-pilot accuracy (\%) before and after acquiring one chosen measurement. Corrections count wrong-to-right transitions against zero right-to-wrong transitions in every pilot.}
  \label{tab:loop-results}
  \centering
  \small
  \setlength{\tabcolsep}{4pt}
  \begin{tabular}{@{}lrrrrr@{}}
    \toprule
    Pilot & Readouts & Initial & Self-review & After evidence & Corrections \\
    \midrule
    Gene interactions & 48 & 77.1 & 77.1 & 89.6 & $+6$ / $-0$ \\
    \rowcolor{rowgray} Drug combinations & 30 & 93.3 & 80.0 & 100.0 & $+2$ / $-0$ \\
    Real microscopy & 9 & 66.7 & 66.7 & 66.7 & $0$ / $-0$ \\
    \rowcolor{rowgray} Signaling & 30 & 66.7 & 60.0 & 70.0 & $+1$ / $-0$ \\
    \bottomrule
  \end{tabular}
\end{table}

\subsection{Scope of the Pilots}
\label{app:loop-scope}

These are one-step replay loops: simulator weights are frozen, no new wet-lab measurement is performed, and corrections update a response-calibration layer rather than the simulator. Case counts are small (36 cases, 108 archived model calls), drug-combination accuracy saturates after evidence, and the microscopy pilot has no simulator at all. The pilots therefore test evidence acquisition and prediction revision on real biological outputs. They do not validate a complete virtual-cell discovery loop, do not establish that \textsc{OmniVCBench} scores transfer to loop performance, and do not isolate the contribution of multimodal inputs, since numerical tasks also expose numeric tables. Free-text mechanisms are retained as explanation traces without an independent gold standard, so no mechanism-discovery claim is made.

\section{AIVC-Judge Dimension-Wise Results}
\label{app:judge-dimensions}

Table~\ref{tab:judge-dimensions} reports dimension scores alongside the raw overall metric used in the main results. Overall scores are holistic judgments, so their means need not equal averages of dimension means. Proprietary models lead on every dimension in this evaluated pool. The production L3 profile separates judgment correctness from source-referenced argumentation and evidence weighing. For GPT-5.6-sol, these scores are 4.04, 1.92, and 1.67, respectively. The human audit shows weaker agreement on the latter two dimensions than on overall scores (Appendix~\ref{app:human-judge-agreement}). We therefore use them to describe behavior under the critique-oriented rubric rather than as a validated scale of proposal quality. Cases~G--I (Appendix~\ref{app:case-studies}) separately examine observational fidelity and the scientific defensibility of proposed hypotheses. For Qwen3-VL-8B, SFT changes the L3 dimensions from 2.91/1.25/1.20 to 3.01/1.44/1.33.

\begin{table}[!ht]
  \caption{AIVC-Judge dimension scores (1--5; higher is better) by interpretation level. L1: C = correctness, ES = evidence support. L2: F = faithfulness, CC = causal completeness, MG = mechanistic granularity, EC = evidence consistency. L3: J = judgment correctness, A = argumentation quality, EW = evidence weighing. $^{\dagger}$ denotes SFT on \textsc{OmniVCTrain}.}
  \label{tab:judge-dimensions}
  \centering
  \small
  \setlength{\tabcolsep}{2.8pt}
  \begin{tabular}{@{}lrr@{\hspace{6pt}}rrrr@{\hspace{6pt}}rrr@{}}
    \toprule
    & \multicolumn{2}{c}{L1} & \multicolumn{4}{c}{L2} & \multicolumn{3}{c}{L3} \\
    \cmidrule(lr){2-3}\cmidrule(lr){4-7}\cmidrule(l){8-10}
    Model & C & ES & F & CC & MG & EC & J & A & EW \\
    \midrule
    GPT-5.6-sol & \textbf{3.35} & \textbf{3.35} & \textbf{3.88} & \textbf{3.72} & \textbf{3.63} & \textbf{3.85} & \textbf{4.04} & 1.92 & 1.67 \\
    \rowcolor{rowgray} GPT-5.6-luna & 3.23 & 3.23 & 3.70 & 3.49 & 3.46 & 3.66 & 3.94 & 1.92 & 1.67 \\
    Grok-4.6 & 3.23 & 3.25 & 3.64 & 3.38 & 3.36 & 3.62 & 3.87 & 1.71 & 1.53 \\
    \rowcolor{rowgray} Claude-Sonnet-4-5 & 2.52 & 2.55 & 3.07 & 2.99 & 2.97 & 3.10 & 3.64 & \textbf{1.99} & \textbf{1.73} \\
    Qwen3-VL-8B-SFT$^{\dagger}$ & 2.50 & 2.55 & 2.71 & 2.25 & 2.34 & 2.50 & 3.01 & 1.44 & 1.33 \\
    \rowcolor{rowgray} Qwen3-VL-8B & 2.45 & 2.50 & 2.89 & 2.30 & 2.32 & 2.68 & 2.91 & 1.25 & 1.20 \\
    InternVL3.5-8B & 2.66 & 2.69 & 2.92 & 2.03 & 1.97 & 2.59 & 2.66 & 1.15 & 1.12 \\
    \rowcolor{rowgray} Qwen3-VL-4B & 2.33 & 2.38 & 2.77 & 2.11 & 2.13 & 2.51 & 2.79 & 1.27 & 1.23 \\
    Qwen2.5-VL-7B & 2.23 & 2.27 & 2.45 & 1.99 & 2.01 & 2.29 & 2.57 & 1.25 & 1.19 \\
    \rowcolor{rowgray} LLaVA-OneVision-7B & 2.30 & 2.35 & 2.84 & 1.53 & 1.46 & 2.02 & 2.30 & 1.09 & 1.07 \\
    LLaVA-Med-Mistral-7B & 2.13 & 2.15 & 2.33 & 1.47 & 1.43 & 1.80 & 2.08 & 1.16 & 1.12 \\
    \bottomrule
  \end{tabular}
\end{table}

% \clearpage
\section{Limitations and Outlook}
\label{app:limitations}

\textbf{Scope of inference.} \textsc{OmniVCBench} evaluates interpretation of closed, literature-derived evidence. Its 1,080 source papers were published in Nature Communications during 2011--2017, and mouse and human studies, microscopy, and immunoblot assays dominate the corpus (Tables~\ref{tab:subject-dist} and~\ref{tab:coverage}). This narrow and dated source distribution is the benchmark's largest limitation, and the resulting comparisons apply to this source and assay distribution. Several considerations nonetheless qualify its severity. The corpus was filtered by MLLM screening followed by manual review toward topics close to virtual-cell construction---single-cell and spatial omics, perturbation atlases, and CRISPR screens---so the evaluated evidence centers on the experimental techniques that AIVC interpretation must handle. Older publication years do not make the underlying conclusions less important: these are peer-reviewed studies whose figures remain valid scientific evidence, and the per-year analysis reports no monotone accuracy trend across 2011--2017 for any evaluated model (Table~\ref{tab:year-accuracy}), providing no evidence that item difficulty drifts with source age within this range. Even where models may have encountered these papers during pretraining, performance remains far from saturated on both tracks (the strongest model reaches 55.1\% MCQ accuracy and 3.28/5 under AIVC-Judge), so potential exposure alone does not resolve the tasks; whether memorization assists a subset of items nonetheless remains unquantified. Source-level filtering and the release-level audit found no confirmed overlap within the assembled resources (Appendix~\ref{app:leakage-audit}); they do not assess pretraining exposure to the published literature. As future work, we plan to extend the source pool to diverse open-access journals from 2018 to the present, for which the construction pipeline and curation criteria transfer directly. Finally, L3 operationalizes bounded hypothesis proposal and assessment. It measures neither experimental execution nor the novelty of a discovery, and MCQ key agreement measures selection among supplied alternatives rather than free hypothesis generation.

\textbf{Text-only shortcuts.} The no-image control (Appendix~\ref{app:item-audits}) shows that GPT-5.6-sol retains 34.7\% MCQ accuracy without the figure: the question text, the option content, domain priors, and potential memorization of the source literature all contribute substantially to MCQ performance. This text-derived signal is a genuine limitation of the controlled track. The 22.6-point gap to the 57.3\% with-image score nonetheless shows that visual evidence carries a large share of the achievable accuracy, and open-response evaluation further constrains purely text-driven answering, since generated explanations are checked against the source evidence by the judge.

\textbf{Scoring and answer validity.} Human scoring provides a check on the production judge under the same reference-conditioned protocol: the panel-mean raw overall score correlates with AIVC-Judge at Spearman $\rho=0.877$ over 884 paired responses (Appendix~\ref{app:human-judge-agreement}). Human scorers applied the same rubric without the original image, so this agreement validates the shared textual evidence and rubric application rather than the judge's image reading; same-input expert scoring remains an open validation step. Agreement is lower on L3 ($\rho=0.709$, 250 responses), and the historical argumentation and evidence-weighing dimensions assess critique and justification. These dimensions are diagnostics under that rubric, not a validated scale of hypothesis-generation ability. L3 overall scores can therefore conflate proposal quality with unrequested critique, and they are not directly comparable with L1/L2 totals, which use different dimensions. A rubric that directly rewards proposal quality---observational fidelity, falsifiable specificity, and discriminative test design---has been piloted on a frozen 24-item L3 subset, where it preserves candidate rankings and improves cross-annotator agreement (Appendix~\ref{app:proposal-pilot}); full-benchmark rescoring under it remains future work. The scoring evidence includes caption, context, and reference information beyond the answering model's image and question. Moreover, AIVC-Judge contributes to both distractor filtering and open-response scoring, so cross-track correlation is an internal consistency measure.

\textbf{System coverage.} The evaluated systems are multimodal language models. The benchmark assesses their interpretation of cellular experiments, rather than the state-prediction accuracy of cell foundation models. The open-weight pool spans 2B--8B parameters, so comparisons with the proprietary systems mix model capacity, compute, and inference configuration and should not be read as open- versus closed-identity effects. Connecting cell models to language interfaces defines an extension of this evaluation setting.

\textbf{Adaptation headroom.} \textsc{OmniVCTrain} provides 548k source-aligned examples, yet our adaptation study covers only LoRA fine-tuning and retrieval augmentation on a single 8B backbone, and the observed gains are configuration-dependent rather than a ceiling on the corpus. We currently lack the computational resources to explore stronger recipes---larger backbones, longer training schedules, preference optimization, or agentic training on evidence-grounded tasks. The corpus's potential for improving evidence-grounded interpretation therefore remains largely untapped, and we view it as a resource for the community as much as a baseline for this paper.

\section{Case Studies of Reasoning and Scoring}
\label{app:case-studies}

We examine nine cases selected purposively after inspecting the stratified 300-item sample (Appendix~\ref{app:item-audits}), with three cases per level from nine different source papers. Cases~A--G illustrate response-level evidence and reasoning errors; Cases~H and~I examine L3 items where a defensible open hypothesis coexists with a flawed option selection. Each exhibit reproduces the source figure, item, reference answer, and six options, with the answer key highlighted in green. The examples provide qualitative diagnoses, not estimates of error prevalence. Any reported judge scores are archived values under the production rubric.

\textbf{Preserving relations in evidence.} Cases~A and~C concern comparison polarity and effect direction. Cases~D and~E require explanations to preserve intervention--reversal and reciprocal-perturbation relations. Case~F requires the correct association of protein pairs, panels, and spatial overlap. These diagnoses describe inconsistencies in the responses without assigning a separate causal contribution to perception, knowledge, or reasoning.

\textbf{Separating observations from predictions.} Case~B concerns a prediction that contradicts the intervention premise. Case~G distinguishes a falsifiable proposal from an unsupported observational premise or necessity claim. In Cases~H and~I, the models' open answers make testable predictions that extend the displayed observations, while their selected options contain direction or mechanism errors; the two response formats therefore carry different diagnoses for the same item.

\textbf{Paired-track interpretation.} Cases~B and~E show that key agreement can coexist with an inconsistent open explanation; Cases~C, D, and~F show that an accurate relation in an open answer need not appear in the selected option. Cases~H and~I extend this divergence to L3 hypothesis items, where a well-formed proposal accompanies a flawed selection. This analysis motivates reading the two tracks together, with explicit attention to the options and scoring rule.

\subsection{Single- versus multi-subfigure difficulty}
\label{app:multi-subfigure}

Multi-subfigure items account for 3,094/6,077 items (50.9\%). All five models in Table~\ref{tab:multi-subfigure} are at least as accurate on multi-subfigure items as on single-subfigure items, with differences of 0.4--7.8 percentage points. For GPT-5.6-sol, accuracy increases from 56.0\% to 63.7\% on L1 and from 58.0\% to 62.6\% on L2, and changes from 44.7\% to 44.1\% on L3. Thus, panel count alone does not define task difficulty; Case~F illustrates the more specific requirement to preserve relations across panels.

\begin{table}[!ht]
  \caption{MCQ accuracy (\%) on single- versus multi-subfigure items (five representative models of the eleven with open-response coverage).}
  \label{tab:multi-subfigure}
  \centering
  \small
  \begin{tabular}{@{}lrrr@{}}
    \toprule
    Model & Single & Multi & $\Delta$ (pp) \\
    \midrule
    GPT-5.6-sol & 54.5 & 55.7 & +1.2 \\
    \rowcolor{rowgray} Claude-Sonnet-4-5 & 45.4 & 53.2 & \textbf{+7.8} \\
    Qwen3-VL-8B & 33.6 & 34.0 & +0.4 \\
    \rowcolor{rowgray} InternVL3.5-8B & 29.6 & 30.9 & +1.3 \\
    LLaVA-OneVision-7B & 25.0 & 26.1 & +1.1 \\
    \bottomrule
  \end{tabular}
\end{table}

\clearpage
\subsection{Case A: LGALS3BP readout polarity (L1)}
\label{app:case-a}

\textit{Source.} LGALS3BP silencing and abnormal centriolar structures in interphase versus mitotic cells (panel b); L1 / Predict; \citet{fogeron2013CASE1}.

\begin{figure}[!ht]
  \centering
  \includegraphics[width=\linewidth,height=0.66\textheight,keepaspectratio]{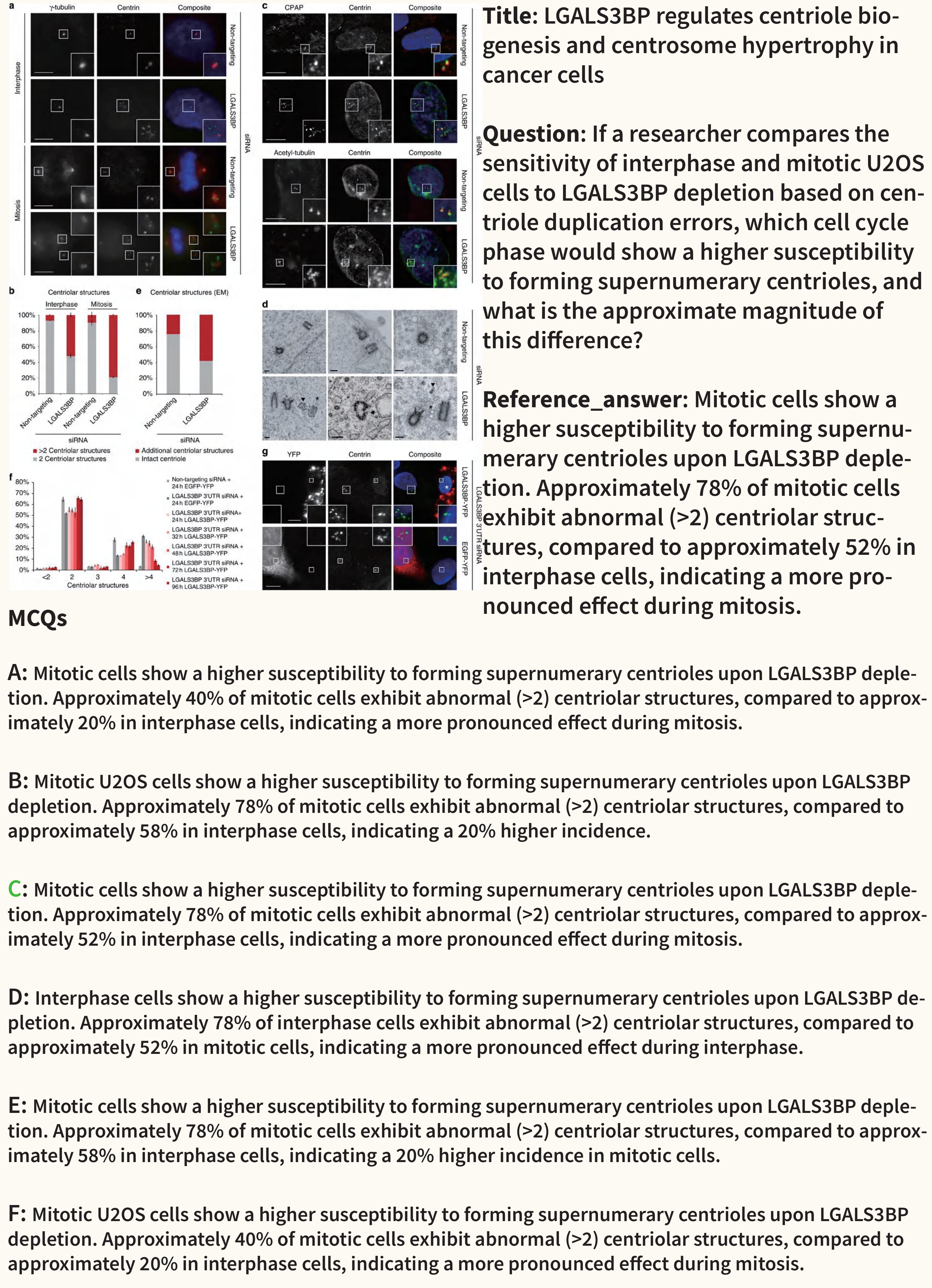}
  \caption{\textbf{Case A exhibit (L1 / Predict).} Source figure, item, reference answer, and six MCQ options; the keyed option (C) is highlighted in green.}
  \label{fig:case-a}
\end{figure}

Panel b marks cells with ${>}2$ centriolar structures in red. After LGALS3BP siRNA, this fraction is approximately 52\% in interphase and 78\% in mitosis. GPT-5.6-sol preserves the comparison (``about 80 versus 50''; overall 5; keyed C). Claude-Sonnet-4-5 and Qwen3-VL-8B reverse it in their open responses (both overall 1). Both select E, whose error is numerical rather than directional, so the open-response diagnosis and the selected letter carry different information. The measured outcome is the number of centrin-positive structures, not the formation of mature, functional centrioles.

\clearpage
\subsection{Case B: IFNAR1 counterfactual across formats (L1)}
\label{app:case-b}

\textit{Source.} IFNAR1 surface loss after stimulation, with cycloheximide pretreatment supplied as the question premise (panels f/g); L1 / Predict; \citet{chmiest2016CASE2}.

\begin{figure}[!ht]
  \centering
  \includegraphics[width=\linewidth,height=0.66\textheight,keepaspectratio]{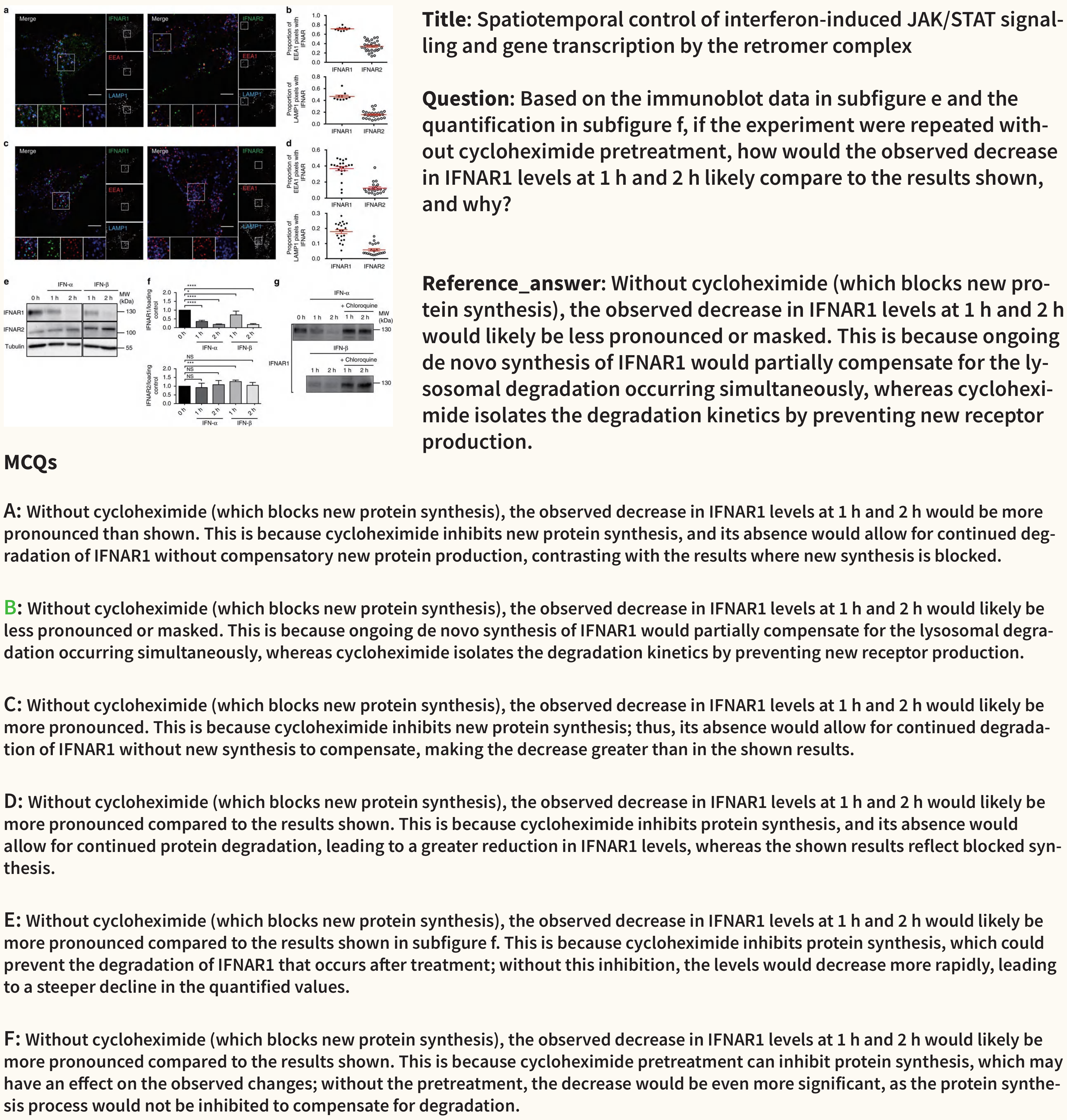}
  \caption{\textbf{Case B exhibit (L1 / Predict).} Source figure, item, reference answer, and six MCQ options; the keyed option (B) is highlighted in green.}
  \label{fig:case-b}
\end{figure}

Under the question's premise, removing the synthesis inhibitor allows resynthesis to offset degradation, predicting a weaker net IFNAR1 decline at 1 and 2\,h. Qwen3-VL-8B instead predicts stronger loss by invoking absent compensatory synthesis, contradicting that premise (overall 1). GPT-5.6-sol and Claude-Sonnet-4-5 preserve the expected direction (both 5). All three select the keyed B, illustrating that a correct choice can coexist with an inconsistent generated explanation. The prediction follows from the counterfactual premise and the inhibitor's role; the displayed panels report the original treatment condition.

\clearpage
\subsection{Case C: C8orf4 effect direction across models (L1)}
\label{app:case-c}

\textit{Source.} C8orf4 knockout in Huh7 (panel b) and knockdown in primary HCC (panel f) and sphere-initiating-cell frequency; L1 / Predict; \citet{zhu2015CASE3}.

\begin{figure}[!ht]
  \centering
  \includegraphics[width=\linewidth,height=0.66\textheight,keepaspectratio]{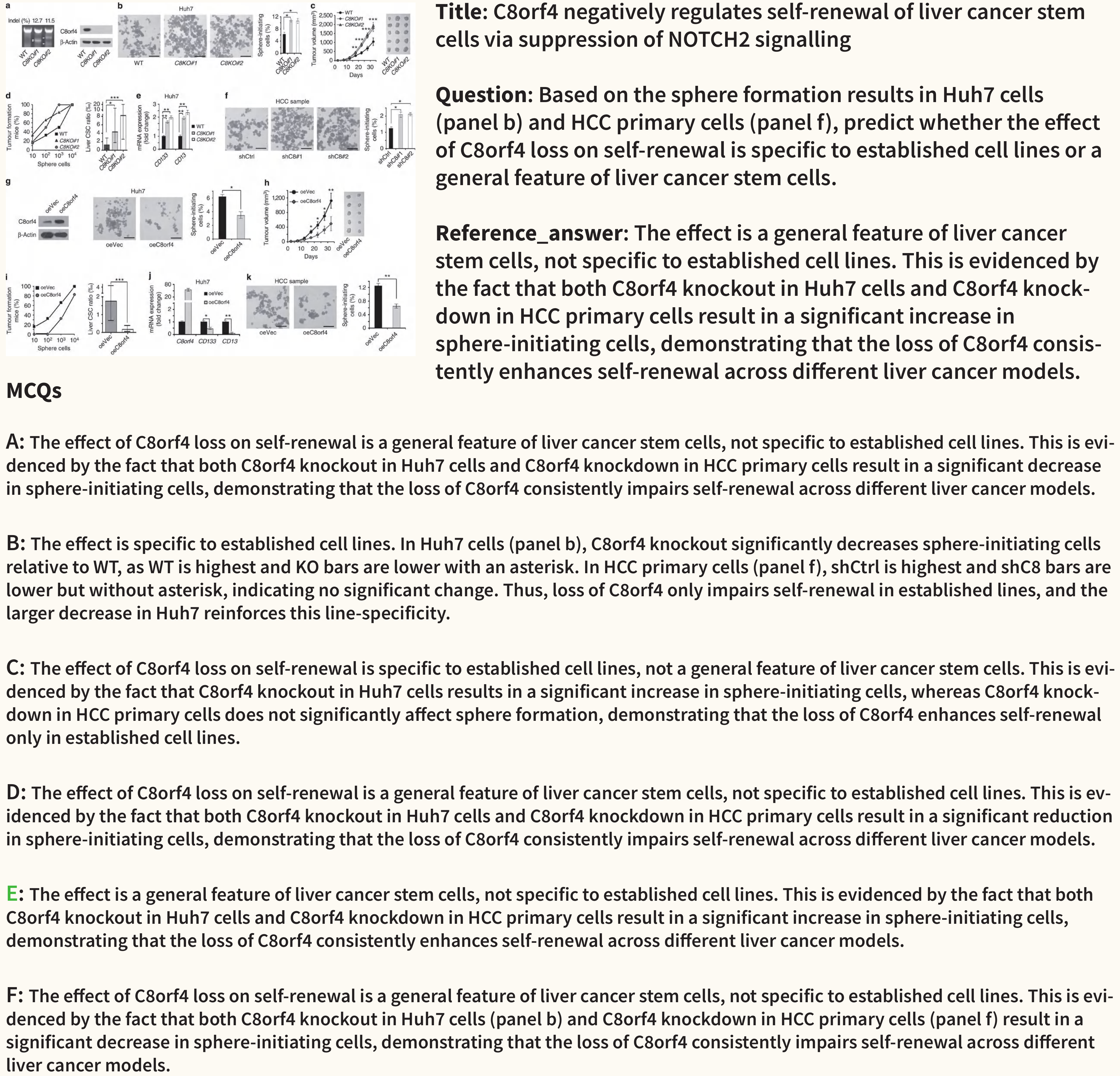}
  \caption{\textbf{Case C exhibit (L1 / Predict).} Source figure, item, reference answer, and six MCQ options; the keyed option (E) is highlighted in green.}
  \label{fig:case-c}
\end{figure}

C8orf4 knockout in Huh7 cells and knockdown in primary HCC both increase the sphere-initiating-cell fraction. Qwen3-VL-8B notes the shared effect but describes it as \emph{impairing} sphere formation (overall 2). GPT-5.6-sol and Claude-Sonnet-4-5 preserve the increase (both 5). GPT-5.6-sol selects the keyed E, whereas Claude-Sonnet-4-5 and Qwen3-VL-8B select F; Claude's open answer and selected option therefore disagree in direction. In fixed-prompt image controls, GPT-5.6-sol changes from E with the original image to A with no image or a swapped image. This is an item-level observation of input sensitivity. The biological comparison concerns sphere formation in these two experimental systems.

\clearpage
\subsection{Case D: ROS--TGFBRI intervention--reversal chain (L2)}
\label{app:case-d}

\textit{Source.} ROS, TGFBRI inhibition, and paired-daughter division symmetry (panels a/d; b/c auxiliary); L2 / Explain; \citet{hinge2017CASE4}.

\begin{figure}[!ht]
  \centering
  \includegraphics[width=\linewidth,height=0.66\textheight,keepaspectratio]{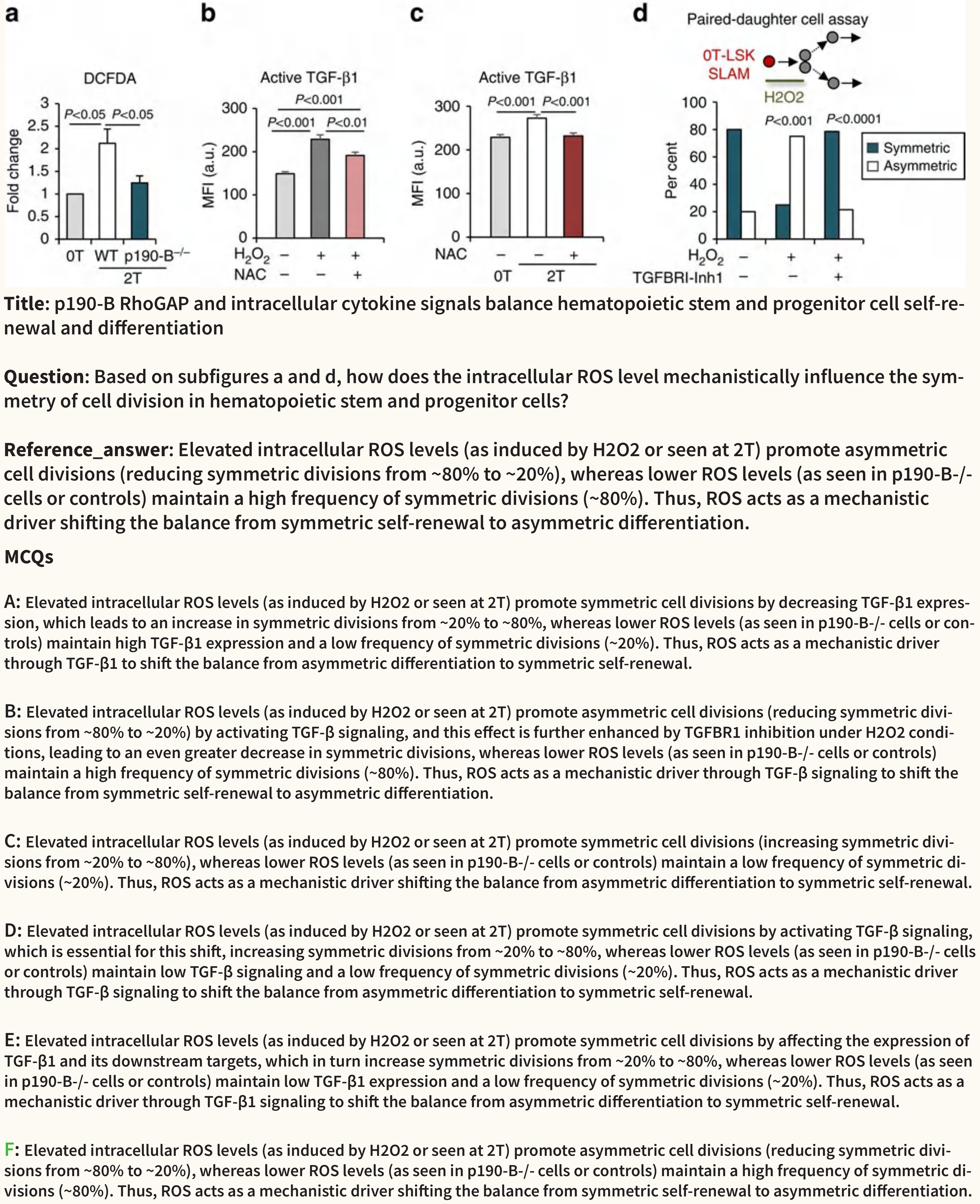}
  \caption{\textbf{Case D exhibit (L2 / Explain).} Source figure, item, reference answer, and six MCQ options; the keyed option (F) is highlighted in green.}
  \label{fig:case-d}
\end{figure}

In panel d, approximately 80\% of control daughter pairs divide symmetrically. H$_2$O$_2$ reduces this fraction to approximately 25\%, and TGFBRI inhibition restores the distribution towards control. GPT-5.6-sol and Claude-Sonnet-4-5 link ROS to TGF-$\beta$-dependent asymmetric division (both overall 5). Qwen3-VL-8B instead states that ROS \emph{promotes symmetric} division, reversing the observed direction (overall 1). GPT-5.6-sol selects the keyed F. Claude-Sonnet-4-5's B preserves the ROS direction but states that the inhibitor enhances the shift, conflicting with the reversal. The evidence supports an H$_2$O$_2$-associated, TGFBRI-inhibitor-sensitive change in this assay; it does not identify a unique direct molecular target.

\clearpage
\subsection{Case E: CCP1 sign from reciprocal perturbations (L2)}
\label{app:case-e}

\textit{Source.} CCP1 deletion and overexpression versus nuclear Msn2 (panels d/g); L2 / Explain; \citet{bodvard2017CASE5}.

\begin{figure}[!ht]
  \centering
  \includegraphics[width=\linewidth,height=0.66\textheight,keepaspectratio]{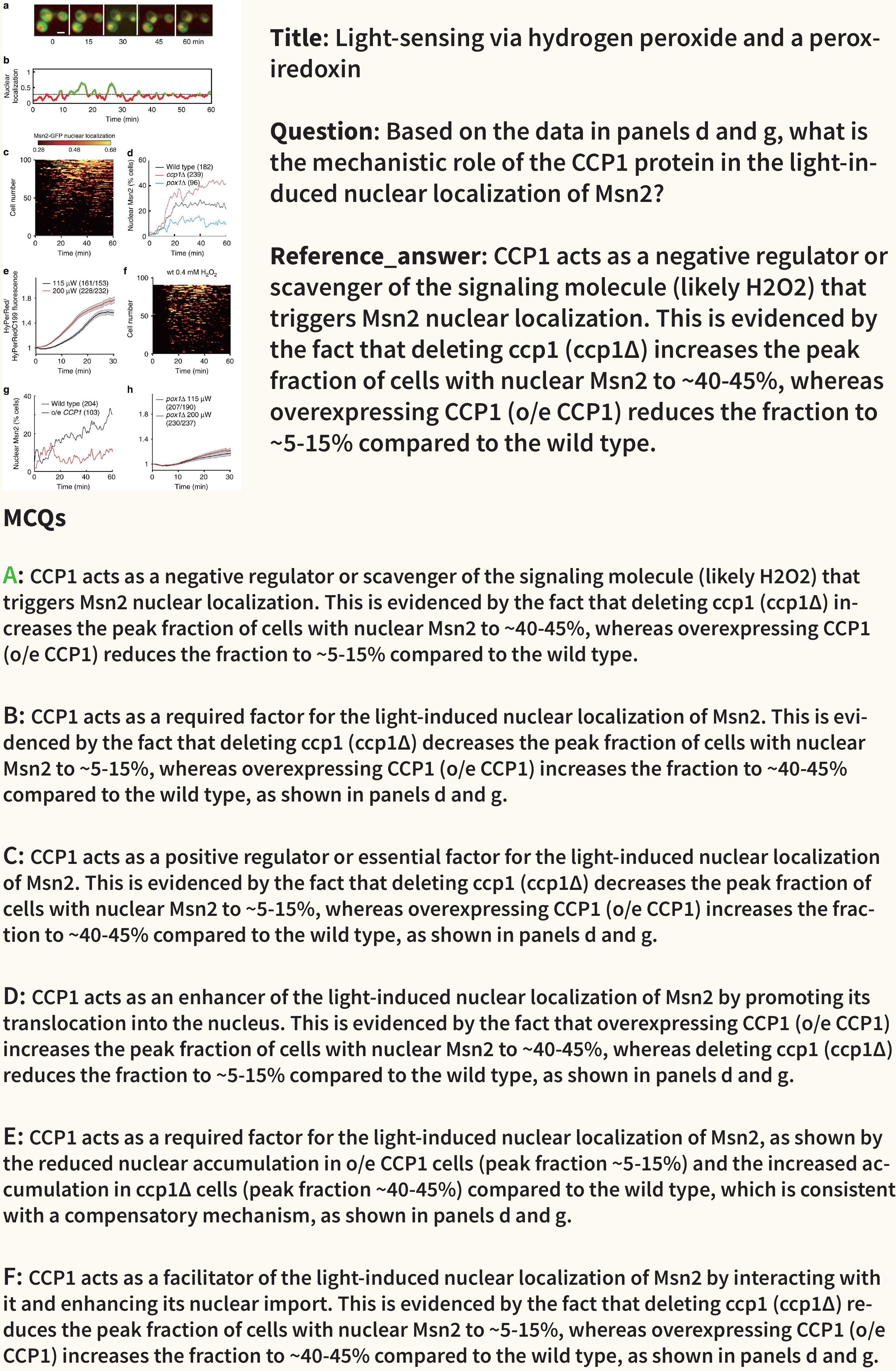}
  \caption{\textbf{Case E exhibit (L2 / Explain).} Source figure, item, reference answer, and six MCQ options; the keyed option (A) is highlighted in green.}
  \label{fig:case-e}
\end{figure}

CCP1 deletion increases the nuclear-Msn2 fraction relative to the wild type in panel d, whereas overexpression decreases it relative to the wild type in panel g. These reciprocal perturbations support negative regulation. Qwen3-VL-8B instead states that CCP1 is required and that deletion impairs translocation (overall 1). GPT-5.6-sol and Claude-Sonnet-4-5 preserve the regulatory sign (overall 5 and 3), and all three select the keyed A. Claude's faithfulness and evidence-consistency scores are both 5, separating correct direction from the overall assessment of mechanistic detail. Comparisons use each panel's own wild-type baseline, and overexpression attenuates the signal.

\clearpage
\subsection{Case F: BRCA1 spatial integration (L2)}
\label{app:case-f}

\textit{Source.} BRCA1--RIF1 versus BRCA1--53BP1 spatial overlap (panels a/c); L2 / Explain; \citet{ha2017CASE6}.

\begin{figure}[!ht]
  \centering
  \includegraphics[width=\linewidth,height=0.66\textheight,keepaspectratio]{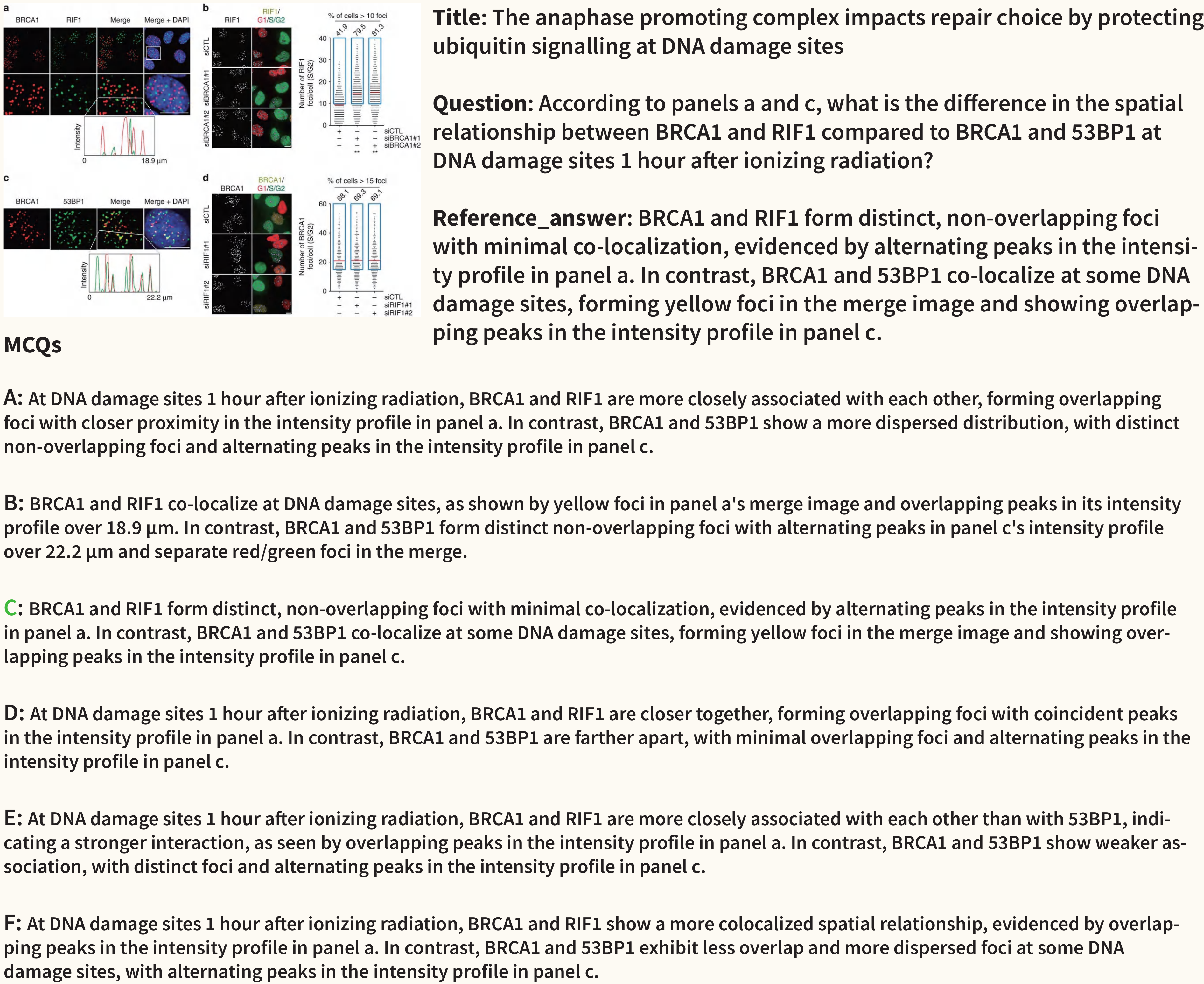}
  \caption{\textbf{Case F exhibit (L2 / Explain).} Source figure, item, reference answer, and six MCQ options; the keyed option (C) is highlighted in green.}
  \label{fig:case-f}
\end{figure}

Panel a shows largely separate or adjacent BRCA1--RIF1 signals; panel c shows partial BRCA1--53BP1 overlap at some sites. Qwen3-VL-8B reverses these relations (overall 1; F). GPT-5.6-sol and Claude-Sonnet-4-5 state the observed comparison (both overall 5), although only GPT-5.6-sol selects the keyed C and Claude selects B. The task requires preserving the association between each protein pair and its spatial relation across panels. These image readouts describe local overlap, which is distinct from direct physical binding or a global colocalization percentage.

\clearpage
\subsection{Case G: hair-regeneration hypothesis on a faulty premise (L3)}
\label{app:case-g}

\textit{Source.} Hair regeneration with SB versus PHM populations; pigmentation and shaft morphology (panels c/d); L3 / Discover; \citet{toyoshima2012CASE7}.

\begin{figure}[!ht]
  \centering
  \includegraphics[width=\linewidth,height=0.66\textheight,keepaspectratio]{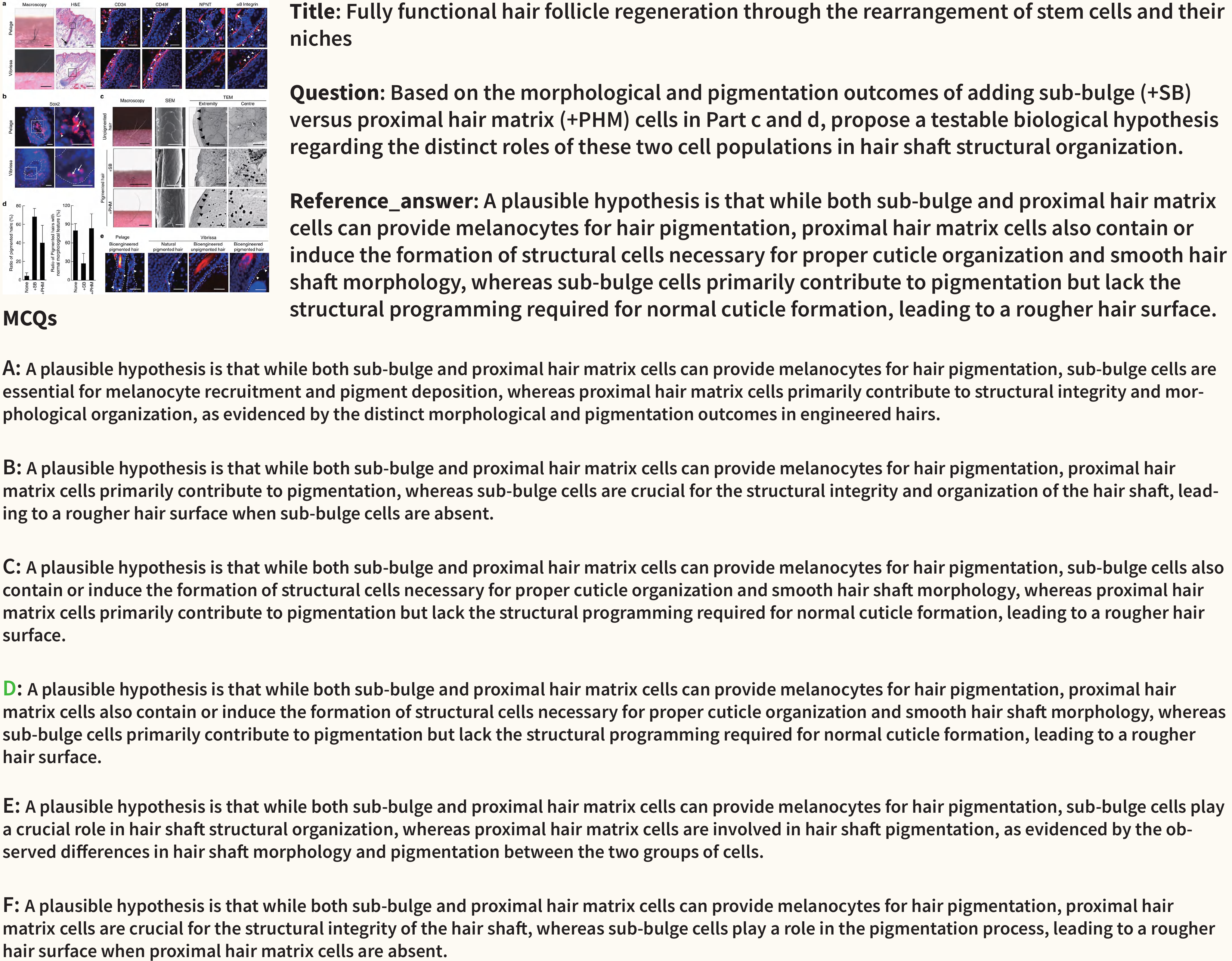}
  \caption{\textbf{Case G exhibit (L3 / Discover).} Source figure, item, reference answer, and six MCQ options; the keyed option (D) is highlighted in green.}
  \label{fig:case-g}
\end{figure}

The source context reports pigmentation recovery of 68.3\% with SB and 40.1\% with PHM, and normal morphology of 27.0\% and 83.3\%, respectively. GPT-5.6-sol proposes different contributions to pigmentation and shaft organization, with selective addition or ablation as a test (overall 4; keyed D). Claude-Sonnet-4-5 builds a cortex/medulla hypothesis on a melanin-localization contrast, although dark granules appear in central TEM fields under both conditions (overall 1; C). The identifiable error is the asserted observation. Qwen3-VL-8B captures the direction but describes PHM as ``essential,'' a necessity claim stronger than the comparison supports (overall 3; C). The observations establish contrasting regeneration outcomes; the proposed division of cellular roles remains a hypothesis.

\clearpage
\subsection{Case H: YAP--Ect2/Fgd3 paired-track divergence (L3)}
\label{app:case-h}

\textit{Source.} Ect2/Fgd3 induction under activated YAP with EtOH/HTVi injury (panels d/e; f is a schematic); L3 / Discover; \citet{miyamura2017CASE8}.

\begin{figure}[!ht]
  \centering
  \includegraphics[width=\linewidth,height=0.66\textheight,keepaspectratio]{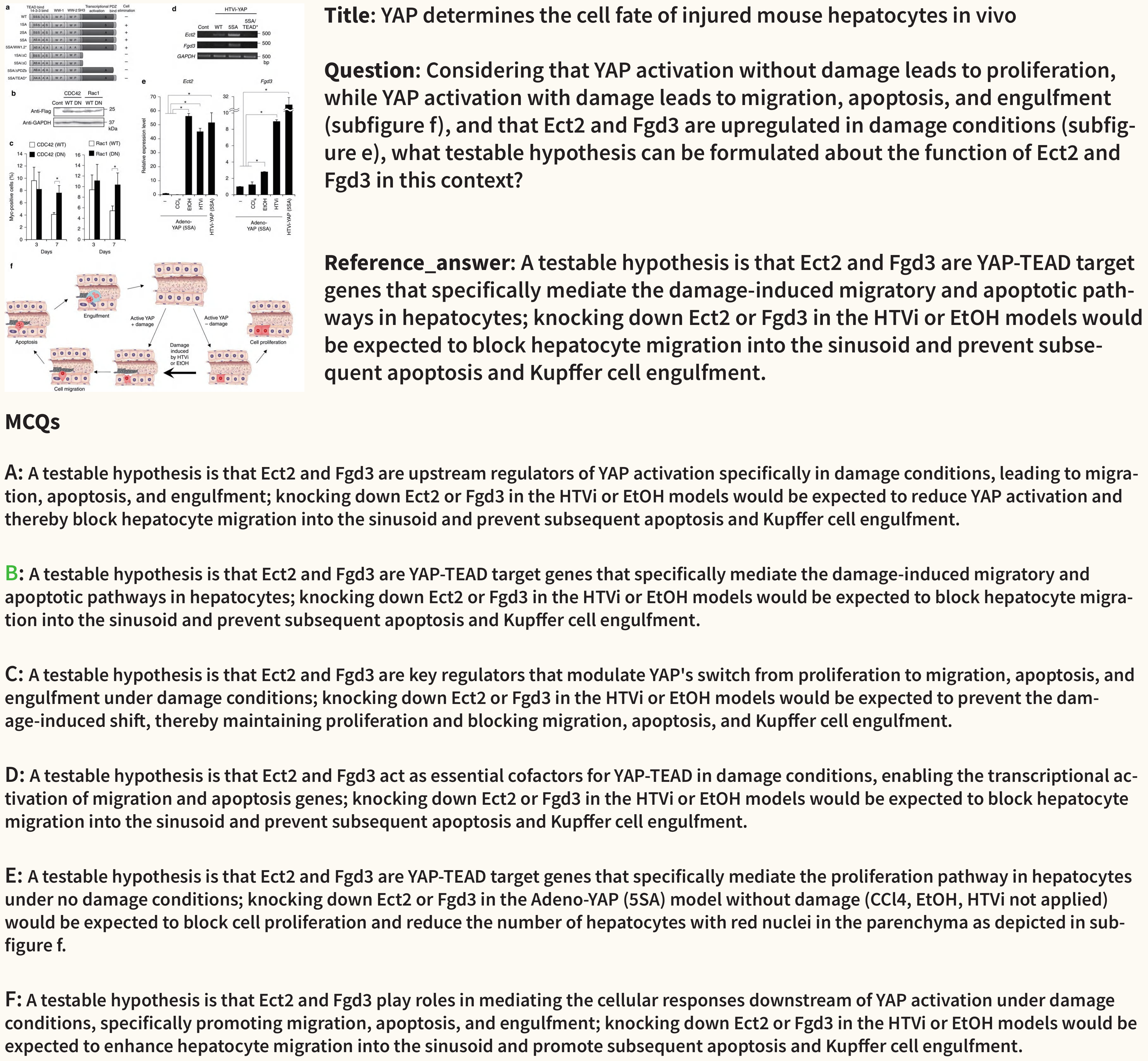}
  \caption{\textbf{Case H exhibit (L3 / Discover).} Source figure, item, reference answer, and six MCQ options; the keyed option (B) is highlighted in green. Option F's knockdown prediction contradicts its own mechanism claim, as discussed below.}
  \label{fig:case-h}
\end{figure}

\textbf{Observed evidence and hypotheses.} Panel e reports Ect2/Fgd3 expression changes under activated YAP with injury, and panel f summarizes the proposed mechanism. GPT-5.6-sol proposes knockdown with migration, elimination, and proliferation readouts; Claude-Sonnet-4-5 also proposes an intervention test. These answers supply falsifiable extensions of the expression evidence. Restoration of proliferation is a separate predicted readout, and expression association does not establish direct promoter binding or a loss-of-function effect.

\textbf{Paired-track divergence.} The key is B. GPT-5.6-sol selects F, Claude-Sonnet-4-5 selects B, and Qwen3-VL-8B selects C. Option F identifies Ect2/Fgd3 as downstream mediators of the damage response, yet predicts that their knockdown would \emph{enhance} migration, apoptosis, and engulfment---contradicting both its own mechanism claim and the direction implied by the expression evidence. GPT-5.6-sol's open answer proposes exactly the knockdown test with the expected direction, so its selected option reverses the prediction that its own hypothesis implies. The case therefore separates the quality of the generated hypothesis from the correctness of the selected option.

\clearpage
\subsection{Case I: integrin--Lck paired-track divergence (L3)}
\label{app:case-i}

\textit{Source.} Lck inhibition/knockdown and downstream paxillin/CrkII phosphorylation, with the integrin$\to$Lck source model (panels a/b/h); L3 / Discover; \citet{ness2013CASE9}.

\begin{figure}[!ht]
  \centering
  \includegraphics[width=\linewidth,height=0.66\textheight,keepaspectratio]{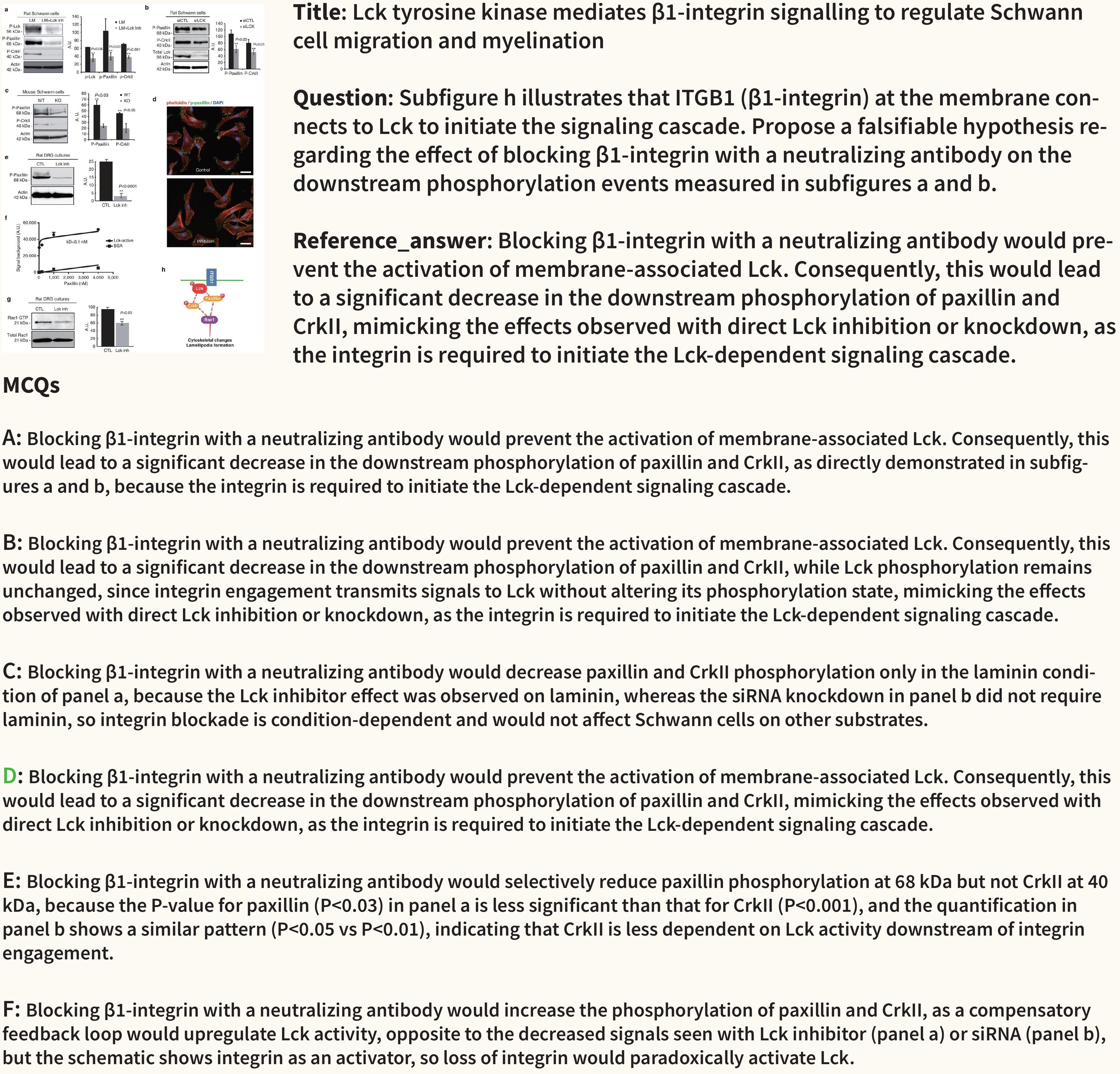}
  \caption{\textbf{Case I exhibit (L3 / Discover).} Source figure, item, reference answer, and six MCQ options; the keyed option (D) is highlighted in green. Option B contradicts the integrin$\to$Lck initiation shown in panel h, as discussed below.}
  \label{fig:case-i}
\end{figure}

\textbf{Observed evidence and hypotheses.} Panels a and b report Lck inhibition and knockdown, while panel h depicts integrin$\to$Lck signalling. Predicting reduced paxillin/CrkII phosphorylation after $\beta$1-integrin antibody blockade is a testable extension of this pathway. GPT-5.6-sol adds an isotype-control comparison and stable total-protein levels as control expectations; Claude-Sonnet-4-5 proposes a conditional pathway test, and Qwen3-VL-8B predicts the same phosphorylation direction. These are intervention predictions, not observations of an antibody experiment in panels a and b.

\textbf{Paired-track divergence.} The key is D. GPT-5.6-sol selects B, while Claude-Sonnet-4-5 and Qwen3-VL-8B select D. Option B predicts reduced paxillin/CrkII phosphorylation but claims that Lck phosphorylation remains unchanged under integrin blockade, contradicting panel h, in which $\beta$1-integrin connects to Lck to initiate the cascade, and contradicting the kinase's phosphorylation-dependent activation. GPT-5.6-sol's open answer supplies a well-formed intervention prediction with appropriate controls, yet its selected option carries this mechanism error. Option A separately conflates the proposed antibody experiment with the displayed panels by treating its effect as directly demonstrated.

\clearpage
\section{Ethics, License, and Data Availability}
\label{app:ethics}

\textbf{Data provenance.} All benchmark items are derived from figures and experimental contexts of peer-reviewed publications, assembled through the OmniScience corpus \citep{tao2026omniscience}. No new biological experiments were conducted, and the benchmark contains no personal or otherwise identifiable information.

\textbf{License.} \textsc{OmniVCBench} and \textsc{OmniVCTrain} are released under the Creative Commons Attribution--NonCommercial--ShareAlike 4.0 International license (CC BY-NC-SA 4.0), the same license as OmniScience \citep{tao2026omniscience}; we impose no additional or differing terms of use. The released packages host the materialized figure crops directly. Redistribution of the underlying source figures remains subject to the terms of the original publishers.

\textbf{Intended use and misuse risks.} The benchmark is intended solely for research evaluation of multimodal scientific reasoning. Reference answers, MDHNM distractors, and AIVC-Judge scores are evaluation annotations; they and the evaluated model outputs are not clinical or experimental decision-making evidence. Evaluated models were accessed through public APIs or their open-weight releases under the respective model licenses.

\textbf{Code and demo release.} The anonymous repository at \url{https://anonymous.4open.science/r/OmniVCBench} includes the full AIVC-Judge implementation (level-specific rubrics, prompts, and the claim-decomposition scoring pipeline) together with a 300-item demo release pairing each open-response QA item with its MDHNM MCQ variant. The complete 6,077-item \textsc{OmniVCBench} and 548,450-item \textsc{OmniVCTrain} datasets will be released after the review process concludes; the demo uses the same stratified 300-item sample as Appendix~\ref{app:item-audits}.

\end{document}